\documentclass{article}

\PassOptionsToPackage{numbers,sort}{natbib}

\usepackage[main, final]{neurips_2026}

\usepackage[utf8]{inputenc}
\usepackage[T1]{fontenc}
\usepackage{xcolor}

\usepackage[pagebackref,breaklinks,colorlinks]{hyperref}
\hypersetup{
    colorlinks=true, 
    linkcolor=blue, 
    filecolor=magenta, 
    urlcolor=blue,    
    citecolor=blue, 
}

\newcommand{\appref}[1]{\hyperref[#1]{Appendix~\ref*{#1}}}

\usepackage{url}
\usepackage{booktabs}
\usepackage{amsfonts}
\usepackage{amssymb}
\usepackage{amsmath}
\usepackage{nicefrac}
\usepackage{microtype}
\usepackage{xcolor}
\usepackage{graphicx}
\usepackage{multirow}
\usepackage[ruled,vlined]{algorithm2e}
\usepackage{import}
\usepackage{subcaption}

\usepackage{multirow}
\usepackage{tikz}
\usetikzlibrary{positioning,calc,arrows.meta}
\usepackage{enumitem}
\usepackage{comment}
\usepackage{caption}
\usepackage{subcaption}
\usepackage{calc}

\newsavebox{\kpnetbox}
\newlength{\kpnetheight}
\newlength{\kpimgheight}

\title{SemGeo-Gen: Unsupervised Generation of Approximate Cross-Instance Semantic-Geometric Correspondences}

\author{
  Roy Amoyal \qquad Shira Ifergane \qquad Oren Freifeld \\[7pt]
  \small The Stein Faculty of Computer and Information Science, \\
  \small Ben-Gurion University of the Negev, Israel \\[4pt]
  \small \texttt{amoyalr@post.bgu.ac.il} \qquad
  \small \texttt{shiraif@post.bgu.ac.il}
  \qquad
  \small \texttt{orenfr@bgu.ac.il }
}

\begin{document}

\maketitle
\begin{abstract}
We introduce SemGeo-Gen, an unsupervised method for generating approximate cross-instance semantic-geometric supervision from weakly aligned 3D object collections. Our goal is not to recover exact point-level correspondences, which are often ambiguous across different instances and would require expensive, impractical manual supervision at scale. Instead, we automatically produce approximate correspondences that are semantically meaningful, geometrically consistent, and diverse enough to train modern correspondence models. Given only coarse category-level rotation alignment, SemGeo-Gen lifts multi-view DINOv2 features to 3D, aligns object instances using continuous piecewise-affine registration, and prunes candidate matches using semantic consistency. The resulting 3D correspondences can be projected into rendered views, yielding scalable 3D--3D, 3D--2D, and 2D--2D supervision without manual annotation. We validate the generated correspondences against sparse human annotations from KeypointNet, obtaining 93\% PCK@0.10 and 97\% PCK@0.15. More importantly, we show that using the generated correspondences for synthetic pretraining improves recent state-of-the-art semantic correspondence models on SPair-71k after fine-tuning on real data, with gains of up to 5 PCK@0.10 points. Together, these results indicate that approximate semantic-geometric supervision generated from 3D assets can improve real-image correspondence learning and serve as a scalable alternative to costly manual annotation. Project page:
\url{https://bgu-cs-vil.github.io/SemGeo-Gen/}.
\end{abstract}

\section{Introduction}
Establishing correspondences across different object instances within the same category is a fundamental problem in computer vision~\cite{min2019spair71k,WahCUB_200_2011,yu2021AP10k,omninocs2024,zhang2023taleTwoFeatures,zhang2024tellingLeftFromRight,shi2021skeletonmerger,you2022ukpgan,ju2025robo_abc}. It underlies semantic matching, category-level reasoning, and downstream tasks such as object manipulation, image alignment, and visual understanding~\cite{zhang2023taleTwoFeatures,zhang2024tellingLeftFromRight,mariotti2024sphericalmap,Amoyal:CVPR:2026:GSA,Hirsch:NeurIPS:2025:FastJAM,Barel:ECCV:2024:SpaceJAM,wang2019visualcorrespondencerobotics,zhu2025densematcher,you2020,luo2023DiffusionHyperFeaturesTimeSpace,hedlin2023unsupervisedstablediffusion}. Compared to same-instance matching, cross-instance matching is substantially harder: objects may differ in geometry, proportions, local structure, texture, and appearance, while sharing only partially overlapping semantic parts. This paper focuses on semantic-geometric correspondences, where matches should preserve both semantic identity and geometric role when such distinctions are meaningful.

Progress in this direction is limited by the scarcity of scalable correspondence supervision. Existing datasets~\cite{min2019spair71k,cuttano2026marco,yu2021AP10k,zhu2025densematcher,WahCUB_200_2011,omninocs2024,you2020,lou2020humanCorrespondenceConsensus,pratikakis2016shrec_16,zuffi2017menagerie,groueix2018b3dcodeddataset,dyke2019shrec_19} rely heavily on manual annotation, which is costly, sparse, and difficult to scale across categories. AP-10K~\cite{yu2021AP10k} required 13 trained annotators, while Human Correspondence Consensus~\cite{lou2020humanCorrespondenceConsensus} uses 80 professional annotators and highlights the ambiguity of semantic point annotation. This ambiguity is inherent: across different instances, exact point-level correspondences are often not well-defined. We therefore ask whether approximate correspondences, if semantically meaningful, geometrically consistent, and sufficiently diverse, can provide effective supervision at scale.

\begin{figure}[t]
    \centering
    \begin{subfigure}[t]{0.33\linewidth}
        \centering
        \includegraphics[
            width=\linewidth,
            trim={60pt 0pt 90pt 0pt},
            clip
        ]{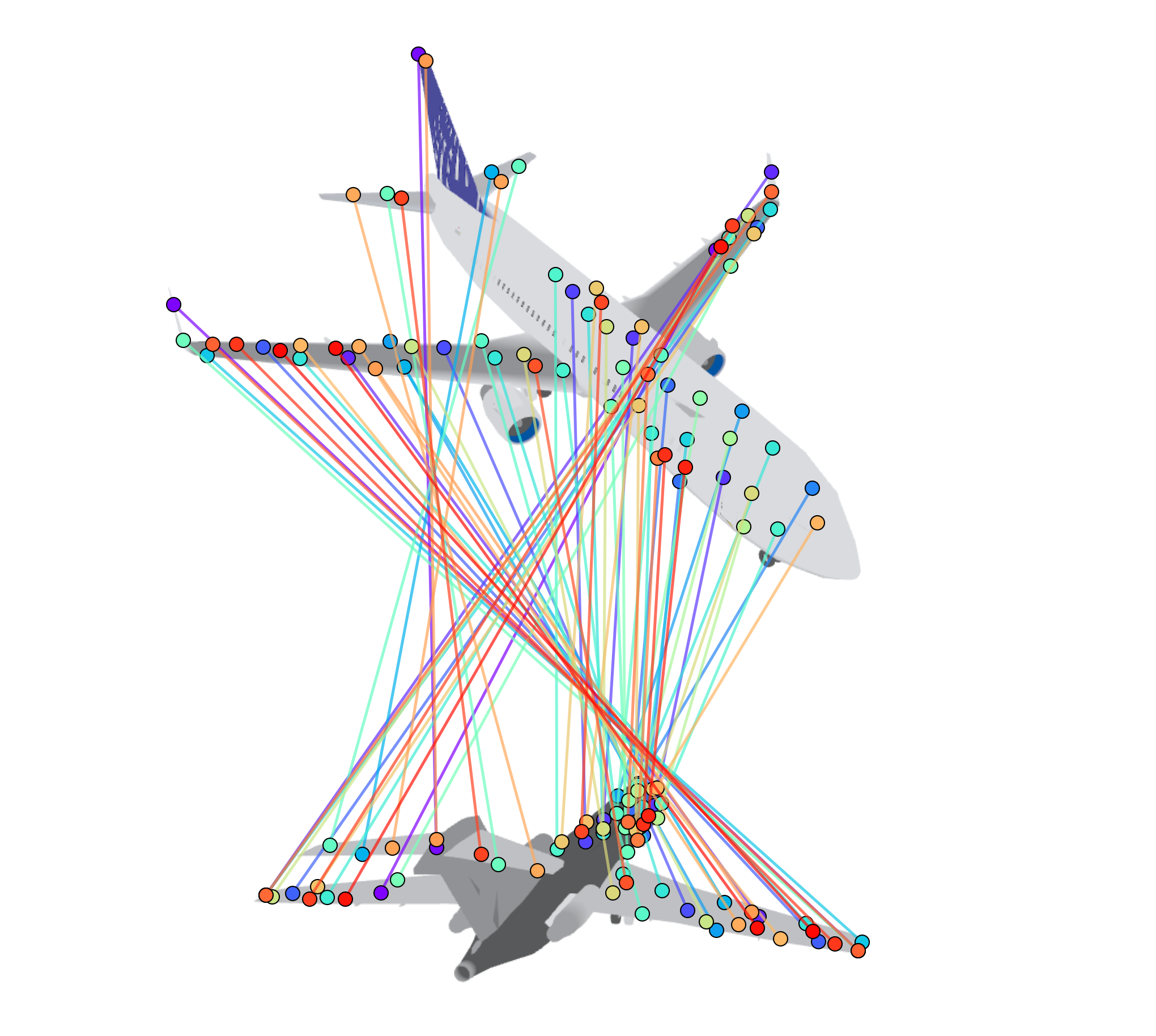}
        \caption{2D--2D}
        \label{fig:corr_2d_2d}
    \end{subfigure}
    \begin{subfigure}[t]{0.30\linewidth}
        \centering
        \includegraphics[
            width=\linewidth,
            trim={75pt 0pt 50pt 0pt},
            clip
        ]{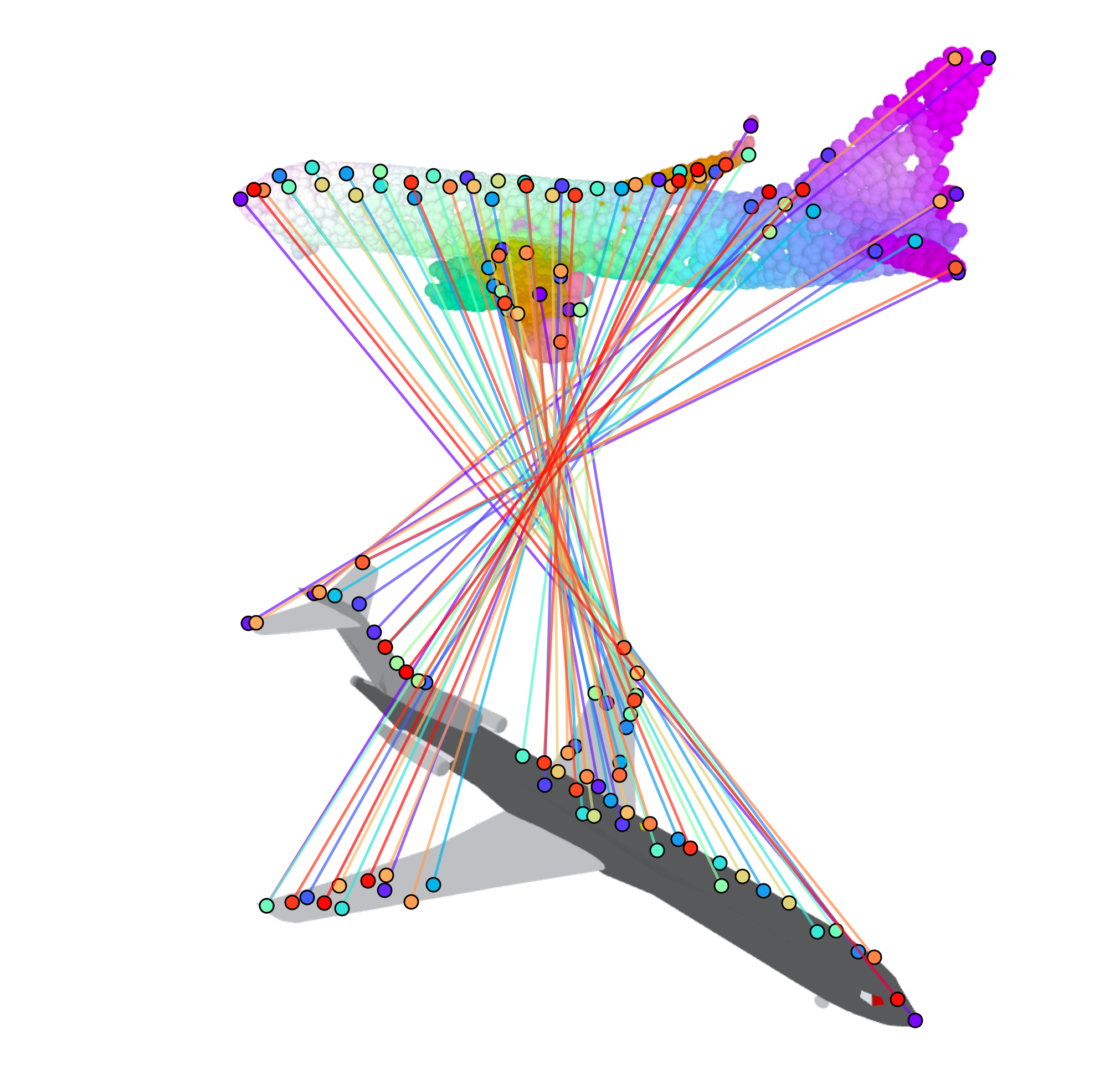}
        \caption{3D--2D}
        \label{fig:corr_3d_2d}
    \end{subfigure}
    \begin{subfigure}[t]{0.33\linewidth}
        \centering
        \includegraphics[
            width=\linewidth,
            trim={50pt 0pt 65pt 0pt},
            clip
        ]{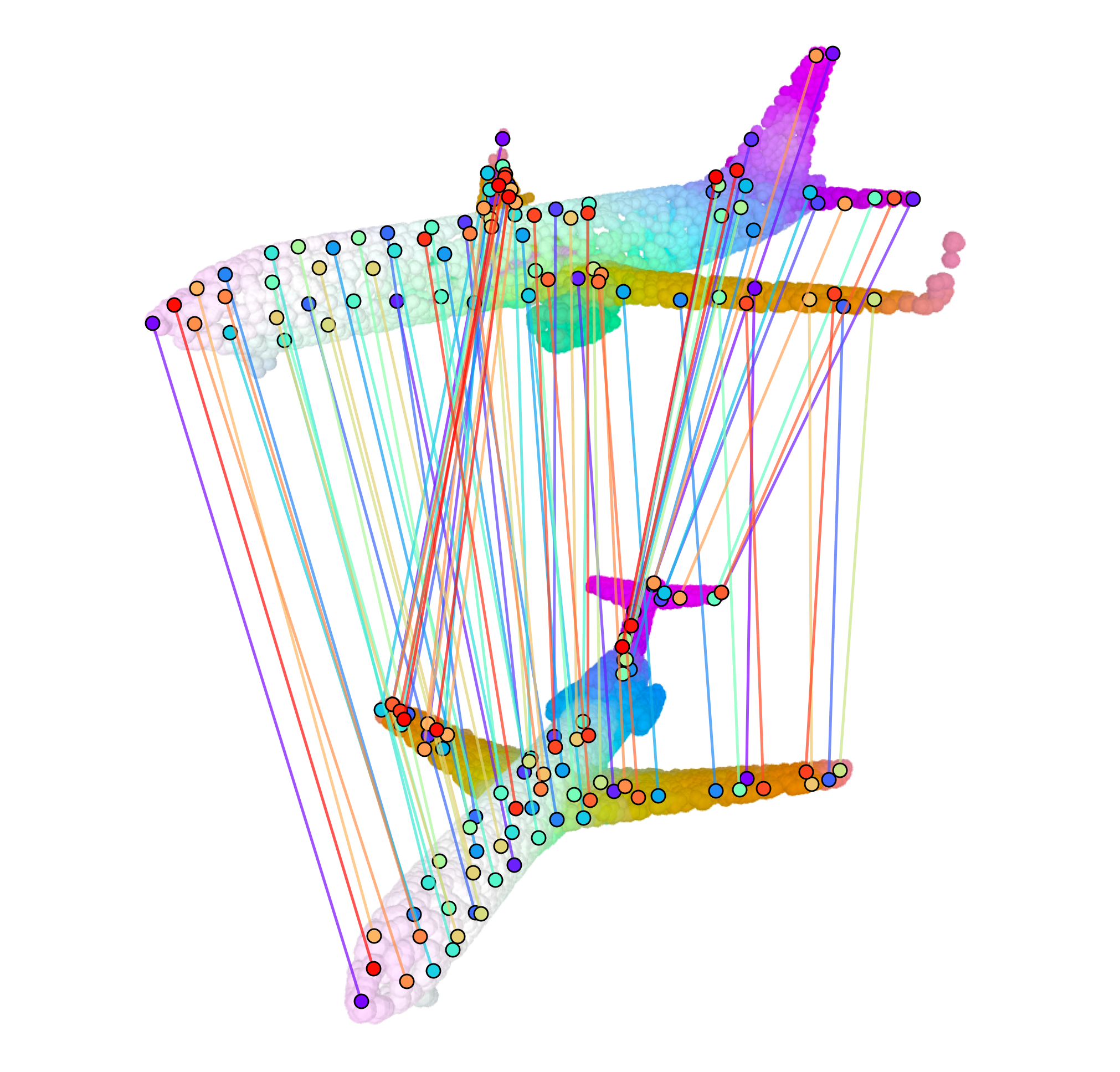}
        \caption{3D--3D}
        \label{fig:corr_3d_3d}
    \end{subfigure}

     \caption{Approximate cross-instance semantic-geometric correspondences between two airplane instances across modalities. Left: 2D--2D correspondences between rendered views. Middle: 3D--2D correspondences between a 3D object and a rendered view. Right: 3D--3D correspondences between object surfaces. Colors in 3D visualize the lifted DINOv2 features reduced to three dimensions.}
    \label{fig:multi_modal_correspondences}
\end{figure}

We introduce SemGeo-Gen, an unsupervised pipeline for generating approximate cross-instance semantic-geometric correspondences from weakly aligned 3D object collections. Given rotation-aligned rigid objects, SemGeo-Gen lifts multi-view DINOv2 features into 3D, uses them to assess semantic compatibility, aligns object instances with continuous piecewise-affine registration, and prunes geometrically plausible but semantically inconsistent matches. The resulting 3D correspondences can be projected into rendered views, yielding 3D--3D, 3D--2D, and 2D--2D supervision without manual annotation; see \autoref{fig:multi_modal_correspondences}. The construction also scales naturally with the number of available assets: a category with \(N\) objects yields up to \(\binom{N}{2}\) cross-instance pairs, and each accepted pair can generate dense 3D correspondences as well as many rendered 2D--2D training pairs.

We evaluate SemGeo-Gen in two complementary ways. First, we use the generated correspondences as synthetic pretraining data for recent semantic correspondence models and then fine-tune them on real SPair-71k annotations~\cite{min2019spair71k}, improving strong recent models without architectural modifications. Second, we compare our generated 3D correspondences to sparse human annotations from KeypointNet~\cite{you2020}, obtaining 93\% PCK@0.10, 97\% PCK@0.15, and 99\% PCK@0.20. These results indicate that automatically generated approximate correspondences, when constructed from 3D geometry and semantic features, can provide scalable supervision for cross-instance correspondence learning.

Our contributions are as follows:
\begin{enumerate}
    \item We introduce SemGeo-Gen, a fully automatic and unsupervised pipeline for generating dense approximate cross-instance semantic-geometric correspondences from weakly aligned 3D object collections, without point, part, or correspondence supervision.

    \item We combine lifted multi-view semantic features, deformable continuous piecewise-affine registration, and semantic correspondence pruning to generate matches constrained by both geometry and semantics.

    \item We project the generated 3D correspondences into rendered views, enabling combinatorially scalable 3D--3D, 3D--2D, and 2D--2D supervision for training image-based correspondence models.

    \item We show that the generated supervision improves recent semantic correspondence models on SPair-71k after real-data fine-tuning, yielding gains of up to 5 PCK points, while also agreeing well with sparse human KeypointNet annotations.
\end{enumerate}

\section{Related Work}
\paragraph{2D and 3D Cross-Instance Correspondence Datasets.}
A wide range of datasets have been proposed for cross-instance correspondence learning, primarily in 2D. Early benchmarks such as PF-PASCAL~\cite{ham2018pascal_proposalflow} and SPair-71k~\cite{min2019spair71k} provide sparse keypoint annotations across object instances under significant viewpoint and scale variation, while CUB~\cite{WahCUB_200_2011} provides fine-grained bird keypoints. AP-10K~\cite{yu2021AP10k}, although originally introduced for animal pose estimation, has been used by GeoAware-SC~\cite{zhang2024tellingLeftFromRight} to derive 2D semantic correspondence supervision from shared pose keypoints. Similarly, MARCO~\cite{cuttano2026marco} derives semantic correspondences from large-scale pose estimation data~\cite{xu2022poseforeverything}, and Jamais Vu~\cite{mariotti2025jamaisvu} extends SPair-71k with SPair-U to evaluate generalization to unseen keypoints. Despite their success, these datasets remain image-based and rely on human-annotated keypoints or pose supervision, providing predominantly sparse correspondence annotations.

In 3D, existing correspondence datasets, including
SHREC~\cite{pratikakis2016shrec_16,dyke2019shrec_19}, parametric animal shape models such as
SMAL~\cite{zuffi2017menagerie}, and deformation benchmarks~\cite{groueix2018b3dcodeddataset}, primarily focus on geometry and often target a single category or articulated shapes such as humans or animals. They also typically lack texture or appearance information, limiting their ability to support projected 2D--2D correspondences for training image-based correspondence models. More recently, DenseMatcher~\cite{zhu2025densematcher} introduced DenseCorr3D, a dataset of colored meshes with part-level annotations across 23 categories, but it does not provide scalable fine-grained point-to-point supervision. KeypointNet~\cite{you2020} provides the largest 3D cross-instance keypoint dataset for rigid objects, but it is limited to 16 categories with sparse annotations, averaging roughly 12 keypoints per object.

Overall, existing datasets rely on expensive and sparse human annotations, limiting scalability. In contrast, our work automatically generates large-scale approximate semantic-geometric correspondences directly between 3D object instances, enabling scalable supervision across diverse rigid object categories and naturally supporting 3D--3D, 3D--2D, and 2D--2D correspondence tasks.

\paragraph{2D Semantic-Geometric Correspondences.}
Semantic correspondence aims to identify analogous locations across different instances of the same category.
However, existing work differs in how it treats the geometric specificity of the correspondence.
A first line of work studies semantic-only correspondence, often formulated as cross-instance image alignment~\cite{gupta2023asic,ofriamar2023neuralCongealing,Barel:ECCV:2024:SpaceJAM,Hirsch:NeurIPS:2025:FastJAM}.
In this setting, geometrically symmetric points may be treated as equivalent if they share the same semantic identity; for example, points on the left and right wings of an airplane may both be considered valid under flipped alignments.
Accordingly, these methods often allow image flips or flipped keypoint annotations during evaluation on datasets such as SPair-71k and CUB~\cite{min2019spair71k,WahCUB_200_2011}.

A second line of work follows the standard semantic correspondence protocol~\cite{hedlin2023unsupervisedstablediffusion,luo2023DiffusionHyperFeaturesTimeSpace,cho2022cats++,mariotti2024sphericalmap,zhang2023taleTwoFeatures,zhang2024tellingLeftFromRight,cuttano2026marco,wandel2025semalign3d,mariotti2025jamaisvu}, typically evaluated on SPair-71k~\cite{min2019spair71k} without annotation flips.
This setting requires models to resolve both semantic identity and geometric orientation when such distinctions are well defined, such as front/back, left/right, or top/bottom.
However, for symmetric categories such as tables, bottles, or potted plants, geometric distinctions can be inherently ambiguous, and evaluation often depends on image-relative location rather than a unique semantic-geometric correspondence.

A third perspective, emphasized by GeoAware-SC~\cite{zhang2024tellingLeftFromRight}, isolates cases of geometry-aware correspondence in which appearance similarity alone is insufficient and resolving the correct match requires reasoning about object orientation or relative part location. Their results show that existing methods often underperform on such cases, motivating our goal of generating richer semantic-geometric supervision at scale. 

\paragraph{3D Semantic Keypoints and Correspondence Methods.}
Recent 3D correspondence methods increasingly combine appearance and geometry to obtain semantic matches across shapes. DenseMatcher~\cite{zhu2025densematcher} learns dense 3D semantic descriptors for textured meshes by projecting multi-view 2D foundation-model features onto mesh vertices, refining them with a 3D network, and computing correspondences with functional maps; its main focus is category-level manipulation from a limited set of annotated objects. UKPGAN~\cite{you2022ukpgan} is an unsupervised 3D keypoint detector that predicts sparse salient keypoints for individual shapes, but does not directly establish cross-instance correspondences. Skeleton Merger~\cite{shi2021skeletonmerger} learns aligned sparse keypoints through skeleton-based reconstruction, yielding category-level keypoint alignment across instances. However, UKPGAN and Skeleton Merger focus on sparse keypoints,
while DenseMatcher relies on costly part-level annotations
and a limited set of annotated objects. None provides scalable,
annotation-free dense cross-instance correspondence supervision
across diverse rigid-object categories.

\section{Method}
\label{sec:method}

\begin{figure*}[t]
    \centering
    \includegraphics[
        width=\textwidth,
        trim={0pt 25pt 0pt 244pt},
        clip
    ]{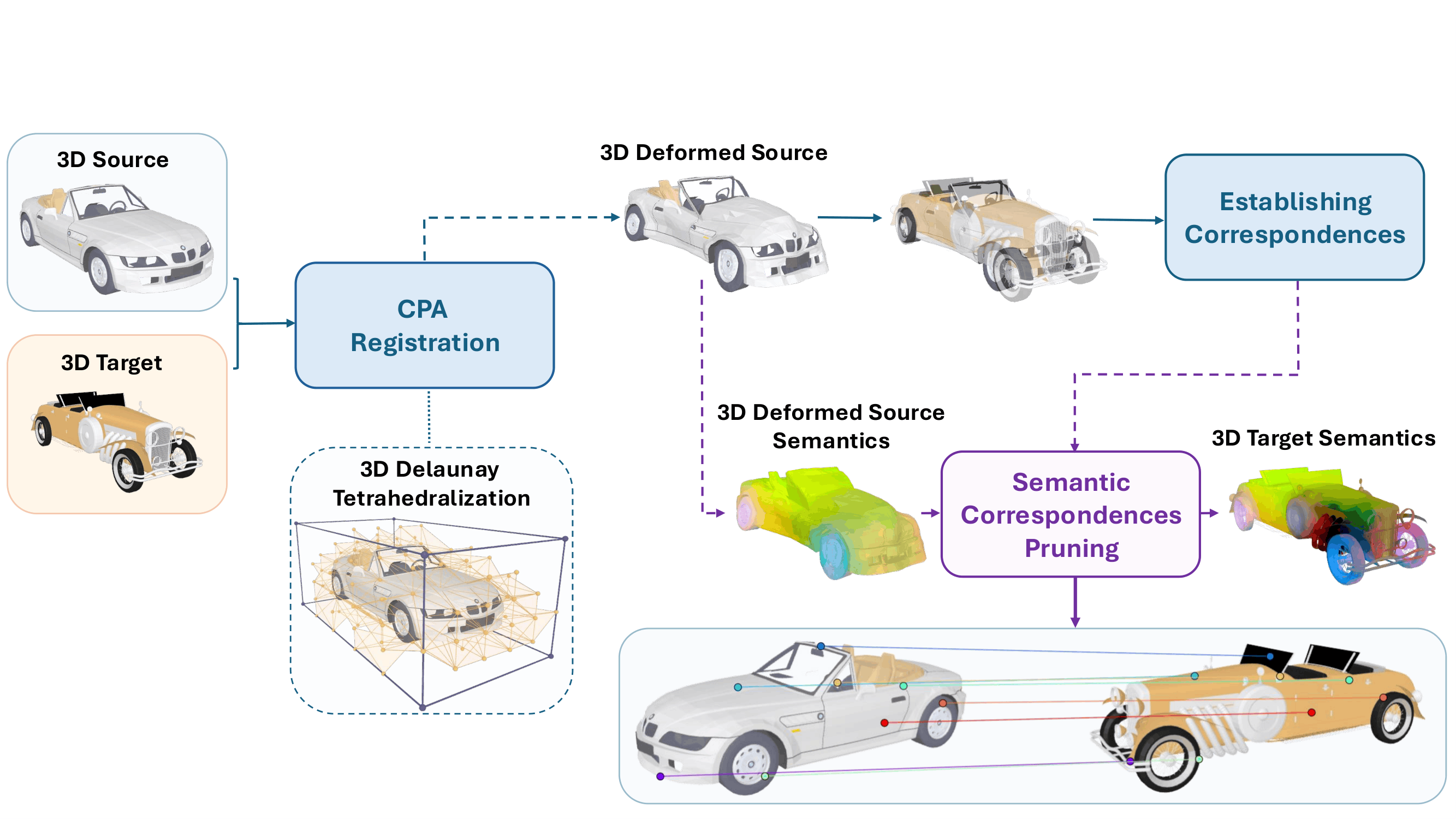}
    \vspace{-4mm}
    \caption{Overview of our correspondence generation pipeline.}
    \label{fig:method_overview}
\end{figure*}

We propose a fully automatic pipeline for generating approximate 3D--3D cross-instance semantic-geometric correspondences between rigid objects, which can be projected into rendered views to obtain 3D--2D and 2D--2D correspondences.
The pipeline's input is a collection of object pairs
\(\mathcal{P}=\{(\mathcal{S}_k,\mathcal{T}_k)\}_{k=1}^{K}\), where each source object \(\mathcal{S}_k\) and target object \(\mathcal{T}_k\) are approximately rotation-aligned. We normalize each object to the unit sphere.
This assumption provides a weak category-level geometric prior, but does not require any point-level, part-level, or correspondence supervision. 

For each object pair, our pipeline proceeds in three main stages.
First, we extract per-point semantic descriptors by rendering each object from multiple views, computing dense DINOv2 features, and lifting and back-projecting them to 3D (\autoref{sec:feature_lifting}).
Second, we estimate a continuous piecewise-affine (CPA) deformation from the source to the target and extract candidate correspondences via nearest-neighbor assignment after alignment.
Third, we prune semantically inconsistent matches, retaining approximate correspondences that are jointly constrained by deformable geometric alignment and multi-view semantic consistency.

In addition, we optionally apply a bidirectional semantic-coverage filter (\autoref{sec:pair_filter}) before registration to reject object pairs with insufficient semantic overlap.
This step reduces the number of retained pairs, but improves the quality of the remaining pairs and the resulting correspondences.
The pipeline is illustrated in \autoref{fig:method_overview}
and summarized in \hyperref[alg:correspondence_generation]{Algorithm~\ref*{alg:correspondence_generation}}.

\subsection{Object Representation and Feature Lifting}
\label{sec:feature_lifting}

Each object \(\mathcal{O}\) is represented as a point cloud with an associated semantic feature map:
\begin{equation}
\label{eq:object_representation}
\mathcal{O}=(\mathcal{X},\mathcal{F}),
\qquad
\mathcal{X}=\{x_i\}_{i=1}^{N}, \; x_i\in\mathbb{R}^{3},
\qquad
\mathcal{F}=\{f_i\}_{i=1}^{N}, \; f_i\in\mathbb{R}^{C}.
\end{equation}
Here, \(x_i\) denotes a 3D point sampled from the object surface, and \(f_i\) denotes its corresponding semantic descriptor.
To obtain these descriptors, we render each object from a set of calibrated virtual cameras, extract dense DINOv2 \cite{oquab2024dinov2} features in image space, and lift them back to the visible 3D points using known camera projection and depth.
We then aggregate multi-view observations by averaging to obtain
a single semantic descriptor for each 3D point.
For a source-target pair \((\mathcal{S},\mathcal{T})\), we write
\begin{equation}
\label{eq:source_target_representation}
\mathcal{S}=(\mathcal{X}_S,\mathcal{F}_S),
\qquad
\mathcal{T}=(\mathcal{X}_T,\mathcal{F}_T),
\end{equation}
where
\(\mathcal{X}_S=\{x_i^S\}_{i=1}^{N_S}\),
\(\mathcal{X}_T=\{x_j^T\}_{j=1}^{N_T}\),
\(\mathcal{F}_S=\{f_i^S\}_{i=1}^{N_S}\), and
\(\mathcal{F}_T=\{f_j^T\}_{j=1}^{N_T}\).

\subsection{Semantic Similarity}
\label{sec:semantic_sim}

We measure semantic compatibility using cosine similarity between
DINOv2~\cite{oquab2024dinov2} descriptors jointly reduced to 64
dimensions via Principal Component Analysis (PCA).
Let $z_i$ denote the reduced descriptor of point $x_i$.
We define
\begin{equation}
\label{eq:semantic_similarity}
s_{\mathrm{sem}}(x_i,x_j)
=
\frac{z_i^\top z_j}{\|z_i\|_2 \|z_j\|_2}.
\end{equation}
Two points are considered semantically compatible if
\begin{equation}
\label{eq:semantic_compatibility}
s_{\mathrm{sem}}(x_i,x_j) > \tau_{\mathrm{sem}},
\end{equation}
where $\tau_{\mathrm{sem}}=0.575$. In our implementation, this corresponds to a squared Euclidean
distance threshold of $0.85$ between L2-normalized descriptors,
since $\|\hat z_i-\hat z_j\|_2^2=2(1-s_{\mathrm{sem}}(x_i,x_j))$,
where $\hat z=z/\|z\|_2$.

\subsection{Optional Semantic Coverage Pair Filtering}
\label{sec:pair_filter}

As an optional preprocessing stage, we apply bidirectional semantic-coverage filtering before registration.

For a source-target pair \((\mathcal{S},\mathcal{T})\), we define the source-to-target semantic coverage as
\begin{equation}
\label{eq:semantic_coverage}
\rho_{\mathrm{sem}}(\mathcal{S} \rightarrow \mathcal{T})
=
\frac{1}{|\mathcal{X}_S|}
\sum_{x_i^S \in \mathcal{X}_S}
\mathbf{1}
\left[
\max_{x_j^T \in \mathcal{X}_T}
s_{\mathrm{sem}}(x_i^S,x_j^T)
>
\tau_{\mathrm{sem}}
\right],
\end{equation}
where \(\mathbf{1}[\cdot]\) denotes the indicator function, which equals \(1\) if the condition inside the brackets is true and \(0\) otherwise. This score measures the fraction of source points that have at least one semantically compatible point in the target.
We also compute the same quantity in the reverse direction, \(\rho_{\mathrm{sem}}(\mathcal{T} \rightarrow \mathcal{S})\), and retain the object pair only if
\begin{equation}
\label{eq:pair_filter}
\rho_{\mathrm{sem}}(\mathcal{S} \rightarrow \mathcal{T}) \geq \gamma
\quad \text{and} \quad
\rho_{\mathrm{sem}}(\mathcal{T} \rightarrow \mathcal{S}) \geq \gamma .
\end{equation}
In our experiments, we use \(\gamma = 0.6\).

\subsection{Continuous Piecewise-Affine Registration}
\label{sec:registration_cpa}

For each retained object pair \((\mathcal{S},\mathcal{T})\), we estimate a non-rigid transformation that aligns the source point set \(\mathcal{X}_S\) to the target \(\mathcal{X}_T\).
We model this transformation using a continuous piecewise-affine (CPA) mapping
\begin{equation}
\label{eq:cpa_transform}
T_{\mathrm{CPA}}:\mathbb{R}^{3}\rightarrow\mathbb{R}^{3},
\end{equation}
which produces a continuous, spatially coherent deformation field while allowing local shape variation across object instances.
This transformation is parameterized by a regular control grid over the normalized 3D domain, which is decomposed into tetrahedral cells using Delaunay triangulation.
Each grid vertex is associated with an optimized displacement vector, and we obtain the displacement of an arbitrary source point by barycentric interpolation from the vertices of the tetrahedron containing it.
This yields a deformation field that is locally affine within each tetrahedron and continuous across neighboring cells. Unlike CPAB transformations~\cite{Freifeld:TPAMI:2017:CPAB,Aflalo:NeurIPS:2026:FG-CPAB}, which are obtained by integrating CPA velocity fields, here we use CPA mappings directly as transformations.

We estimate the transformation by optimizing a Chamfer distance loss~\cite{fan2017chamfer} with smoothness regularization via gradient descent:
\begin{equation}
\label{eq:cpa_objective}
\mathcal{L}_{\mathrm{CPA}}
=
\mathcal{L}_{\mathrm{Chamfer}}
\left(
T_{\mathrm{CPA}}(\mathcal{X}_S),
\mathcal{X}_T
\right)
+
\lambda_{\mathrm{smooth}}
\mathcal{R}(T_{\mathrm{CPA}}),
\end{equation}
where \(\lambda_{\mathrm{smooth}} = 0.3\) in all experiments, and 
\begin{equation}
\label{eq:chamfer_loss}
\mathcal{L}_{\mathrm{Chamfer}}
(\widehat{\mathcal{X}}_S,\mathcal{X}_T)
=
\frac{1}{|\widehat{\mathcal{X}}_S|}
\sum_{\hat{x}_i^S \in \widehat{\mathcal{X}}_S}
\min_{x_j^T \in \mathcal{X}_T}
\left\|
\hat{x}_i^S-x_j^T
\right\|_2^2
+
\frac{1}{|\mathcal{X}_T|}
\sum_{x_j^T \in \mathcal{X}_T}
\min_{\hat{x}_i^S \in \widehat{\mathcal{X}}_S}
\left\|
x_j^T-\hat{x}_i^S
\right\|_2^2 .
\end{equation}

The first term aligns the deformed source to the target, while the second term encourages the target surface to be covered by the deformed source.

The smoothness term regularizes neighboring control-grid displacements:
\begin{equation}
\label{eq:cpa_smoothness}
\mathcal{R}(T_{\mathrm{CPA}})
=
\frac{1}{|\mathcal{E}|}
\sum_{(v_a,v_b)\in\mathcal{E}}
\left\|
u_{v_a}-u_{v_b}
\right\|_2^2,
\end{equation}
where \(\mathcal{E}\) denotes edges between neighboring vertices
in the tetrahedral control grid, and \(u_v\in\mathbb{R}^3\)
denotes the displacement vector associated with vertex \(v\).
This regularization promotes locally coherent deformations while
still allowing non-rigid cross-instance variation.
After optimization, the transformed source point cloud is given by
\begin{equation}
\label{eq:deformed_source}
\widehat{\mathcal{X}}_S
=
T_{\mathrm{CPA}}(\mathcal{X}_S)
=
\{\hat{x}_i^S\}_{i=1}^{N_S}.
\end{equation}

\begin{algorithm}[t]
\DontPrintSemicolon
\caption{Automatic Approximate Semantic-Geometric Correspondence Generation}
\label{alg:correspondence_generation}

\KwIn{
Rotation-aligned and scale-normalized object pairs 
\(\mathcal{P}=\{(\mathcal{S}_k,\mathcal{T}_k)\}_{k=1}^{K}\);
lifted 3D DINOv2 features;
semantic similarity threshold \(\tau_{\mathrm{sem}}\);
optional coverage threshold \(\gamma\).
}
\KwOut{
Approximate semantic-geometric correspondences.
}

\ForEach{\((\mathcal{S},\mathcal{T}) \in \mathcal{P}\)}{

    Let \((\mathcal{X}_S,\mathcal{F}_S)\) and \((\mathcal{X}_T,\mathcal{F}_T)\) be the source and target representations\;

    \If{semantic coverage filtering is enabled}{
        \If{\(\rho_{\mathrm{sem}}(\mathcal{S}\!\rightarrow\!\mathcal{T}) < \gamma\) 
        \textbf{or} 
        \(\rho_{\mathrm{sem}}(\mathcal{T}\!\rightarrow\!\mathcal{S}) < \gamma\)}{
            \textbf{continue}\;
        }
    }

    Estimate \(T_{\mathrm{CPA}}\) by minimizing Eq.~\eqref{eq:cpa_objective}\;

    Compute the deformed source 
    \(\widehat{\mathcal{X}}_S = T_{\mathrm{CPA}}(\mathcal{X}_S)\)\;

    Assign each \(\hat{x}_i^S \in \widehat{\mathcal{X}}_S\) to its nearest neighbor in \(\mathcal{X}_T\) using Eq.~\eqref{eq:nearest_neighbor_assignment}\;

    Construct candidate correspondences \(\mathcal{C}_{\mathcal{S},\mathcal{T}}^{0}\)\;

    Prune correspondences using the semantic criterion in Eq.~\eqref{eq:semantic_pruning_condition}\;

    Store the final correspondence set \(\mathcal{C}_{\mathcal{S},\mathcal{T}}\)\;
}

\textbf{Return} all processed object pairs that were not filtered out and their correspondence sets.

\end{algorithm}

\subsection{Approximate Correspondence Extraction}
\label{sec:establish_correspondences}

After CPA registration, the source and target are approximately aligned geometrically.
We then assign each transformed source point to its nearest neighbor in the target point cloud.
For each \(\hat{x}_i^S \in \widehat{\mathcal{X}}_S\), the target index is
\begin{equation}
\label{eq:nearest_neighbor_assignment}
a(i)
=
\arg\min_{j\in\{1,\ldots,N_T\}}
\left\|
\hat{x}_i^S-x_j^T
\right\|_2 .
\end{equation}
This produces an initial set of candidate correspondences:
\begin{equation}
\label{eq:initial_correspondences}
\mathcal{C}_{\mathcal{S},\mathcal{T}}^{0}
=
\left\{
\left(x_i^S,x_{a(i)}^T\right)
\; \middle| \;
i=1,\ldots,N_S
\right\}.
\end{equation}
Importantly, the assignment is computed using the deformed source point \(\hat{x}_i^S\), while the stored correspondence is defined between the original source point \(x_i^S\) and the target point \(x_{a(i)}^T\).
These matches should therefore be interpreted as approximate cross-instance correspondences induced by deformation-based alignment, rather than exact point-level annotation.

\subsection{Semantic Consistency Pruning}
\label{sec:semantic_pruning}

Geometric proximity alone may produce implausible matches in close regions with different semantics or regions with no meaningful semantic counterpart.
We therefore apply a final semantic pruning step to retain only candidate matches that are also semantically compatible.
A candidate correspondence \((x_i^S,x_{a(i)}^T)\in\mathcal{C}_{\mathcal{S},\mathcal{T}}^{0}\) is retained only if
\begin{equation}
\label{eq:semantic_pruning_condition}
s_{\mathrm{sem}}
\left(
x_i^S,
x_{a(i)}^T
\right)
>
\tau_{\mathrm{sem}}.
\end{equation}
The final semantic-geometric correspondence set is
\begin{equation}
\label{eq:final_correspondences}
\mathcal{C}_{\mathcal{S},\mathcal{T}}
=
\left\{
\left(x_i^S,x_{a(i)}^T\right)
\in
\mathcal{C}_{\mathcal{S},\mathcal{T}}^{0}
\; \middle| \;
s_{\mathrm{sem}}
\left(
x_i^S,x_{a(i)}^T
\right)
>
\tau_{\mathrm{sem}}
\right\}.
\end{equation}
This step removes matches that are geometrically nearby after deformation but semantically inconsistent according to the lifted DINOv2 \cite{oquab2024dinov2} features.
The remaining set forms our final approximate semantic-geometric correspondences.

\subsection{Projection to 2D Correspondences}
\label{sec:projection_2d}

The generated correspondences, established in 3D, can be directly projected into rendered views to produce aligned 3D--3D, 3D--2D, or 2D--2D supervision. Given calibrated rendering views with camera projections \(\pi_{c_S}\) and \(\pi_{c_T}\), a 3D correspondence \((x_i^S,x_j^T)\in\mathcal{C}_{\mathcal{S},\mathcal{T}}\) induces the 2D projections
\begin{equation}
\label{eq:projection_to_2d}
(u_i^S,u_j^T)
=
\left(
\pi_{c_S}(x_i^S),
\pi_{c_T}(x_j^T)
\right),
\end{equation}
where \(c_S\) and \(c_T\) denote the selected source and target camera views.
Projecting one endpoint yields 3D--2D supervision, while projecting both endpoints yields 2D--2D supervision.
In either case, we retain only correspondences whose projected endpoints are visible in their respective rendered views.
By rendering multiple views of each object pair, we obtain large-scale automatically generated training data for image-based semantic correspondence models.

\section{Experiments}\label{sec:experiments}

We evaluate the proposed correspondence-generation pipeline in two complementary settings. First, we assess whether the generated correspondences can improve real-image semantic correspondence models when used as synthetic pretraining data (\autoref{spair71k_eval}). Second, we directly evaluate the quality of the generated 3D correspondences against sparse human-annotated 3D ground truth(\autoref{annotation_quality}).
All experiments were conducted on a single NVIDIA RTX A6000 GPU.

\subsection{Experimental Data}
\label{sec:exp_data}

Our method assumes collections of rigid 3D objects that are approximately rotation-aligned, providing a weak category-level geometric prior. In this work, we use two complementary 3D sources, 
\textbf{KeypointNet}~\cite{you2020} and \textbf{Objaverse-OA}~\cite{lu2025orientationmatters}, 
to validate the proposed pipeline. 
\textbf{KeypointNet} provides sparse human-annotated semantic keypoints on approximately aligned 3D object models. It has relatively few rigid categories, but many objects per category, making it suitable for evaluating how well the generated correspondences  agree with human semantic annotations. These annotations are used only for evaluation, not for training or correspondence generation.
\textbf{Objaverse-OA} provides far broader category and shape diversity, containing over 100 rigid categories after curation, but without correspondence annotations. \autoref{sup:qualitative}  shows example correspondences generated by our method on several Objaverse-OA categories beyond those covered by existing correspondence datasets.

For downstream SPair-71k~\cite{min2019spair71k} experiments,
we generate synthetic correspondences from nine rigid categories
(aeroplane; bicycle; boat; bus; car; chair; motorbike; train; tvmonitor), excluding non-rigid and strongly symmetric objects.
We evaluate two settings: fine-tuning on the same Rigid-9 categories,
and fine-tuning on all 18 categories, additionally including
Non-rigid-7 (bird; cat; cow; dog; horse; person; sheep)
and Symmetric-2 (bottle; potted plant).
Both settings use the same Rigid-9 synthetic pretraining data.

\begin{table*}[t]
\captionsetup{justification=centering}
\small
\setlength{\tabcolsep}{2.6pt}
\renewcommand{\arraystretch}{0.95}
\caption{Category-wise PCK@0.10 results on selected SPair-71k categories with well-defined semantic-geometric correspondences. P.T. denotes pretraining.}
\label{tab:spair_pck010}
\resizebox{\textwidth}{!}{
\begin{tabular}{llccccccccc|c}
\toprule
Method & Training regime & aero & bike & boat & bus & car & chair & mbike & train & tv & Mean \\
\midrule
\multicolumn{12}{c}{\textit{Zero-shot / weakly supervised methods}} \\
\midrule
SD \cite{rombach2022stablediffusion,zhang2023taleTwoFeatures} & Zero-shot & 62.8 & 52.7 & 31.2 & 39.1 & 35.6 & 32.0 & 51.0 & 61.8 & 52.9 & 46.57 \\
DINOv2 \cite{oquab2024dinov2,zhang2023taleTwoFeatures} & Zero-shot & 73.4 & 60.2 & 43.2 & 46.7 & 45.1 & 33.4 & 60.7 & 54.2 & 23.9 & 48.98 \\
DINOv2+SD \cite{zhang2023taleTwoFeatures} & Zero-shot & 73.8 & 61.0 & 40.2 & 47.4 & 44.1 & 41.5 & 61.7 & 63.5 & 52.4 & 53.96 \\
SphericalMaps \cite{mariotti2024sphericalmap} & Weakly supervised (SPair-71k) & 76.2 & 60.1 & 46.5 & 74.9 & 68.0 & 45.1 & 69.1 & 73.9 & 58.1 & 63.54 \\
\midrule
\multicolumn{12}{c}{\textit{Supervised methods}} \\
\midrule
SCorrSan \cite{huang2022SCorrSan} & SPair-71k & 57.1 & 40.3 & 38.1 & 57.8 & 47.1 & 25.2 & 45.3 & 77.7 & 69.7 & 50.92 \\
CATS++ \cite{cho2022cats++}  & SPair-71k & 60.6 & 46.9 & 41.6 & 64.9 & 50.4 & 29.2 & 50.9 & 80.9 & 74.9 & 55.59 \\
DHF \cite{luo2023DiffusionHyperFeaturesTimeSpace} & SPair-71k & 74.0 & 61.0 & 40.7 & 70.0 & 74.4 & 38.5 & 66.6 & 87.4 & 60.3 & 63.66 \\
DINO+SD (S) \cite{zhang2023taleTwoFeatures} & SPair-71k & 84.7 & 67.5 & 64.5 & 85.7 & 82.0 & 57.0 & 75.9 & 93.6 & 70.5 & 75.71 \\
Jamais Vu \cite{mariotti2025jamaisvu} & SPair-71k 
& 83.6 & 70.1 & 64.1 & 91.4 & 83.2 & 61.4 & 78.4 & 95.5 & 80.1 & 78.64 \\
\midrule
\multicolumn{12}{c}{\textit{Recent supervised state-of-the-art methods under different supervision settings}} \\
\midrule
\multirow{3}{*}{GeoAware-SC \cite{zhang2024tellingLeftFromRight}}
& SPair-71k & 83.42 & 71.07 & 66.68 & 90.93 & 86.10 & 62.18 & 79.54 & 95.33 & 79.00 & 79.36 \\
& SPair-71k + Ours & 86.28 & 74.60 & 68.33 & 93.12 & 84.73 & 68.94 & 79.73 & 95.31 & 80.73 & 81.31 \\
& Ours P.T. $\rightarrow$ SPair-71k & \textbf{89.38} & \textbf{76.11} & \textbf{71.15} & \textbf{93.80} & \textbf{86.74} & \textbf{73.80} & \textbf{82.33} & \textbf{95.67} & \textbf{84.59} & \textbf{83.73} \\
\midrule
\multirow{3}{*}{MARCO \cite{cuttano2026marco}}
& SPair-71k 
& 92.16 & 76.76 & 76.53 & 93.17 & 88.57 & 82.49 & 82.03 & 93.62 & 91.72 & 86.34 \\
& SPair-71k + Ours 
& 92.26 & 77.49 & 73.74 & 92.59 & 84.86 & 82.71 & 79.78 & 93.22 & 89.72 & 85.15 \\
& Ours P.T. $\rightarrow$ SPair-71k 
& \textbf{95.59} & \textbf{80.66} & \textbf{79.35} & \textbf{94.59} & \textbf{91.89} & \textbf{91.16} & \textbf{82.30} & \textbf{94.69} & \textbf{93.49} & \textbf{89.30} \\
\bottomrule
\end{tabular}
}
\end{table*}

\subsection{Results on SPair-71k}
\label{spair71k_eval}
We evaluate the synthetic 2D correspondence supervision generated by our pipeline on SPair-71k~\cite{min2019spair71k}, a standard benchmark for cross-instance semantic correspondence.
Concretely, we generate 12K synthetic 2D training pairs per category from diverse 3D object instances sampled from Objaverse-OA~\cite{lu2025orientationmatters}. To preserve category and shape diversity in the synthetic pretraining data, we avoid pair-level semantic filtering in this experiment. We focus on 9 rigid categories with well-defined 3D semantic-geometric structure: aeroplane; bicycle; boat; bus; car; chair; motorbike; train; tvmonitor. We report reproduced results for recent methods, including GeoAware-SC~\cite{zhang2024tellingLeftFromRight}, Jamais Vu~\cite{mariotti2025jamaisvu}, and MARCO~\cite{cuttano2026marco}, alongside reported results for earlier methods in \autoref{tab:spair_pck010}.
For the state-of-the-art GeoAware-SC and MARCO, we also evaluate the effect of our generated data in three training regimes: training on SPair-71k only; joint training with our projected correspondences; pretraining on our generated correspondences followed by fine-tuning on SPair-71k. 

Since our data is generated from synthetic 3D objects, a synthetic-to-real domain gap remains,
and naive joint training gives mixed results. In contrast, the standard pretrain-and-finetune
regime consistently improves both evaluated models. GeoAware-SC~\cite{zhang2024tellingLeftFromRight}
improves from 79.36 to 83.73 mean PCK (+4.37), and MARCO~\cite{cuttano2026marco}
improves from 86.34 to 89.30 (+2.96). These gains are obtained without any architectural
modification, highlighting that even approximate semantic-geometric correspondences provide
a useful geometric-semantic training signal that is easier to obtain than manual annotations.
Additional qualitative examples in \autoref{fig:qual_marco_main} further illustrate the
improved semantic-geometric consistency obtained after pretraining on our generated data.

\noindent\textbf{Beyond rigid categories.}
We also investigate whether synthetic pretraining on rigid
categories transfers to non-rigid and symmetric objects.
As \autoref{tab:spair_all18} shows, pretraining GeoAware-SC
on our Rigid-9 synthetic correspondences followed by
fine-tuning on all 18 SPair-71k categories improves
PCK@0.10 by 3.72 on rigid categories, 1.29 on non-rigid
categories, and 2.39 overall, compared to real-only training.
Notably, these gains extend to categories excluded
from synthetic pretraining.
\begin{table}[h!]
    \centering
    \captionsetup{justification=centering}
    \small
    \setlength{\tabcolsep}{3pt}
    \renewcommand{\arraystretch}{0.95}

    \caption{
        \textbf{Generalization across all SPair-71k categories.}
        GeoAware-SC pretrained on Rigid-9 synthetic correspondences
        and fine-tuned on all 18 SPair-71k categories.
        PCK@0.10 (\%); improvements over real-only
        training are shown in parentheses.
    }
    \label{tab:spair_all18}

    \vspace{4pt}

    \centering
    \begin{tabular}{lcccc}
        \toprule
        \textbf{Training regime} &
        Rigid-9 &
        Non-rigid-7 &
        Sym.-2 &
        All-18 \\
        \midrule

        Real only
        & 80.48 & 86.94 & 72.35 & 82.09 \\

        DenseMatcher \cite{zhu2025densematcher} P.T. $\rightarrow$ Real
        & 80.37 & 86.54 & 71.97 & 81.83 \\

        \textbf{Ours P.T. $\rightarrow$ Real}
        & \textbf{84.20} & \textbf{88.23}
        & \textbf{72.58} & \textbf{84.48} \\
        & \textbf{(+3.72)} & \textbf{(+1.29)}
        & \textbf{(+0.23)} & \textbf{(+2.39)} \\

        \bottomrule
    \end{tabular}
\end{table}

We additionally compare against DenseMatcher~\cite{zhu2025densematcher}
under the same training protocol, which performs similarly to
real-only training (81.83 vs.\ 82.09 PCK overall).
DenseMatcher combines semantic features with geometric refinement
but requires costly part annotations, with only 5 animals, 5 chairs,
and 6 vehicles in its released training data.
Its functional-map matching may also struggle with large structural
variations, consistent with the symmetry-mismatched and structurally
inconsistent correspondences we observed on diverse Objaverse-OA instances.
In contrast, SemGeo-Gen generates dense approximate semantic-geometric
correspondences from scalable collections of rotation-aligned 3D models,
without part or correspondence annotations.

\subsection{Evaluation of Annotation Quality}
\label{annotation_quality}

We evaluate our generated correspondences on KeypointNet~\cite{you2020},
which provides human-annotated semantic keypoints and correspondences
on approximately rotation-aligned 3D objects.
We assess both \textbf{3D--3D} correspondence accuracy against human
annotations and \textbf{3D--2D} accuracy after projection into image space.

\begin{figure*}[t]
\centering

\sbox{\kpnetbox}{%
\begin{minipage}[t]{0.48\textwidth}
\captionsetup{type=table}
\caption{Evaluation of annotation quality on KeypointNet \cite{you2020} correspondences.}
\label{tab:keypointnet_table}
\centering
\footnotesize
\setlength{\tabcolsep}{3.5pt}
\renewcommand{\arraystretch}{1.0}
\resizebox{\linewidth}{!}{%
\begin{tabular}{lcccc}
    \toprule
    Category & PCK@.05 $\uparrow$ & PCK@.10 $\uparrow$ & PCK@.15 $\uparrow$ & PCK@.20 $\uparrow$ \\
    \midrule
    Airplane   & 0.79 & 0.94 & 0.98 & 0.99 \\
    Bathtub    & 0.70 & 0.88 & 0.96 & 0.99 \\
    Bed        & 0.68 & 0.91 & 0.96 & 0.99 \\
    Bottle     & 0.83 & 0.96 & 0.98 & 0.99 \\
    Cap        & 0.80 & 0.97 & 1.0 & 1.00 \\
    Car        & 0.78 & 0.95 & 0.99 & 1.00 \\
    Chair      & 0.77 & 0.96 & 0.99 & 0.99 \\
    Guitar     & 0.81 & 0.96 & 0.98 & 0.99 \\
    Helmet     & 0.55 & 0.84 & 0.94 & 0.98 \\
    Knife      & 0.83 & 0.93 & 0.96 & 0.98 \\
    Laptop     & 0.89 & 0.99 & 1.00 & 1.00 \\
    Motorcycle & 0.75 & 0.93 & 0.98 & 0.99 \\
    Mug        & 0.73 & 0.96 & 0.99 & 1.00 \\
    Skateboard & 0.86 & 0.97 & 0.99 & 1.00 \\
    Table      & 0.76 & 0.96 & 0.99 & 1.00 \\
    Vessel     & 0.45 & 0.71 & 0.87 & 0.94 \\
    \midrule
    Mean       & 0.75 & 0.93 & 0.97 & 0.99 \\
    Median     & 0.78 & 0.96 & 0.98 & 0.99 \\
    \bottomrule
\end{tabular}%
}
\end{minipage}%
}
\setlength{\kpnetheight}{\dimexpr\ht\kpnetbox+\dp\kpnetbox\relax}

\setlength{\kpimgheight}{\dimexpr(\kpnetheight - 3\baselineskip)/3\relax}

\makebox[\textwidth][c]{%
\usebox{\kpnetbox}%
\hfill%
\begin{minipage}[t][\kpnetheight][t]{0.504\textwidth}
\captionof{figure}{Qualitative SPair-71k results for MARCO \cite{cuttano2026marco}. Left: SPair-71k-only. Right: ours-pretrained $\rightarrow$ SPair-71k fine-tuned.}
\label{fig:qual_marco_main}
\centering
\scriptsize
\setlength{\tabcolsep}{0pt}

\begin{tabular}{cc}
\textbf{SPair-71k only} & \textbf{Ours P.T.$\rightarrow$SPair-71k} \\[0.2mm]

\includegraphics[width=0.49\linewidth]{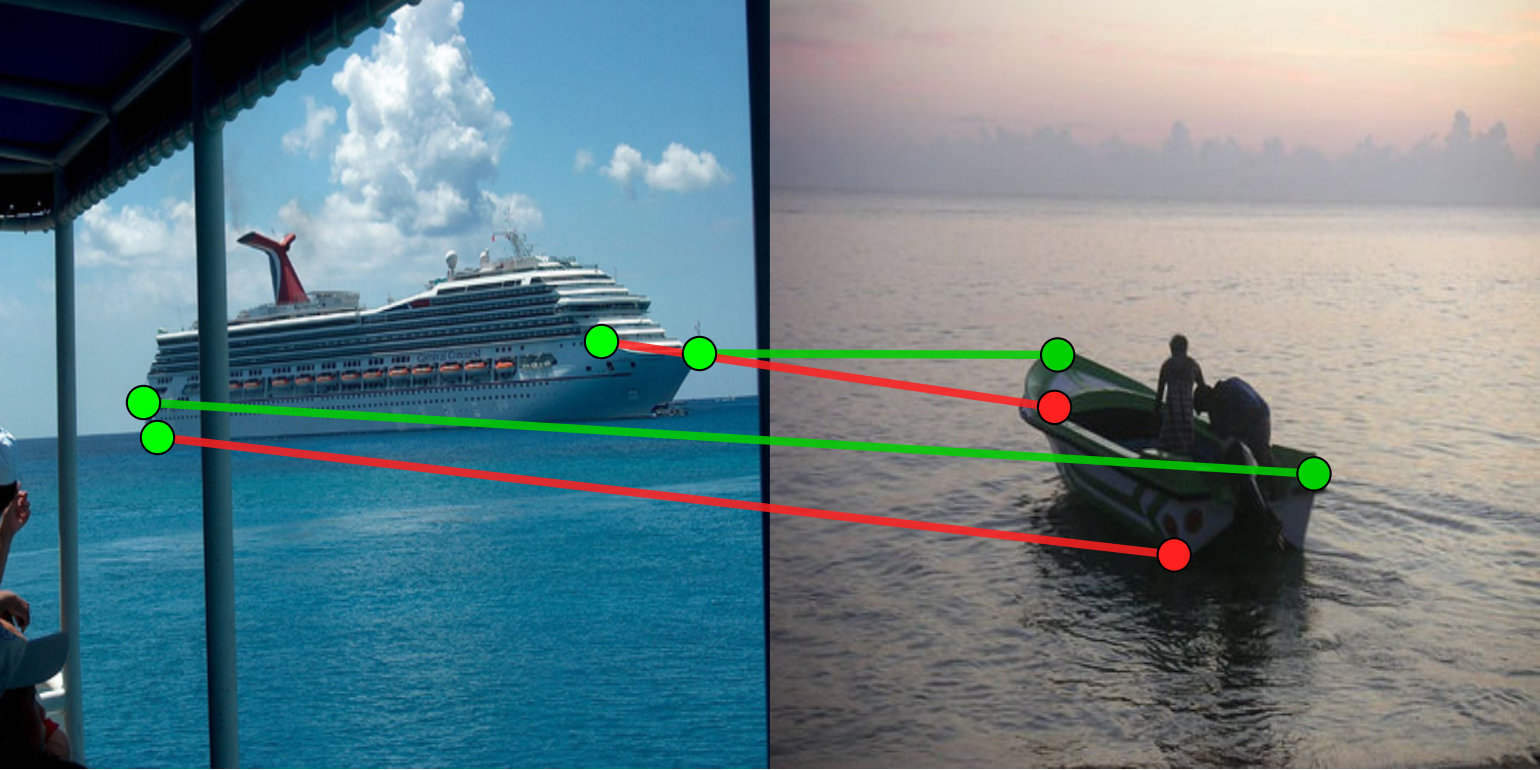} &
\includegraphics[width=0.49\linewidth]{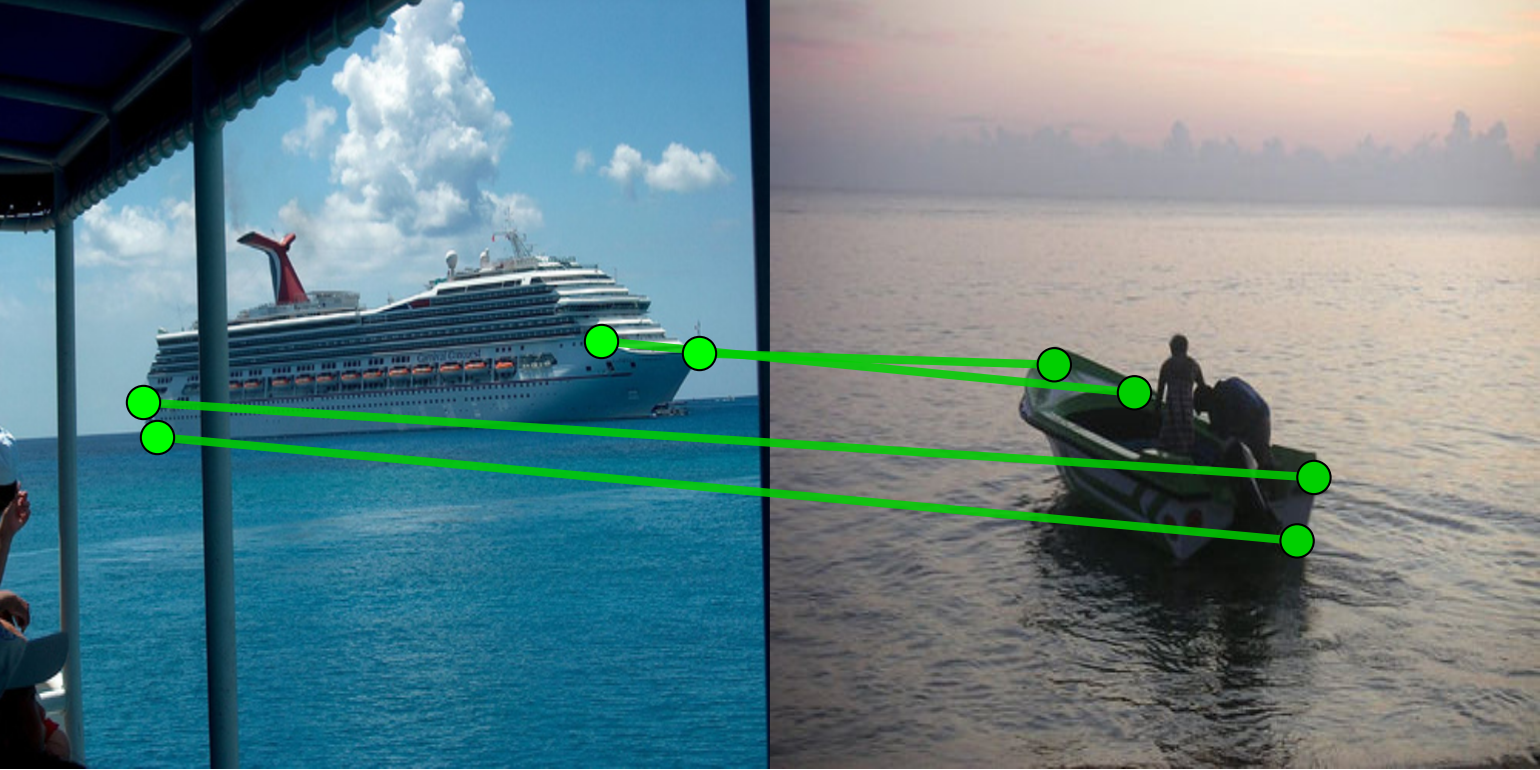} \\[-1mm]

\includegraphics[width=0.49\linewidth]{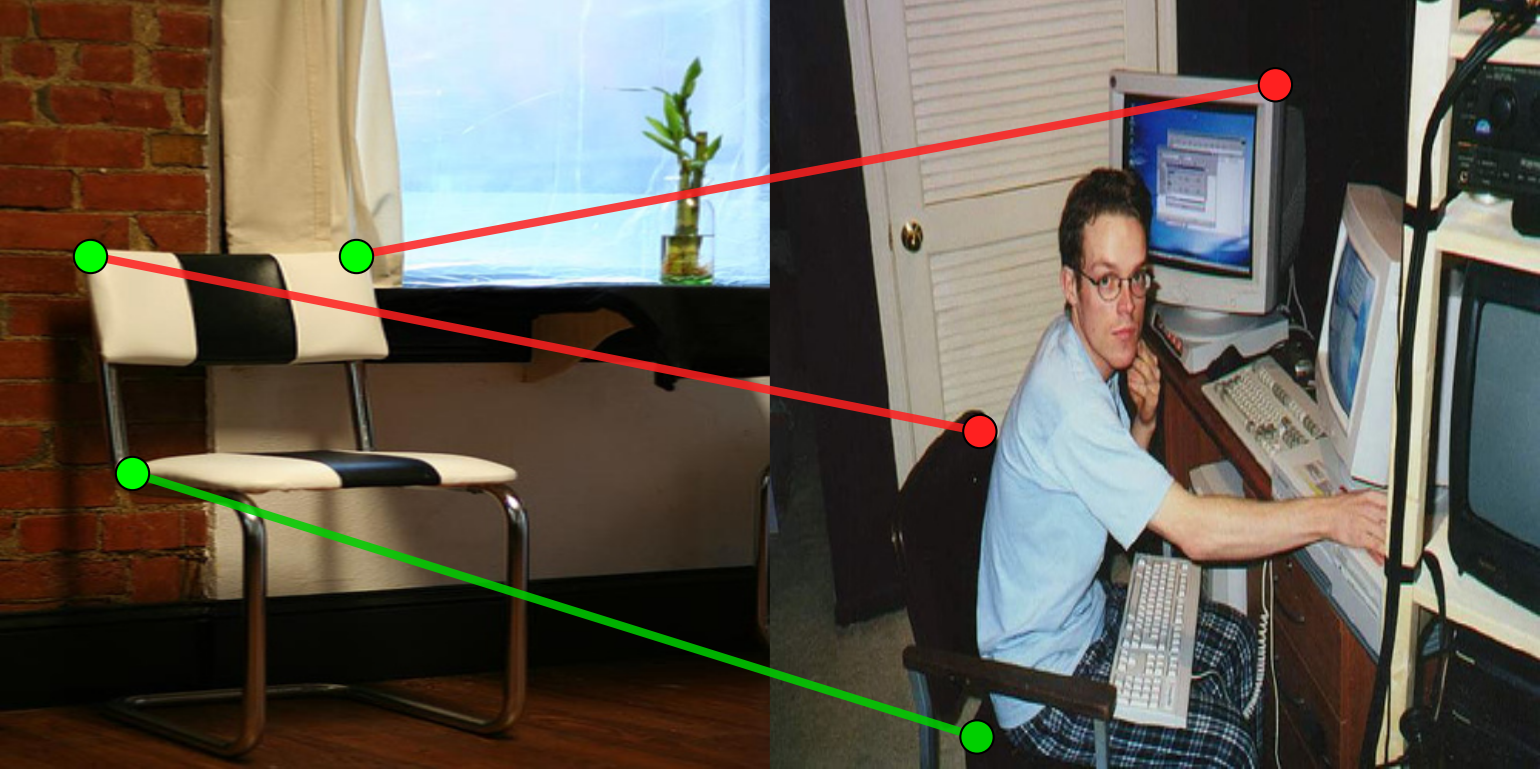} &
\includegraphics[width=0.49\linewidth]{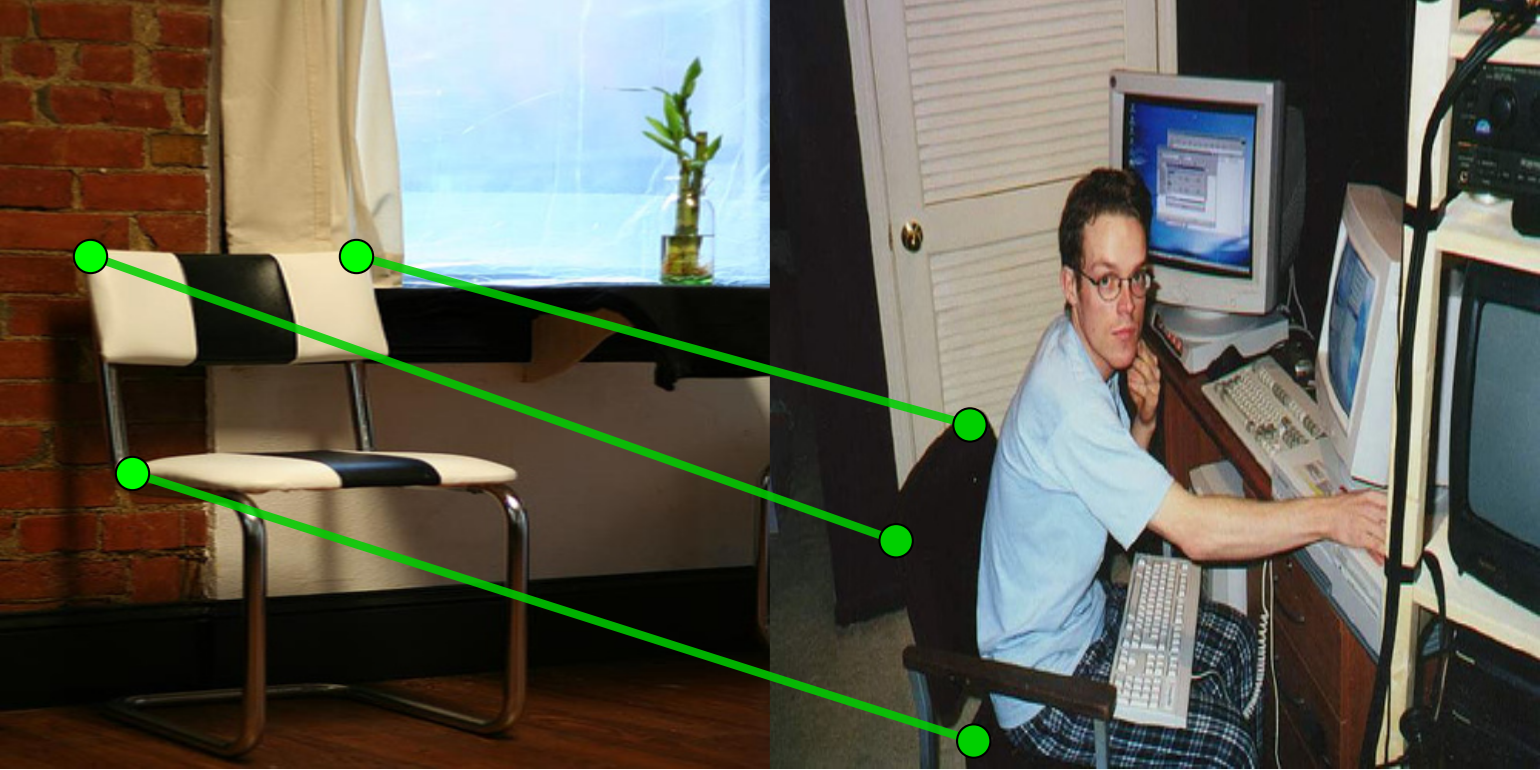}
 \\[-1mm]
\includegraphics[width=0.49\linewidth]{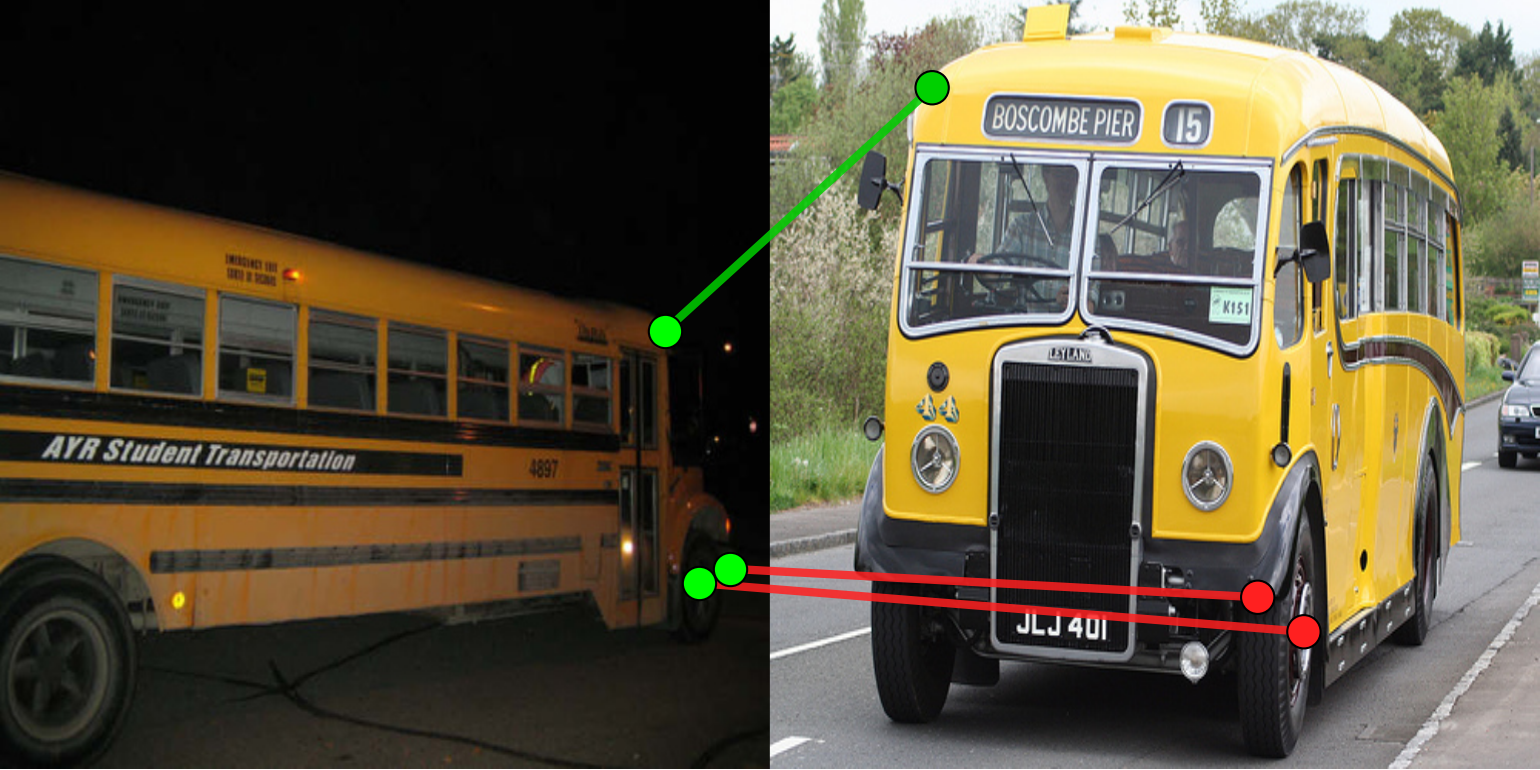} &
\includegraphics[width=0.49\linewidth]{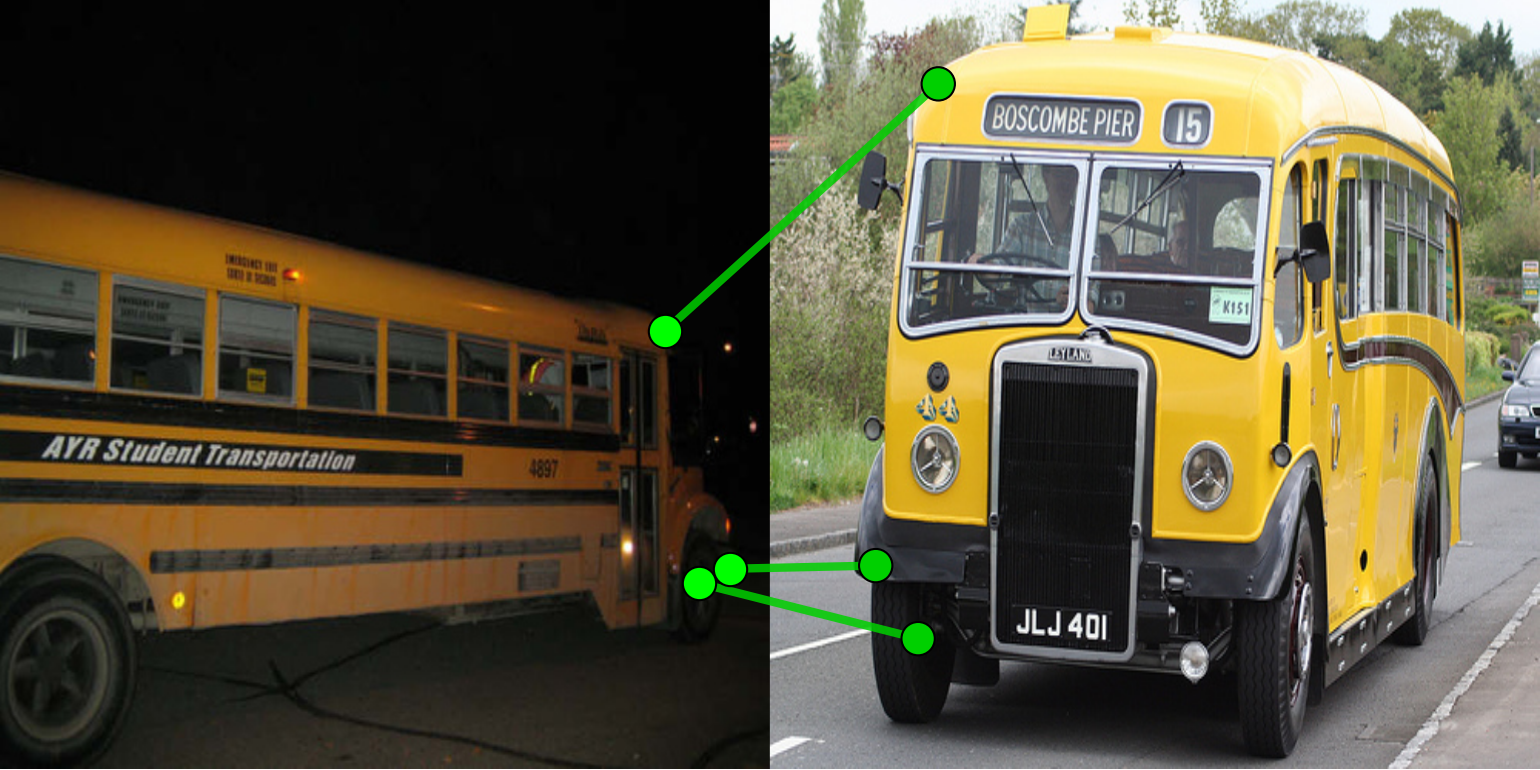}
\end{tabular}

\end{minipage}%
}
\end{figure*}

\noindent\textbf{3D--3D evaluation.}
For each category, we evaluate all model pairs within a subset,
retaining pairs where at least $60\%$ of each model's points have
a DINOv2 match with cosine similarity above $0.575$
($68\%$ of pairs retained on average).
Within retained pairs, we prune correspondences with endpoint
feature similarity below $0.575$.
For each ground-truth query keypoint, we select the correspondence
with the nearest source point, provided its distance is below
$0.05$ times the target bounding-box diagonal.
Keypoints without a surviving correspondence are excluded from PCK,
and their fraction is reflected in the GT-used rate.
Predictions are correct if their target locations lie within
$\alpha$ times the target bounding-box diagonal of the GT keypoint.
We average results over three random subsets of 100 models per
category (38 and 90 models for Cap and Helmet, respectively),
as shown in \autoref{tab:keypointnet_table}.
Our method achieves 93\% PCK@0.10, 97\% PCK@0.15,
and 99\% PCK@0.20, with a median GT-used rate of 57.6\%.
Reduced coverage reflects semantic pruning that avoids
uncertain matches between structurally different instances.
Detailed per-label results and evaluable keypoint counts
appear in \autoref{Sup:per_label}.
On the same pairs, we also outperform Back to 3D~\cite{wimmer2024back}
in both 1-shot and 3-shot settings without requiring keypoint
annotations (see \autoref{sup:backto3d}).

\noindent\textbf{3D--2D evaluation.}
We further evaluate our generated 3D correspondences in image
space by projecting them onto randomly rendered views.
As in our data generation, only correspondences whose target
points are visible in the rendered image are retained.
Across 16 categories, we achieve 94.2\% PCK@0.10 in 2D,
with an average GT-used rate of 45.3\% of ground-truth keypoints in the rendered views.
On the same visible correspondences, 3D PCK@0.10 reaches 95.1\%,
compared to 93\% in the original 3D--3D evaluation,
which includes all found keypoints regardless of visibility.
The slight decrease may result from geometric inaccuracies
in the approximate 3D correspondences, which can introduce
projection errors or visibility inconsistencies in image space.
Nevertheless, the results demonstrate that correspondence
accuracy is largely preserved when transferred to 2D.
The full evaluation protocol and per-category results
are provided in \autoref{sup:3d2d_eval}.

\subsection{Ablations and Hyperparameter Sensitivity}
\label{sec:ablation}
\begingroup
\setlength{\parskip}{0pt}
We summarize the main findings of our ablation studies here and defer detailed quantitative analyses, per-category tables, and additional sensitivity experiments to \autoref{sup:ablation_study} and \autoref{Sup:per_label}.
\\

\noindent\textbf{Hyperparameters.}
Performance is stable across a reasonable range of the CPA smoothness weight \(\lambda_{\mathrm{smooth}}\) and semantic threshold \(\tau_{\mathrm{sem}}\). As detailed in \autoref{sup:ablation_semantic_filtering}, \(\tau_{\mathrm{sem}}\) controls an accuracy--coverage trade-off, and \(\tau_{\mathrm{sem}}=0.575\) provides a practical balance between retaining reliable correspondences and preserving sufficient coverage. For the CPA smoothness weight, we observe only minor variation in KeypointNet \cite{you2020} PCK across tested values, likely because the sparse annotations are mostly concentrated on salient corners and extremities. Qualitatively, however, \(\lambda_{\mathrm{smooth}}=0.3\) better preserves the source object structure while still allowing sufficient local deformation to align with the target, producing more coherent dense correspondences in regions not captured by the sparse KeypointNet annotations. We therefore use \(\lambda_{\mathrm{smooth}}=0.3\) and \(\tau_{\mathrm{sem}}=0.575\) in all experiments.

\noindent\textbf{Registration model.}
We compare continuous piecewise-affine (CPA) registration with a global affine transformation and thin-plate splines (TPS); see \autoref{sup:ablation_registration} for details. Affine performs substantially worse at all PCK thresholds, while TPS achieves accuracy closer to CPA. CPA nonetheless achieves comparable or slightly better accuracy at most evaluated thresholds and is approximately \(3\times\) faster in our setup, likely because its tetrahedral piecewise-affine parameterization is local, whereas TPS relies on a global radial-basis deformation.

\par\noindent\textbf{Semantic filtering.}
Semantic filtering consistently improves correspondence quality (see \autoref{sup:ablation_semantic_filtering}). Without correspondence-level or pair-level filtering, geometric registration alone achieves over \(84\%\) PCK@0.10. Adding correspondence-level semantic pruning improves performance to about \(91\%\) PCK@0.10, while pair-level filtering yields a further improvement to about \(93\%\) PCK@0.10 on the retained pairs. We additionally provide a detailed per-semantic-label analysis in \autoref{Sup:per_label}, following the evaluation protocol of~\cite{wimmer2024back}, as well as a qualitative analysis of DINOv2-based semantic filtering and correspondence pruning in \autoref{sup:supp_semantic_analy}.

\noindent\textbf{Sensitivity to rotation misalignment.}
We evaluate robustness to imperfect canonical alignment.
Mean PCK@0.10 decreases from 93\% to 89\% and 85\%
under 5$^\circ$ and 10$^\circ$ rotation perturbations,
respectively, but drops substantially for larger rotations,
confirming the importance of approximate alignment
(see \autoref{sup:rotation_sensitivity}).
\endgroup

\subsection{Discussion}

Our results demonstrate that approximate semantic-geometric
correspondences generated from rigid 3D objects provide effective
supervision for cross-instance matching. KeypointNet evaluations
validate our 3D--3D and projected 3D--2D correspondences,
while SPair-71k results show improvements in real-image
2D--2D matching, including generalization to non-rigid categories. The scalability of our method lies in leveraging large-scale,
approximately aligned 3D collections without manual correspondence
annotations. While alignment remains a prerequisite, existing
datasets such as Objaverse-OA~\cite{lu2025orientationmatters} and advances
in orientation estimation~\cite{orient_anything} and scalable canonicalization~\cite{jin2025oneshot,jin2026canoverse}
make aligned 3D assets increasingly available.
Moreover, SemGeo-Gen scales combinatorially, generating
correspondences for up to $\binom{N}{2}$ pairs from $N$
aligned objects, with multiple rendered view pairs per pair.
Together, these enable scalable semantic-geometric supervision
without manual point-level annotations.

\subsection{Limitations}

Our approach has several limitations. First, the CPA deformation model cannot fully capture large cross-instance structural differences without supervision, potentially leading to imperfect alignments. Second, our semantic filtering relies on DINOv2 descriptors, although our modular pipeline can benefit from stronger future feature extractors. Third, our method assumes rigid, approximately aligned 3D objects and does not directly handle non-rigid pose variations; a promising extension is to cluster objects by pose and apply CPA within each cluster to accommodate residual variations. Finally, our synthetic 3D assets introduce a domain gap when transferring to real images, although pretraining followed by real-data fine-tuning proves effective. Growing collections of real, canonically aligned 3D assets could further reduce this gap.

\section{Conclusion}

We presented a scalable unsupervised pipeline for generating cross-instance semantic-geometric correspondences from 3D assets without manual annotation. Although approximate, the generated correspondences align well with human annotations and provide an effective pretraining signal for real-image semantic correspondence models. Our results demonstrate that scalable semantic-geometric data can complement manual annotations and improve cross-instance correspondence learning.

\begin{ack}
\begin{flushleft}
Roy Amoyal was partially supported by the Kreitman School of Advanced
Graduate Studies. Shira Ifergane was partially supported by the VATAT
National Excellence Scholarship for MSc Students in AI and Data Science.
\end{flushleft}
\end{ack}
{\small
\bibliographystyle{plainnat}
\bibliography{refs}

\begin{thebibliography}{42}
\providecommand{\natexlab}[1]{#1}
\providecommand{\url}[1]{\texttt{#1}}
\expandafter\ifx\csname urlstyle\endcsname\relax
  \providecommand{\doi}[1]{doi: #1}\else
  \providecommand{\doi}{doi: \begingroup \urlstyle{rm}\Url}\fi

\bibitem[Aflalo et~al.(2026)Aflalo, Treister, Baskin, and Freifeld]{Aflalo:NeurIPS:2026:FG-CPAB}
Amit Aflalo, Eran Treister, Chaim Baskin, and Oren Freifeld.
\newblock Factorized gradients for scalable highly-expressive parametric diffeomorphisms.
\newblock In \emph{Advances in Neural Information Processing Systems}, 2026.

\bibitem[Amoyal et~al.(2026)Amoyal, Freifeld, and Baskin]{Amoyal:CVPR:2026:GSA}
Roy Amoyal, Oren Freifeld, and Chaim Baskin.
\newblock {Cross-Instance Gaussian Splatting Registration via Geometry-Aware Feature-Guided Alignment}.
\newblock In \emph{CVPR}, 2026.

\bibitem[Barel et~al.(2024)Barel, Weber, Mualem, Finder, and Freifeld]{Barel:ECCV:2024:SpaceJAM}
Nir Barel, Ron~Shapira Weber, Nir Mualem, Shahaf~E. Finder, and Oren Freifeld.
\newblock {SpaceJAM}: a lightweight and regularization-free method for fast joint alignment of images.
\newblock In \emph{European Conference on Computer Vision}, 2024.

\bibitem[Cho et~al.(2023)Cho, Hong, and Kim]{cho2022cats++}
Seokju Cho, Sunghwan Hong, and Seungryong Kim.
\newblock {CATs++}: Boosting cost aggregation with convolutions and transformers.
\newblock \emph{IEEE Transactions on Pattern Analysis and Machine Intelligence}, 45\penalty0 (6):\penalty0 7174--7194, 2023.
\newblock \doi{10.1109/TPAMI.2022.3218727}.

\bibitem[Cuttano et~al.(2026)Cuttano, Trivigno, Masone, and Roth]{cuttano2026marco}
Claudia Cuttano, Gabriele Trivigno, Carlo Masone, and Stefan Roth.
\newblock {MARCO}: Navigating the unseen space of semantic correspondence.
\newblock In \emph{Proceedings of the IEEE/CVF Conference on Computer Vision and Pattern Recognition (CVPR)}, 2026.

\bibitem[Dyke et~al.(2019)Dyke, Stride, Lai, Rosin, Aubry, Boyarski, Bronstein, Bronstein, Cremers, Fisher, Groueix, Guo, Kim, Kimmel, L{\"a}hner, Li, Litany, Remez, Rodol{\`a}, Russell, Sahillioglu, Slossberg, Tam, Vestner, Wu, and Yang]{dyke2019shrec_19}
R.~M. Dyke, C.~Stride, Y.-K. Lai, P.~L. Rosin, M.~Aubry, A.~Boyarski, A.~M. Bronstein, M.~M. Bronstein, D.~Cremers, M.~Fisher, T.~Groueix, D.~Guo, V.~G. Kim, R.~Kimmel, Z.~L{\"a}hner, K.~Li, O.~Litany, T.~Remez, E.~Rodol{\`a}, B.~C. Russell, Y.~Sahillioglu, R.~Slossberg, G.~K.~L. Tam, M.~Vestner, Z.~Wu, and J.~Yang.
\newblock {SHREC}'19: Shape correspondence with isometric and non-isometric deformations.
\newblock In \emph{Eurographics Workshop on {3D} Object Retrieval}, pages 111--119. The Eurographics Association, 2019.
\newblock \doi{10.2312/3dor.20191069}.

\bibitem[Fan et~al.(2017)Fan, Su, and Guibas]{fan2017chamfer}
Haoqiang Fan, Hao Su, and Leonidas Guibas.
\newblock A point set generation network for {3D} object reconstruction from a single image.
\newblock In \emph{Proceedings of the IEEE Conference on Computer Vision and Pattern Recognition (CVPR)}, 2017.

\bibitem[Freifeld et~al.(2017)Freifeld, Hauberg, Batmanghelich, and Fisher~III]{Freifeld:TPAMI:2017:CPAB}
Oren Freifeld, S{\o}ren Hauberg, Kayhan Batmanghelich, and John~W. Fisher~III.
\newblock Transformations based on continuous piecewise-affine velocity fields.
\newblock \emph{IEEE Transactions on Pattern Analysis and Machine Intelligence}, 39\penalty0 (12):\penalty0 2496--2509, 2017.

\bibitem[Groueix et~al.(2018)Groueix, Fisher, Kim, Russell, and Aubry]{groueix2018b3dcodeddataset}
Thibault Groueix, Matthew Fisher, Vladimir~G. Kim, Bryan Russell, and Mathieu Aubry.
\newblock {3D-CODED}: {3D} correspondences by deep deformation.
\newblock In \emph{ECCV}, 2018.

\bibitem[Gupta et~al.(2023)Gupta, Jampani, Esteves, Shrivastava, Makadia, Snavely, and Kar]{gupta2023asic}
Kamal Gupta, Varun Jampani, Carlos Esteves, Abhinav Shrivastava, Ameesh Makadia, Noah Snavely, and Abhishek Kar.
\newblock {ASIC}: Aligning sparse in-the-wild image collections.
\newblock In \emph{Proceedings of the IEEE/CVF International Conference on Computer Vision (ICCV)}, 2023.

\bibitem[Ham et~al.(2018)Ham, Cho, Schmid, and Ponce]{ham2018pascal_proposalflow}
Bumsub Ham, Minsu Cho, Cordelia Schmid, and Jean Ponce.
\newblock Proposal flow: Semantic correspondences from object proposals.
\newblock \emph{IEEE Transactions on Pattern Analysis and Machine Intelligence}, 40\penalty0 (7):\penalty0 1711--1725, 2018.
\newblock \doi{10.1109/TPAMI.2017.2724510}.

\bibitem[Hedlin et~al.(2023)Hedlin, Sharma, Mahajan, Isack, Kar, Tagliasacchi, and Yi]{hedlin2023unsupervisedstablediffusion}
Eric Hedlin, Gopal Sharma, Shweta Mahajan, Hossam Isack, Abhishek Kar, Andrea Tagliasacchi, and Kwang~Moo Yi.
\newblock Unsupervised semantic correspondence using stable diffusion.
\newblock In \emph{Advances in Neural Information Processing Systems 36 (NeurIPS 2023)}, 2023.

\bibitem[Hirsch et~al.(2025)Hirsch, {Shapira Weber}, Ifergane, and Freifeld]{Hirsch:NeurIPS:2025:FastJAM}
Omri Hirsch, Ron {Shapira Weber}, Shira Ifergane, and Oren Freifeld.
\newblock {FastJAM}: a fast joint alignment model for images.
\newblock In \emph{Advances in Neural Information Processing Systems (NeurIPS)}, 2025.

\bibitem[Huang et~al.(2022)Huang, Yang, He, Zhang, He, and Shrivastava]{huang2022SCorrSan}
Shuaiyi Huang, Luyu Yang, Bo~He, Songyang Zhang, Xuming He, and Abhinav Shrivastava.
\newblock Learning semantic correspondence with sparse annotations.
\newblock In \emph{European Conference on Computer Vision (ECCV)}, 2022.

\bibitem[Jin et~al.(2025)Jin, Wang, Chen, Dai, Gao, Qin, and Chen]{jin2025oneshot}
Li~Jin, Yujie Wang, Wenzheng Chen, Qiyu Dai, Qingzhe Gao, Xueying Qin, and Baoquan Chen.
\newblock One-shot {3D} object canonicalization based on geometric and semantic consistency.
\newblock In \emph{Proceedings of the IEEE/CVF Conference on Computer Vision and Pattern Recognition (CVPR)}, pages 16850--16859, 2025.

\bibitem[Jin et~al.(2026)Jin, Yang, Chen, Wang, Hao, Jia, Yin, Hu, Zhang, Luo, Yuan, Quan, Wang, and Qin]{jin2026canoverse}
Li~Jin, Yuchen Yang, Weikai Chen, Yujie Wang, Dehao Hao, Tanghui Jia, Yingda Yin, Zeyu Hu, Runze Zhang, Keyang Luo, Li~Yuan, Long Quan, Xin Wang, and Xueying Qin.
\newblock {CanoVerse}: {3D} object scalable canonicalization and dataset for generation and pose.
\newblock In \emph{European Conference on Computer Vision (ECCV)}, 2026.

\bibitem[Ju et~al.(2024)Ju, Hu, Zhang, Zhang, Jiang, and Xu]{ju2025robo_abc}
Yuanchen Ju, Kaizhe Hu, Guowei Zhang, Gu~Zhang, Mingrun Jiang, and Huazhe Xu.
\newblock {Robo-ABC}: Affordance generalization beyond categories via semantic correspondence for robot manipulation.
\newblock In \emph{European Conference on Computer Vision (ECCV)}, pages 222--239. Springer, 2024.

\bibitem[Krishnan et~al.(2024)Krishnan, Kundu, Maninis, Hays, and Brown]{omninocs2024}
Akshay Krishnan, Abhijit Kundu, Kevis-Kokitsi Maninis, James Hays, and Matthew Brown.
\newblock {OmniNOCS}: A unified {NOCS} dataset and model for {3D} lifting of {2D} objects.
\newblock In \emph{European Conference on Computer Vision (ECCV)}, 2024.

\bibitem[Lou et~al.(2020)Lou, You, Li, Cheng, Li, Ma, Wang, and Lu]{lou2020humanCorrespondenceConsensus}
Yujing Lou, Yang You, Chengkun Li, Zhoujun Cheng, Liangwei Li, Lizhuang Ma, Weiming Wang, and Cewu Lu.
\newblock Human correspondence consensus for {3D} object semantic understanding.
\newblock In \emph{Proceedings of the European Conference on Computer Vision (ECCV)}, 2020.

\bibitem[Lu et~al.(2025)Lu, Tian, Jiang, Zhao, Yang, Ouyang, Hu, Yu, Shen, and Liao]{lu2025orientationmatters}
Yichong Lu, Yuzhuo Tian, Zijin Jiang, Yikun Zhao, Yuanbo Yang, Hao Ouyang, Haoji Hu, Huimin Yu, Yujun Shen, and Yiyi Liao.
\newblock Orientation matters: Making {3D} generative models orientation-aligned.
\newblock In \emph{Advances in Neural Information Processing Systems}, volume~38, 2025.

\bibitem[Luo et~al.(2023)Luo, Dunlap, Park, Holynski, and Darrell]{luo2023DiffusionHyperFeaturesTimeSpace}
Grace Luo, Lisa Dunlap, Dong~Huk Park, Aleksander Holynski, and Trevor Darrell.
\newblock Diffusion hyperfeatures: Searching through time and space for semantic correspondence.
\newblock In \emph{Advances in Neural Information Processing Systems}, 2023.

\bibitem[Mariotti et~al.(2024)Mariotti, Mac~Aodha, and Bilen]{mariotti2024sphericalmap}
Octave Mariotti, Oisin Mac~Aodha, and Hakan Bilen.
\newblock Improving semantic correspondence with viewpoint-guided spherical maps.
\newblock In \emph{Proceedings of the IEEE/CVF Conference on Computer Vision and Pattern Recognition (CVPR)}, pages 19521--19530, June 2024.

\bibitem[Mariotti et~al.(2025)Mariotti, Du, Bhalgat, Mac~Aodha, and Bilen]{mariotti2025jamaisvu}
Octave Mariotti, Zhipeng Du, Yash Bhalgat, Oisin Mac~Aodha, and Hakan Bilen.
\newblock Jamais vu: Exposing the generalization gap in supervised semantic correspondence.
\newblock In \emph{Advances in Neural Information Processing Systems (NeurIPS)}, 2025.

\bibitem[Min et~al.(2019)Min, Lee, Ponce, and Cho]{min2019spair71k}
Juhong Min, Jongmin Lee, Jean Ponce, and Minsu Cho.
\newblock {SPair-71k}: A large-scale benchmark for semantic correspondence.
\newblock \emph{arXiv preprint arXiv:1908.10543}, 2019.

\bibitem[Ofri-Amar et~al.(2023)Ofri-Amar, Geyer, Kasten, and Dekel]{ofriamar2023neuralCongealing}
Dolev Ofri-Amar, Michal Geyer, Yoni Kasten, and Tali Dekel.
\newblock Neural congealing: Aligning images to a joint semantic atlas.
\newblock In \emph{Proceedings of the IEEE/CVF Conference on Computer Vision and Pattern Recognition (CVPR)}, pages 19403--19412, 2023.

\bibitem[Oquab et~al.(2024)Oquab, Darcet, Moutakanni, Vo, Szafraniec, Khalidov, Fernandez, Haziza, Massa, El-Nouby, et~al.]{oquab2024dinov2}
Maxime Oquab, Timoth{\'e}e Darcet, Th{\'e}o Moutakanni, Huy Vo, Marc Szafraniec, Vasil Khalidov, Pierre Fernandez, Daniel Haziza, Francisco Massa, Alaaeldin El-Nouby, et~al.
\newblock {DINOv2}: Learning robust visual features without supervision.
\newblock \emph{Transactions on Machine Learning Research}, 2024.

\bibitem[Pratikakis et~al.(2016)Pratikakis, Savelonas, Arnaoutoglou, Ioannakis, Koutsoudis, Theoharis, Tran, Nguyen, Pham, Nguyen, Le, Tran, To, Truong, Phan, Nguyen, Than, Mac, Do, Duong, Furuya, Ohbuchi, Aono, Tashiro, Pickup, Sun, Rosin, and Martin]{pratikakis2016shrec_16}
I.~Pratikakis, M.~A. Savelonas, F.~Arnaoutoglou, G.~Ioannakis, A.~Koutsoudis, T.~Theoharis, M.~T. Tran, V.~T. Nguyen, V.~K. Pham, H.~D. Nguyen, H.~A. Le, B.~H. Tran, Q.~H. To, M.~B. Truong, T.~V. Phan, M.~D. Nguyen, T.~A. Than, K.~N.~C. Mac, M.~N. Do, A.~D. Duong, T.~Furuya, R.~Ohbuchi, M.~Aono, S.~Tashiro, D.~Pickup, X.~Sun, P.~L. Rosin, and R.~R. Martin.
\newblock {SHREC}'16 track: Partial shape queries for {3D} object retrieval.
\newblock In \emph{Eurographics Workshop on 3D Object Retrieval}, pages 79--88. The Eurographics Association, 2016.
\newblock \doi{10.2312/3dor.20161091}.

\bibitem[Rombach et~al.(2022)Rombach, Blattmann, Lorenz, Esser, and Ommer]{rombach2022stablediffusion}
Robin Rombach, Andreas Blattmann, Dominik Lorenz, Patrick Esser, and Bj{\"o}rn Ommer.
\newblock High-resolution image synthesis with latent diffusion models.
\newblock In \emph{Proceedings of the IEEE/CVF Conference on Computer Vision and Pattern Recognition (CVPR)}, 2022.

\bibitem[Shi et~al.(2021)Shi, Xue, You, and Lu]{shi2021skeletonmerger}
Ruoxi Shi, Zhengrong Xue, Yang You, and Cewu Lu.
\newblock {Skeleton Merger}: An unsupervised aligned keypoint detector.
\newblock In \emph{Proceedings of the IEEE/CVF Conference on Computer Vision and Pattern Recognition (CVPR)}, pages 43--52, 2021.

\bibitem[Wah et~al.(2011)Wah, Branson, Welinder, Perona, and Belongie]{WahCUB_200_2011}
C.~Wah, S.~Branson, P.~Welinder, P.~Perona, and S.~Belongie.
\newblock The {Caltech-UCSD} birds-200-2011 dataset.
\newblock Technical Report CNS-TR-2011-001, California Institute of Technology, 2011.

\bibitem[Wandel and Wang(2025)]{wandel2025semalign3d}
Krispin Wandel and Hesheng Wang.
\newblock {SemAlign3D}: Semantic correspondence between {RGB}-images through aligning {3D} object-class representations.
\newblock In \emph{Proceedings of the IEEE/CVF Conference on Computer Vision and Pattern Recognition (CVPR)}, 2025.

\bibitem[Wang(2019)]{wang2019visualcorrespondencerobotics}
Xiaolong Wang.
\newblock \emph{Learning and Reasoning with Visual Correspondence in Time}.
\newblock PhD thesis, Carnegie Mellon University, Pittsburgh, PA, September 2019.

\bibitem[Wang et~al.(2025)Wang, Zhang, Pang, Du, Zhao, and Zhao]{orient_anything}
Zehan Wang, Ziang Zhang, Tianyu Pang, Chao Du, Hengshuang Zhao, and Zhou Zhao.
\newblock Orient anything: Learning robust object orientation estimation from rendering {3D} models.
\newblock In \emph{Proceedings of the International Conference on Machine Learning (ICML)}, 2025.

\bibitem[Wimmer et~al.(2024)Wimmer, Wonka, and Ovsjanikov]{wimmer2024back}
Thomas Wimmer, Peter Wonka, and Maks Ovsjanikov.
\newblock Back to {3D}: Few-shot {3D} keypoint detection with back-projected {2D} features.
\newblock In \emph{Proceedings of the IEEE/CVF Conference on Computer Vision and Pattern Recognition}, 2024.

\bibitem[Xu et~al.(2022)Xu, Jin, Zeng, Liu, Qian, Ouyang, Luo, and Wang]{xu2022poseforeverything}
Lumin Xu, Sheng Jin, Wang Zeng, Wentao Liu, Chen Qian, Wanli Ouyang, Ping Luo, and Xiaogang Wang.
\newblock Pose for everything: Towards category-agnostic pose estimation.
\newblock In \emph{European Conference on Computer Vision (ECCV)}, 2022.

\bibitem[You et~al.(2020)You, Lou, Li, Cheng, Li, Ma, Lu, and Wang]{you2020}
Yang You, Yujing Lou, Chengkun Li, Zhoujun Cheng, Liangwei Li, Lizhuang Ma, Cewu Lu, and Weiming Wang.
\newblock {KeypointNet}: A large-scale {3D} keypoint dataset aggregated from numerous human annotations.
\newblock In \emph{Proceedings of the IEEE/CVF Conference on Computer Vision and Pattern Recognition (CVPR)}, pages 13647--13656, 2020.

\bibitem[You et~al.(2022)You, Liu, Ze, Li, Wang, and Lu]{you2022ukpgan}
Yang You, Wenhai Liu, Yanjie Ze, Yong-Lu Li, Weiming Wang, and Cewu Lu.
\newblock {UKPGAN}: A general self-supervised keypoint detector.
\newblock In \emph{Proceedings of the IEEE/CVF Conference on Computer Vision and Pattern Recognition}, 2022.

\bibitem[Yu et~al.(2021)Yu, Xu, Zhang, Zhao, Guan, and Tao]{yu2021AP10k}
Hang Yu, Yufei Xu, Jing Zhang, Wei Zhao, Ziyu Guan, and Dacheng Tao.
\newblock {AP-10K}: A benchmark for animal pose estimation in the wild.
\newblock In \emph{Thirty-fifth Conference on Neural Information Processing Systems Datasets and Benchmarks Track (Round 2)}, 2021.

\bibitem[Zhang et~al.(2023)Zhang, Herrmann, Hur, Polania~Cabrera, Jampani, Sun, and Yang]{zhang2023taleTwoFeatures}
Junyi Zhang, Charles Herrmann, Junhwa Hur, Luisa Polania~Cabrera, Varun Jampani, Deqing Sun, and Ming-Hsuan Yang.
\newblock A tale of two features: {Stable Diffusion} complements {DINO} for zero-shot semantic correspondence.
\newblock In \emph{Advances in Neural Information Processing Systems (NeurIPS)}, volume~36, 2023.

\bibitem[Zhang et~al.(2024)Zhang, Herrmann, Hur, Chen, Jampani, Sun, and Yang]{zhang2024tellingLeftFromRight}
Junyi Zhang, Charles Herrmann, Junhwa Hur, Eric Chen, Varun Jampani, Deqing Sun, and Ming-Hsuan Yang.
\newblock Telling left from right: Identifying geometry-aware semantic correspondence.
\newblock In \emph{Proceedings of the IEEE/CVF Conference on Computer Vision and Pattern Recognition}, 2024.

\bibitem[Zhu et~al.(2025)Zhu, Ju, Zhang, Wang, Yuan, Hu, and Xu]{zhu2025densematcher}
Junzhe Zhu, Yuanchen Ju, Junyi Zhang, Muhan Wang, Zhecheng Yuan, Kaizhe Hu, and Huazhe Xu.
\newblock {DenseMatcher}: Learning {3D} semantic correspondence for category-level manipulation from a single demo.
\newblock In \emph{International Conference on Learning Representations (ICLR)}, 2025.

\bibitem[Zuffi et~al.(2017)Zuffi, Kanazawa, Jacobs, and Black]{zuffi2017menagerie}
Silvia Zuffi, Angjoo Kanazawa, David~W. Jacobs, and Michael~J. Black.
\newblock {3D} menagerie: Modeling the {3D} shape and pose of animals.
\newblock In \emph{Proceedings of the IEEE Conference on Computer Vision and Pattern Recognition (CVPR)}, pages 5524--5532, 2017.
\newblock \doi{10.1109/CVPR.2017.586}.

\end{thebibliography}
}

\clearpage
\appendix
\begin{center}
{\LARGE \textbf{SemGeo-Gen: Unsupervised Generation of Approximate Cross-Instance Semantic-Geometric Correspondences}}\\[4pt]
{\large ------------}\\[4pt]
{\LARGE \textbf{Supplementary Material}}
\end{center}

\vspace{1em}

In this supplementary material, we provide additional qualitative results, extended ablations, and detailed per-label analysis to complement the main paper.

\section*{Contents}
\begin{enumerate}[]
    \item \autoref{sup:qualitative} - Qualitative Results on SPair-71k and KeypointNet.
\item \autoref{sup:additional_quant} - Additional Quantitative Results.
    \item \autoref{sup:ablation_study} - Extended Ablation Study.
    \item \autoref{sup:supp_semantic_analy} - Qualitative Analysis of DINOv2 Based Semantic Filtering.
    \item \autoref{sup:3d2d_eval} - 3D--2D Correspondence Evaluation on KeypointNet.
    \item \autoref{Sup:per_label} - Detailed PCK Results by Semantic Label and KeypointNet Category.
\end{enumerate}
\clearpage

\section{Qualitative Results}
\label{sup:qualitative}

\subsection{SPair-71k Transfer Results}

\begin{figure*}[h]
\centering
\setlength{\tabcolsep}{3pt}
\begin{tabular}{cc}
\textbf{MARCO \cite{cuttano2026marco} Real-only} & \textbf{MARCO \cite{cuttano2026marco} Ours P.T.$\rightarrow$Real} \\
\includegraphics[width=0.48\textwidth]{figures/qualitative/marco/match_real_chair_bold.png} &
\includegraphics[width=0.48\textwidth]{figures/qualitative/marco/match_ours_chair_bold.png} \\
\includegraphics[width=0.48\textwidth]{figures/qualitative/marco/match_real_boat_bold.png} &
\includegraphics[width=0.48\textwidth]{figures/qualitative/marco/match_ours_boat_bold.png} \\
\includegraphics[width=0.48\textwidth]{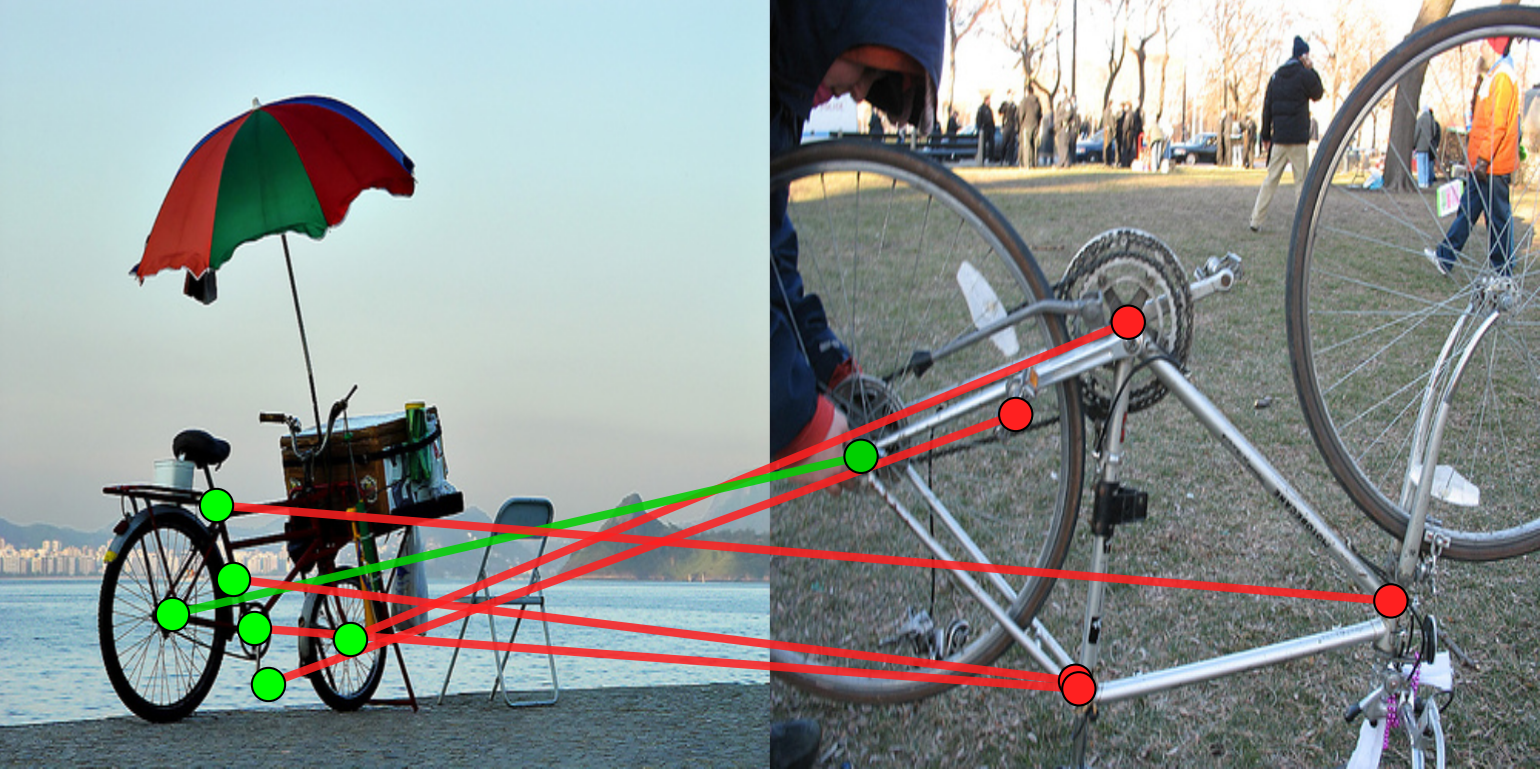} &
\includegraphics[width=0.48\textwidth]{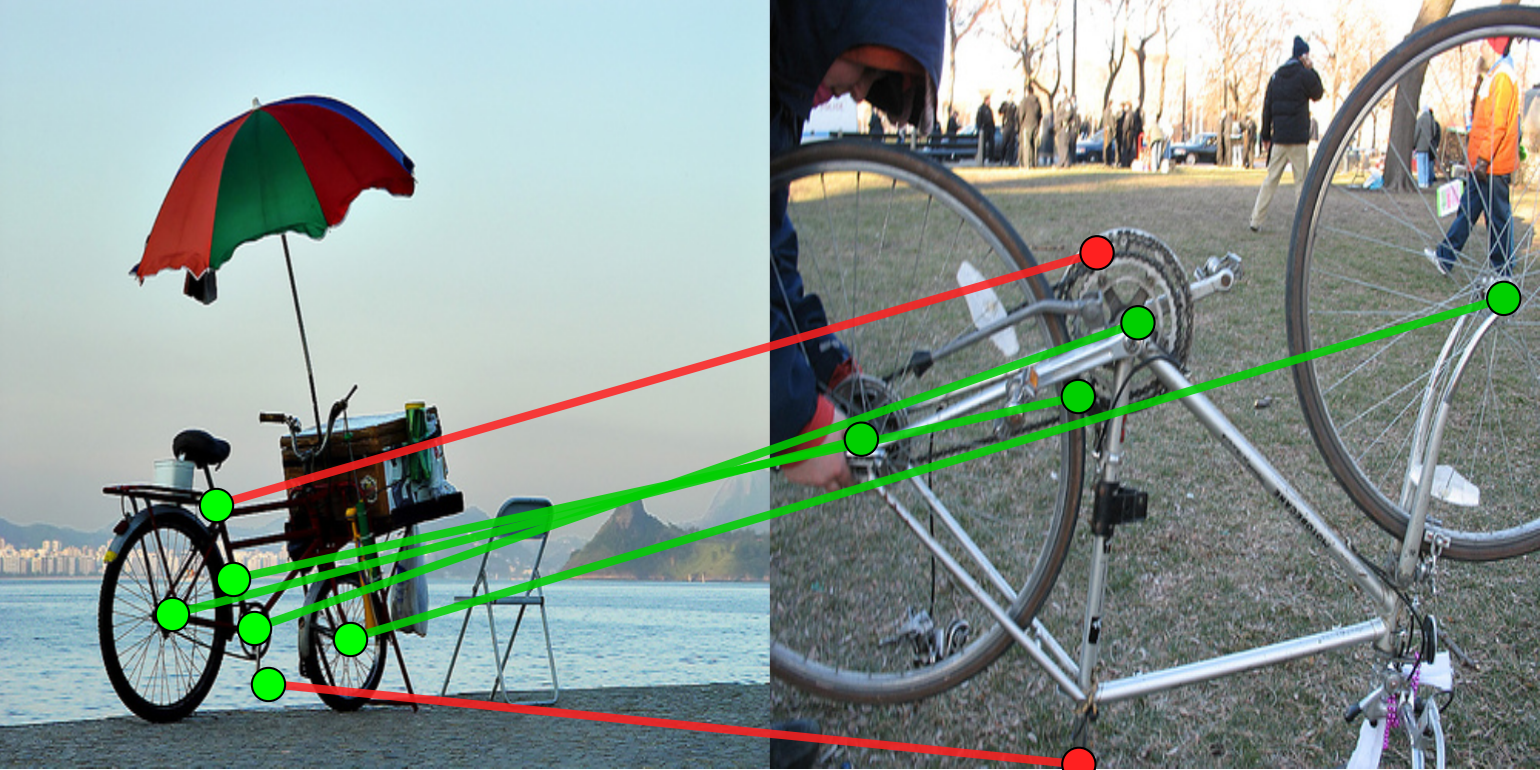} \\
\end{tabular}
\caption{\textbf{Qualitative SPair-71k \cite{min2019spair71k} results for MARCO \cite{cuttano2026marco}.}
Each image shows a source--target pair with predicted correspondences. We compare MARCO trained on SPair-71k against MARCO pretrained on our generated correspondences and then fine-tuned on SPair-71k.}
\label{fig:qual_marco}
\end{figure*}

\begin{figure*}[t]
\centering
\setlength{\tabcolsep}{3pt}
\begin{tabular}{cc}
\textbf{GeoAware-SC Real-only} & \textbf{GeoAware-SC Ours P.T.$\rightarrow$Real} \\
\includegraphics[width=0.48\textwidth]{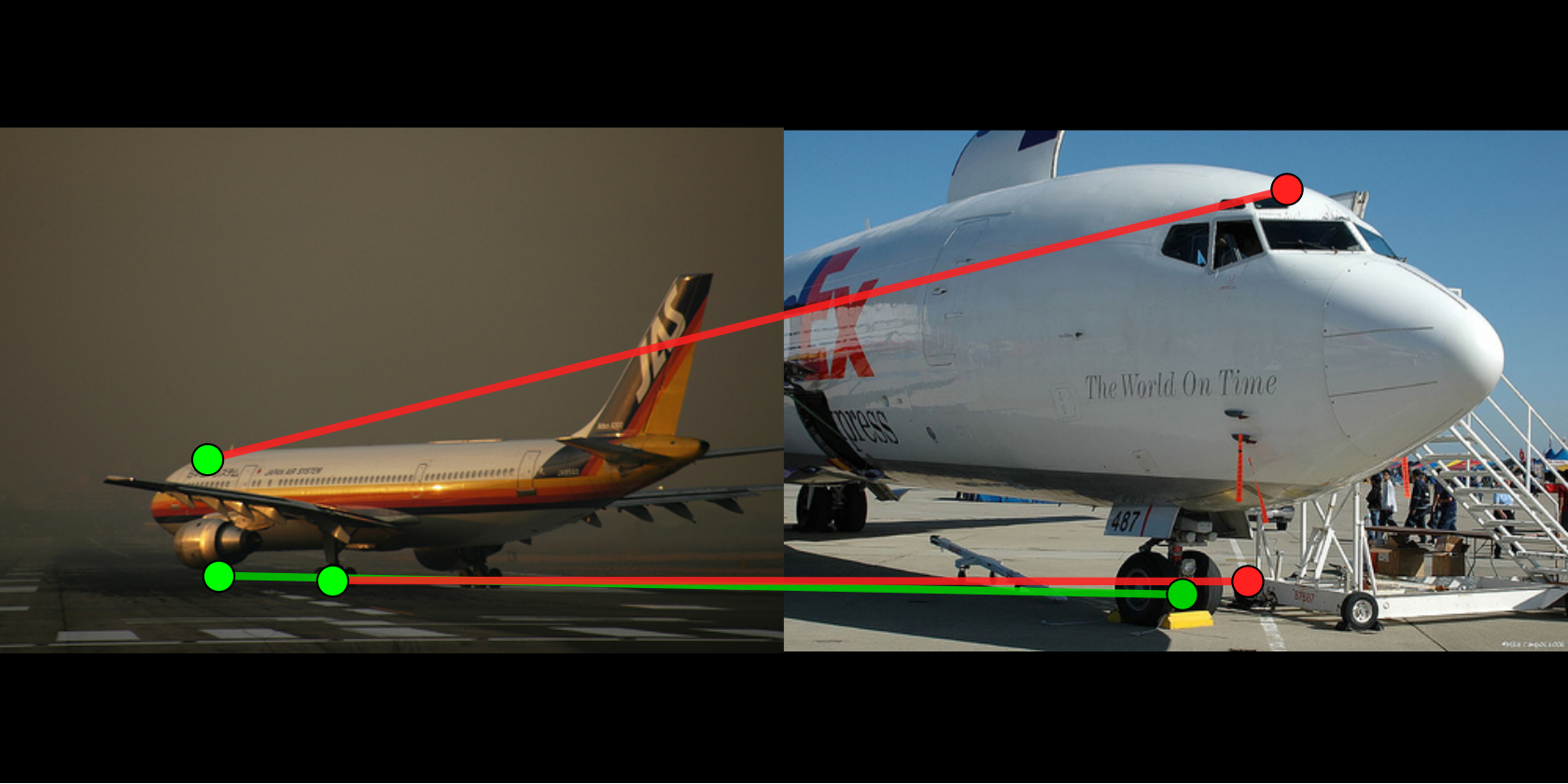} &
\includegraphics[width=0.48\textwidth]{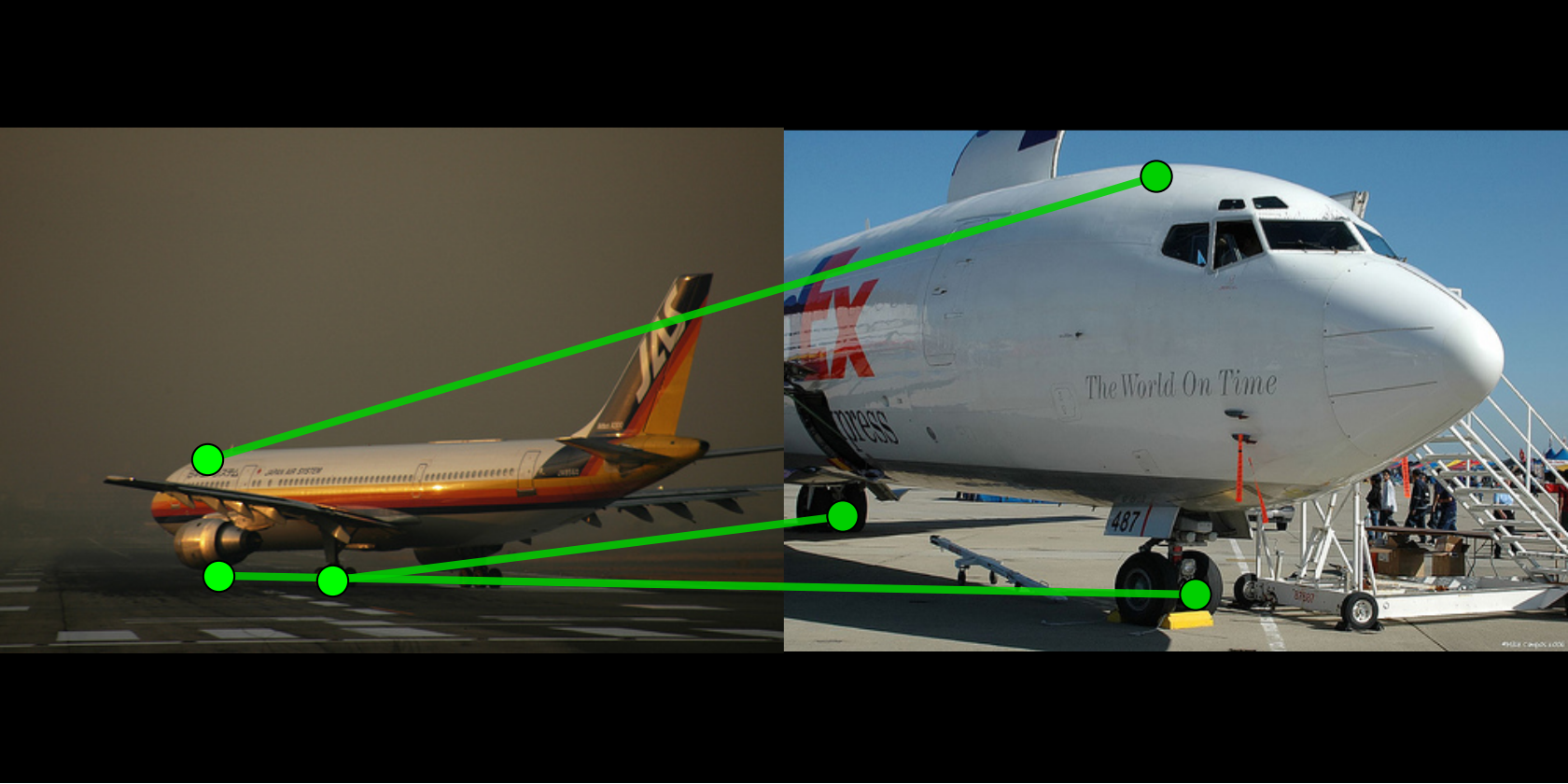} \\
\includegraphics[width=0.48\textwidth]{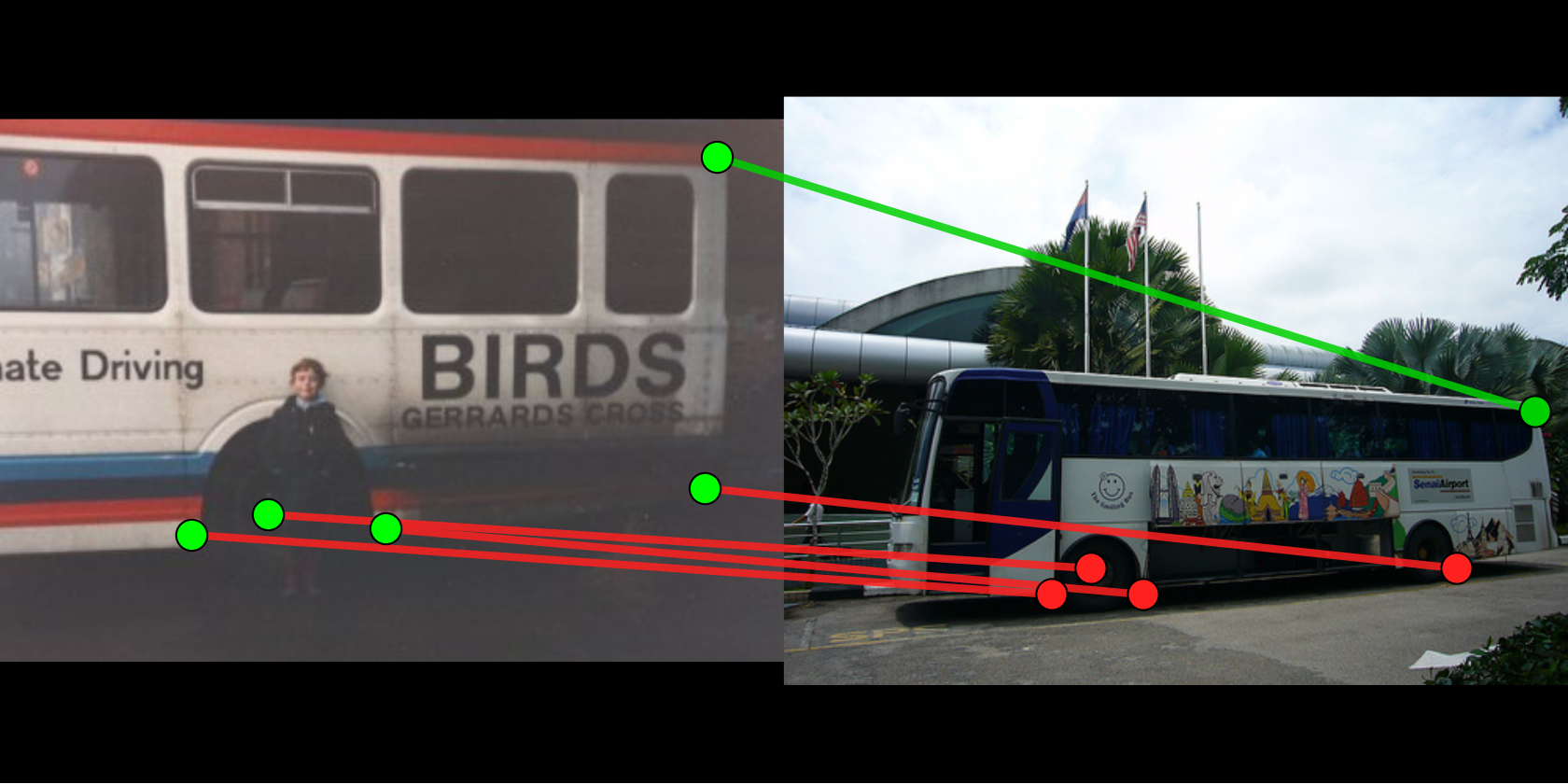} &
\includegraphics[width=0.48\textwidth]{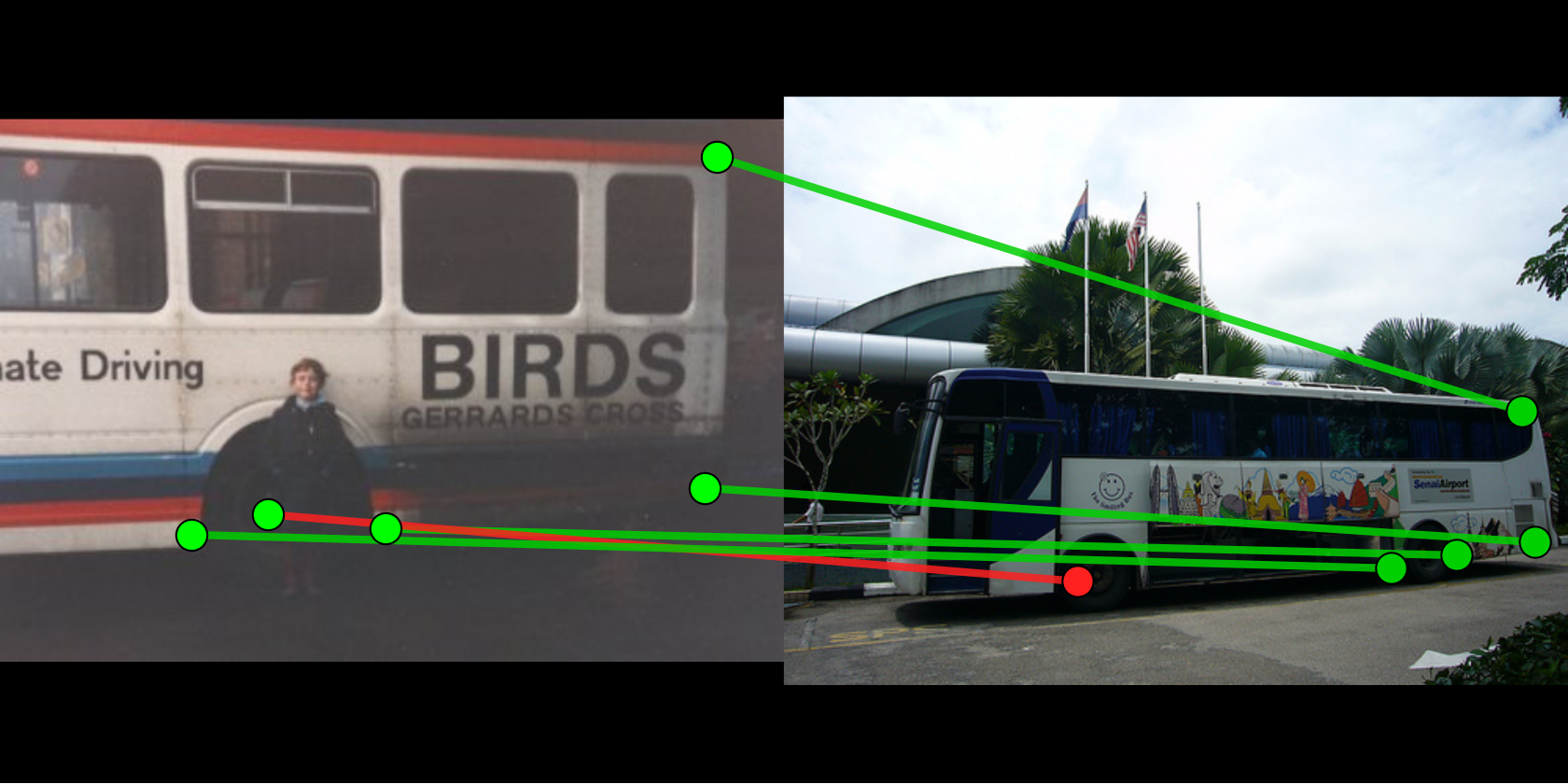} \\
\includegraphics[width=0.48\textwidth]{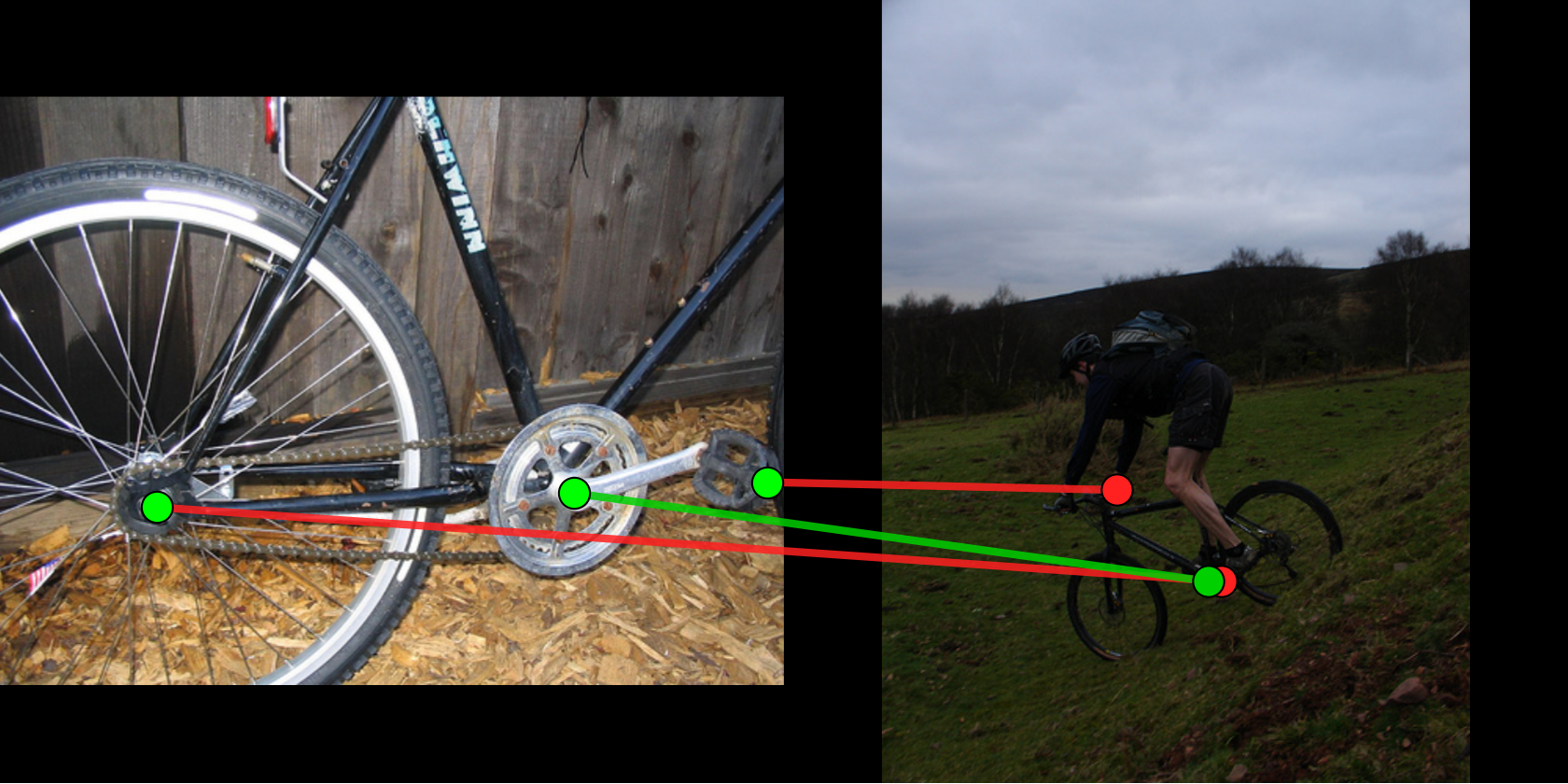} &
\includegraphics[width=0.48\textwidth]{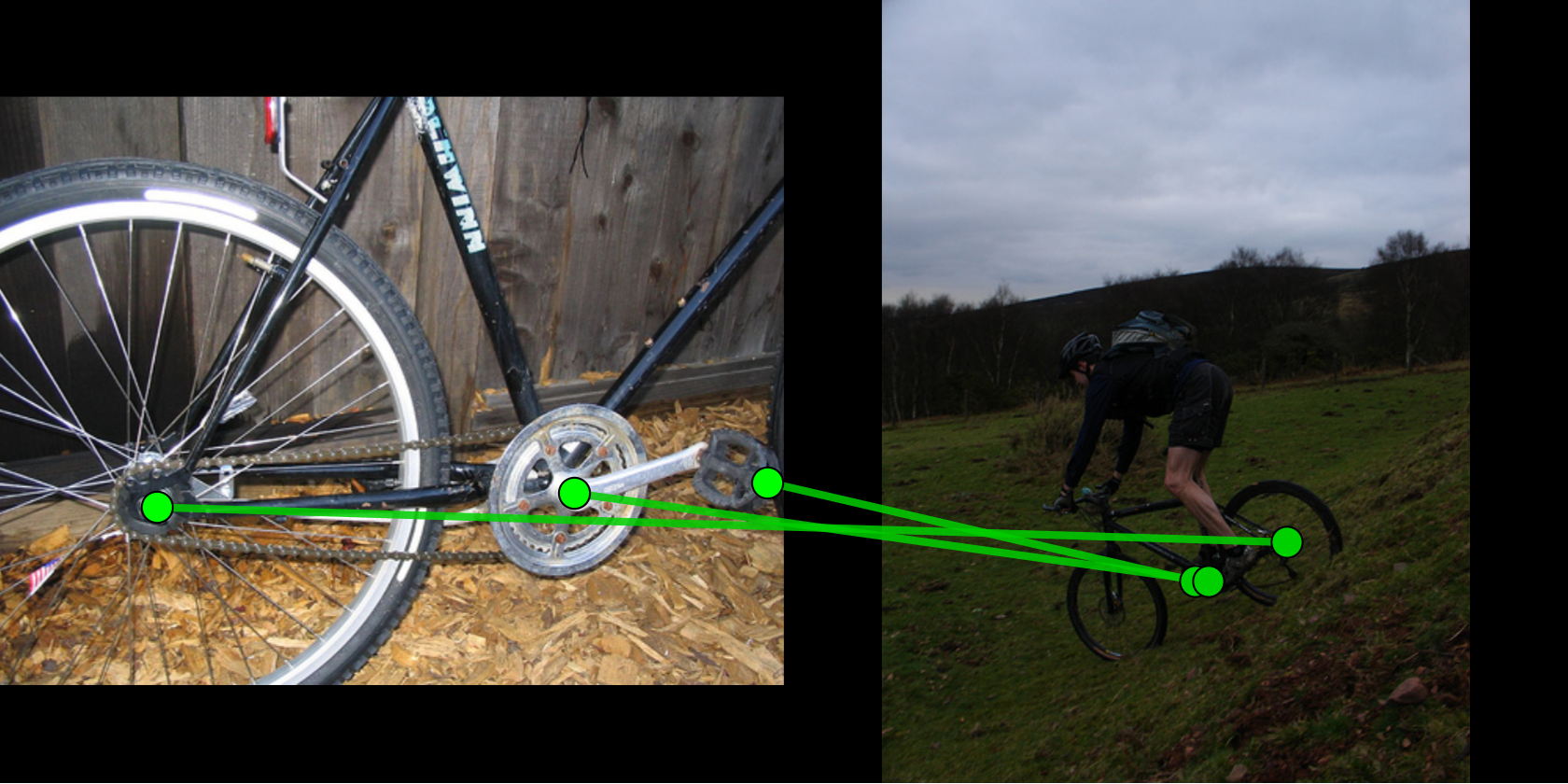} \\
\end{tabular}
\caption{\textbf{Qualitative SPair-71k \cite{min2019spair71k} results for GeoAware-SC \cite{zhang2024tellingLeftFromRight}.}
Each image shows a source--target pair with predicted correspondences. We compare GeoAware-SC trained on SPair-71k against GeoAware-SC pretrained on our generated correspondences and then fine-tuned on SPair-71k.}
\label{fig:qual_geoaware}
\end{figure*}

We provide qualitative SPair-71k \cite{min2019spair71k} examples comparing models trained only on real SPair-71k annotations with the same models pretrained on our generated synthetic correspondences and then fine-tuned on SPair-71k. We show results for both MARCO \cite{cuttano2026marco} (see \autoref{fig:qual_marco}) and GeoAware-SC \cite{zhang2024tellingLeftFromRight} (see \autoref{fig:qual_geoaware}).

\clearpage
\subsection{Correspondence Examples on New Categories}

\begin{figure*}[h]
\centering
\includegraphics[width=0.92\linewidth,height=0.22\textheight,keepaspectratio,trim={100 150 100 40},clip]{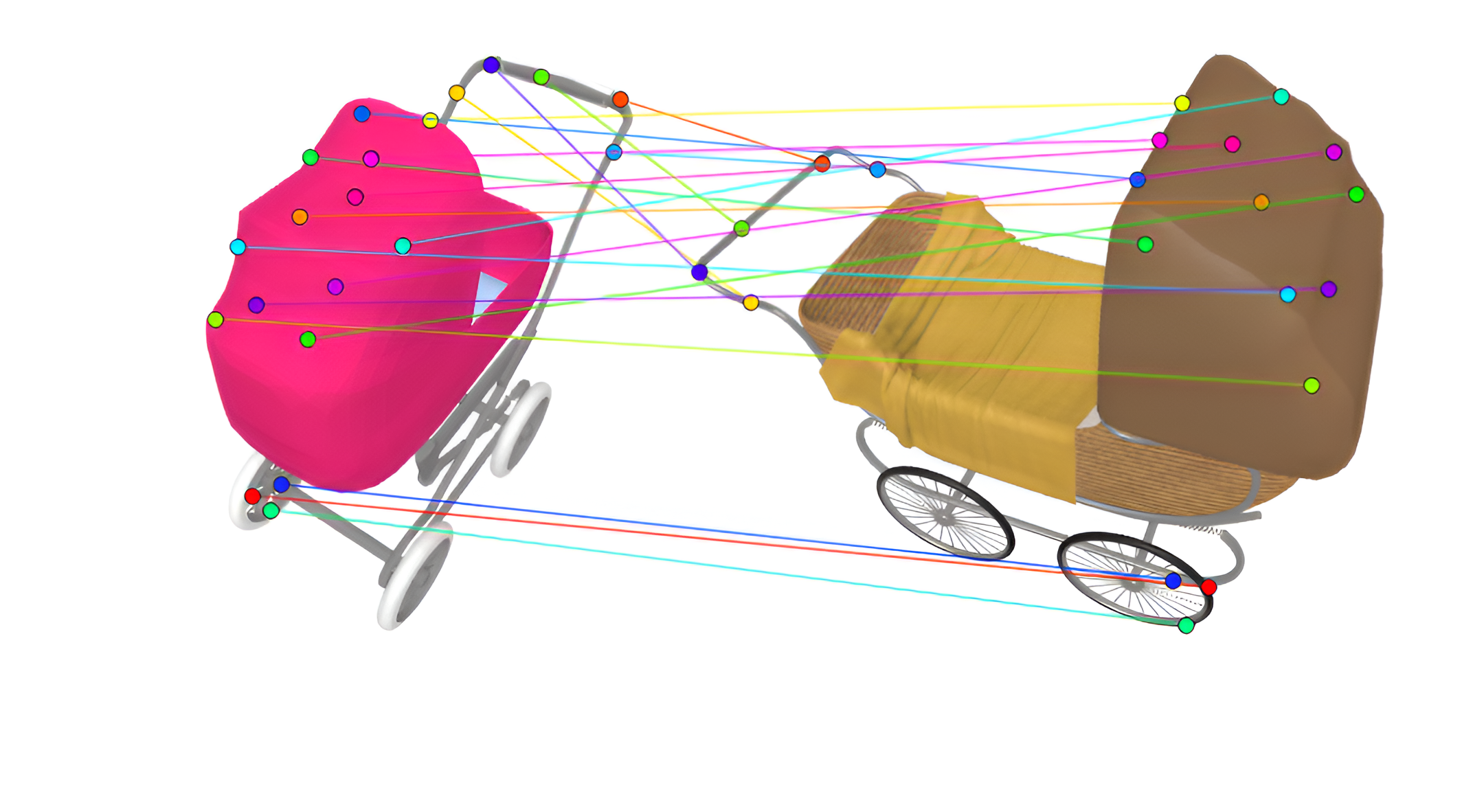}\vspace{1mm}

\includegraphics[width=0.92\linewidth,height=0.22\textheight,keepaspectratio,trim={0 20 70 40},clip]{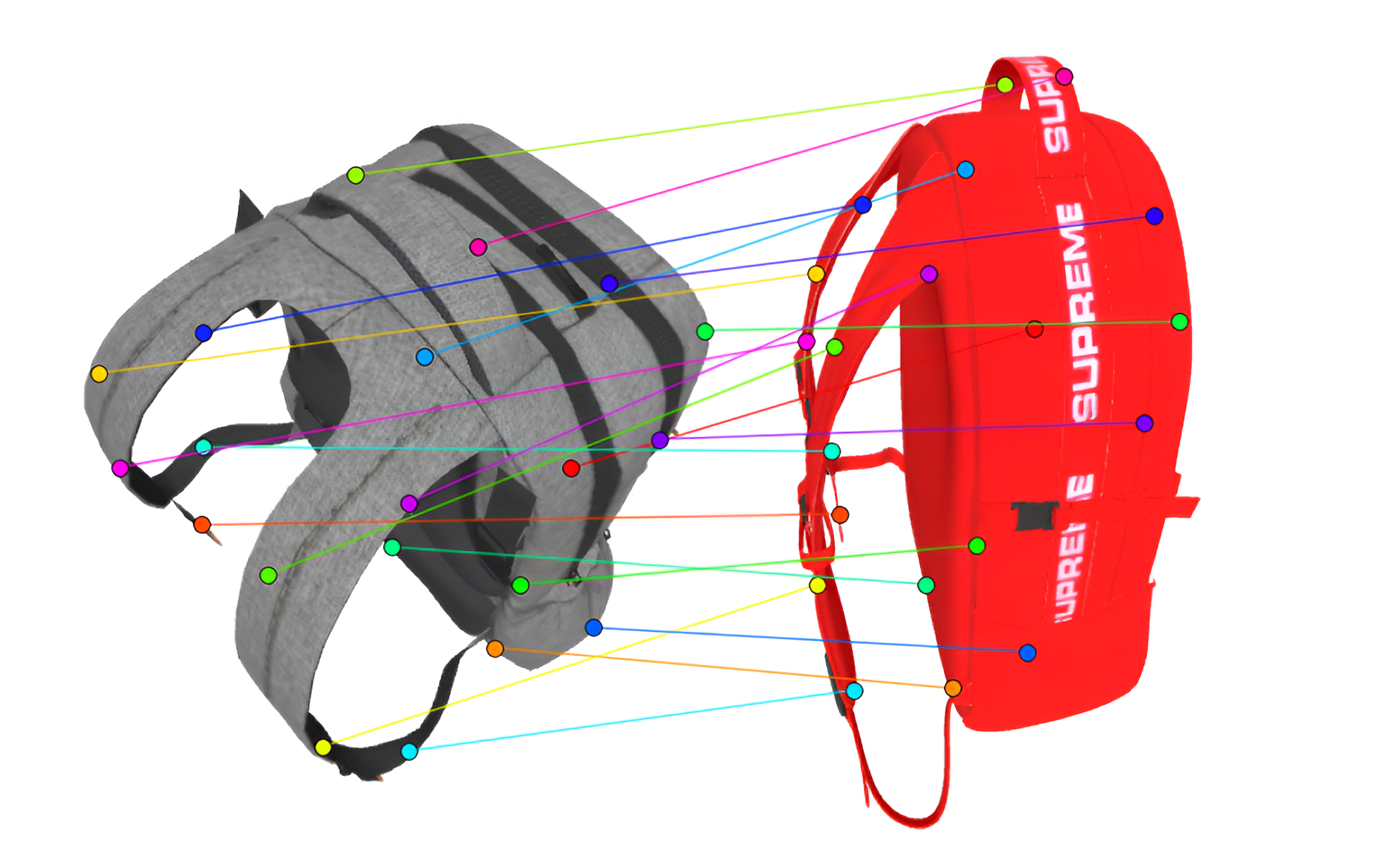}\vspace{1mm}

\includegraphics[width=0.92\linewidth,height=0.22\textheight,keepaspectratio,trim={0 45 50 0},clip]{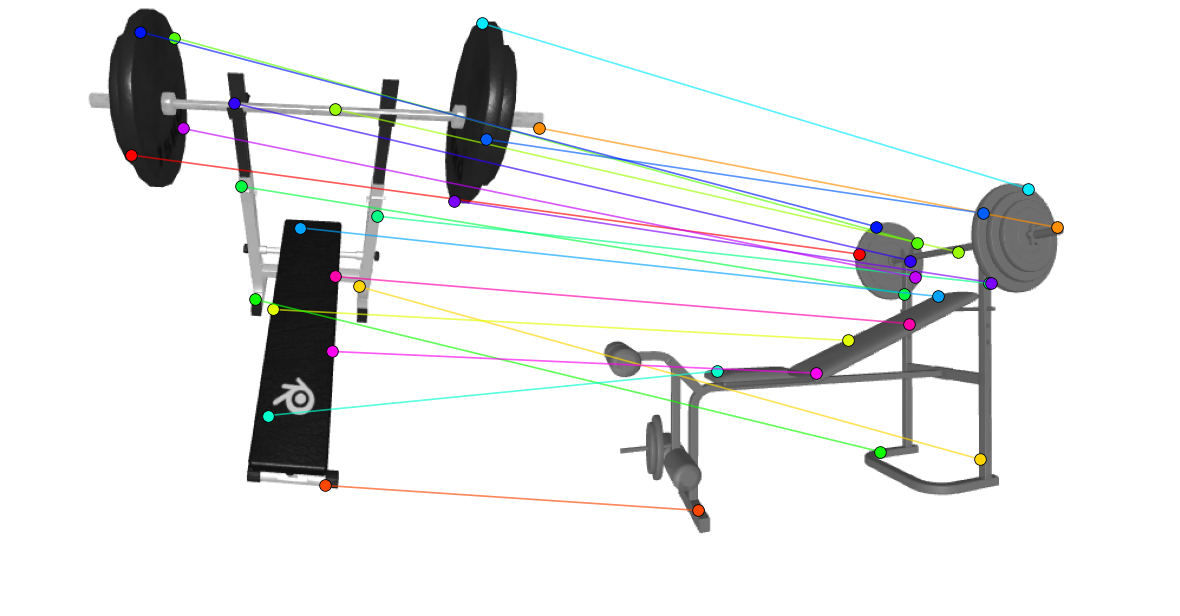}

\caption{
Qualitative correspondence examples on categories beyond existing correspondence benchmarks. Colors indicate sampled semantic-geometric correspondences established by our method.
}
\label{fig:new_categories_examples}
\end{figure*}

We additionally show (\autoref{fig:new_categories_examples}) qualitative correspondence examples on rigid categories beyond those covered by existing correspondence benchmarks such as SPair-71k \cite{min2019spair71k}, KeypointNet \cite{you2020}, and AP-10K \cite{yu2021AP10k}. These examples demonstrate the ability of our method to establish approximate semantic-geometric correspondences on diverse categories from Objaverse-OA \cite{lu2025orientationmatters}.

\clearpage
\subsection{Qualitative Results on KeypointNet}

\begin{figure}[h]
    \centering

    \begin{subfigure}[b]{0.48\linewidth}
        \centering
        \includegraphics[width=\linewidth]{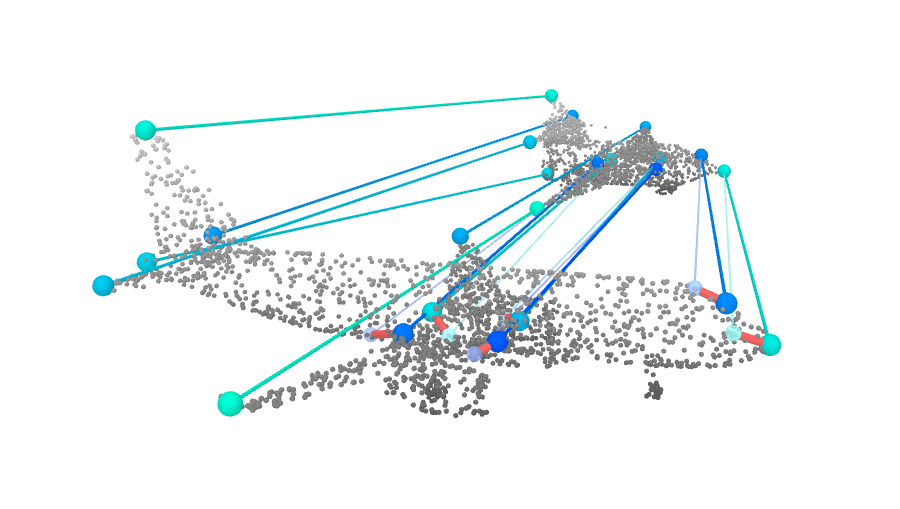}
        \caption{Airplanes}
        \label{fig:example_a}
    \end{subfigure}
    \hfill
    \begin{subfigure}[b]{0.48\linewidth}
        \centering
        \includegraphics[width=\linewidth]{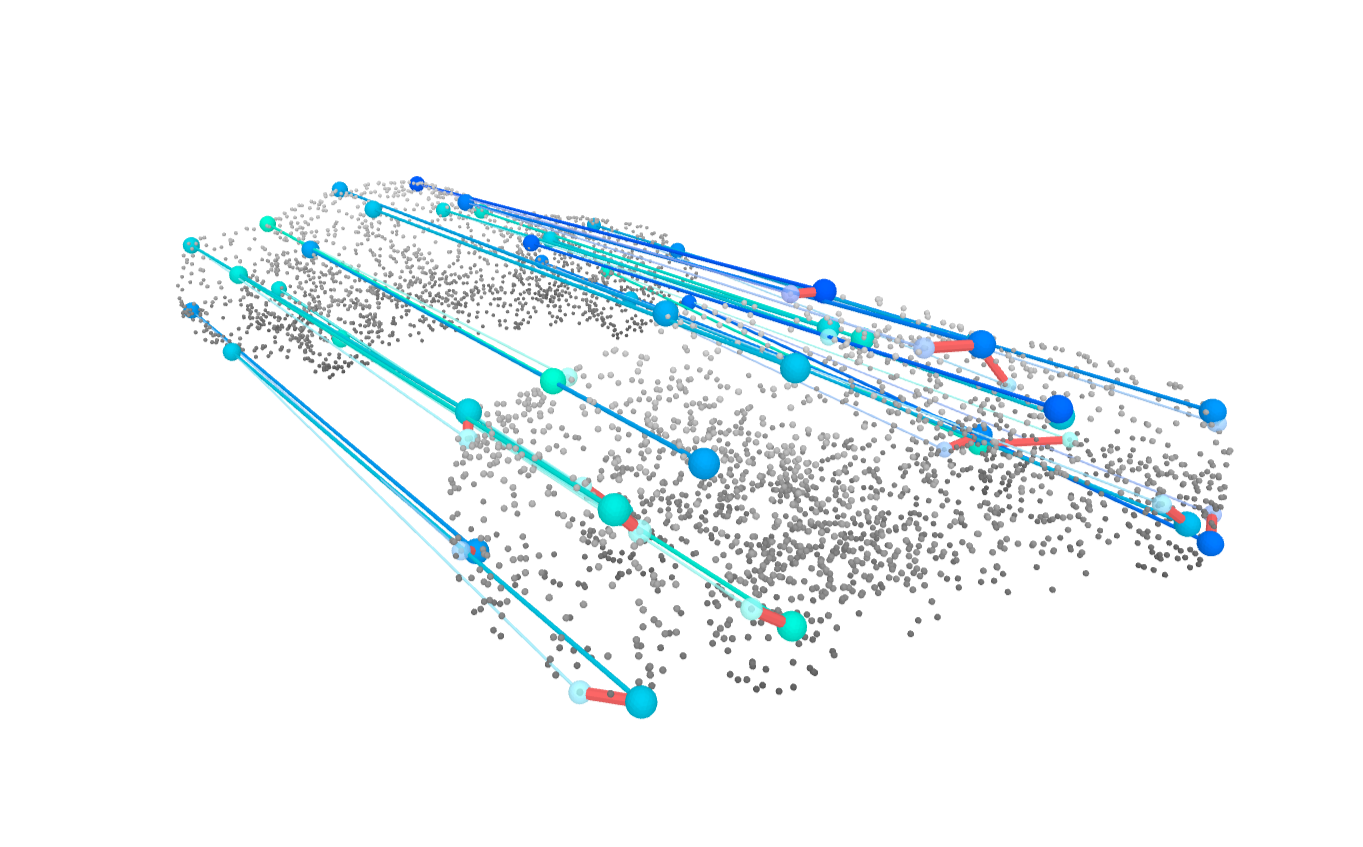}
        \caption{Cars}
        \label{fig:example_b}
    \end{subfigure}

    \caption{\textbf{Qualitative correspondence results on KeypointNet \cite{you2020}}
    We compare our predicted correspondences with human-annotated keypoints on the target object. Predicted correspondences are shown in lighter colors, while human annotations are shown in darker colors. Red lines connect each prediction to its corresponding ground-truth annotation, visualizing the localization difference between them.}
    \label{fig:keypointnet_examples}
\end{figure}

We provide qualitative correspondence results on KeypointNet \cite{you2020}, comparing our predicted correspondences against sparse human-annotated semantic keypoints. For each object pair, we use the human-annotated keypoints on the source object as queries and predict the corresponding points on the target object. As shown in \autoref{fig:keypointnet_examples}, our predictions generally align well with the human annotations across different categories. Predicted correspondences are shown in lighter colors, while the human annotations are shown in darker colors. Red lines connect each prediction to its corresponding ground-truth annotation.

\begin{figure}[h]
    \centering

    \begin{subfigure}[b]{0.45\linewidth}
        \centering
        \includegraphics[trim={10 350 10 400},clip,width=\linewidth,height=\linewidth,keepaspectratio]{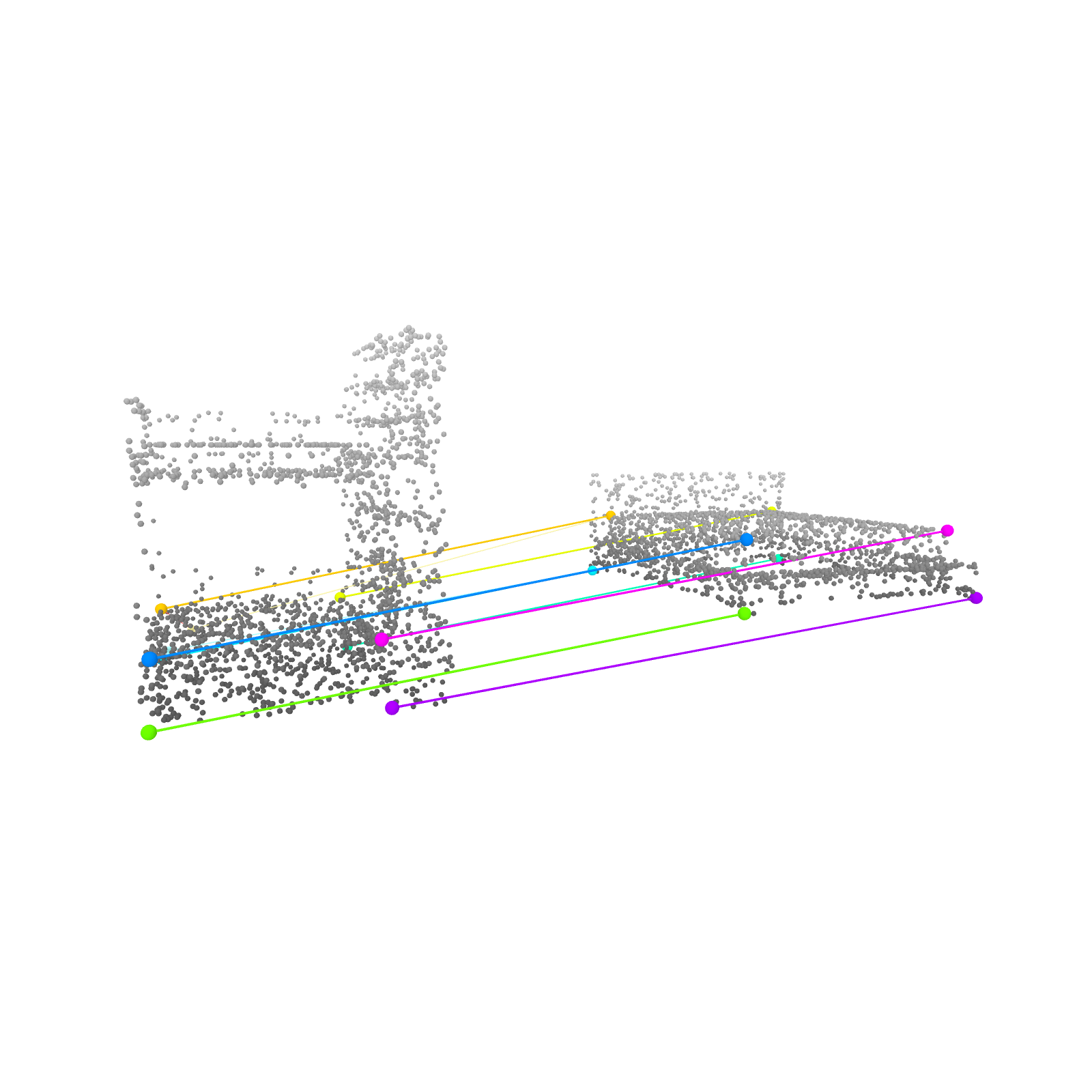}
        \caption{Bed}
        \label{fig:square_bed}
    \end{subfigure}
    \hfill
    \begin{subfigure}[b]{0.45\linewidth}
        \centering
        \includegraphics[trim={10 150 10 400},clip,width=\linewidth,height=\linewidth,keepaspectratio]{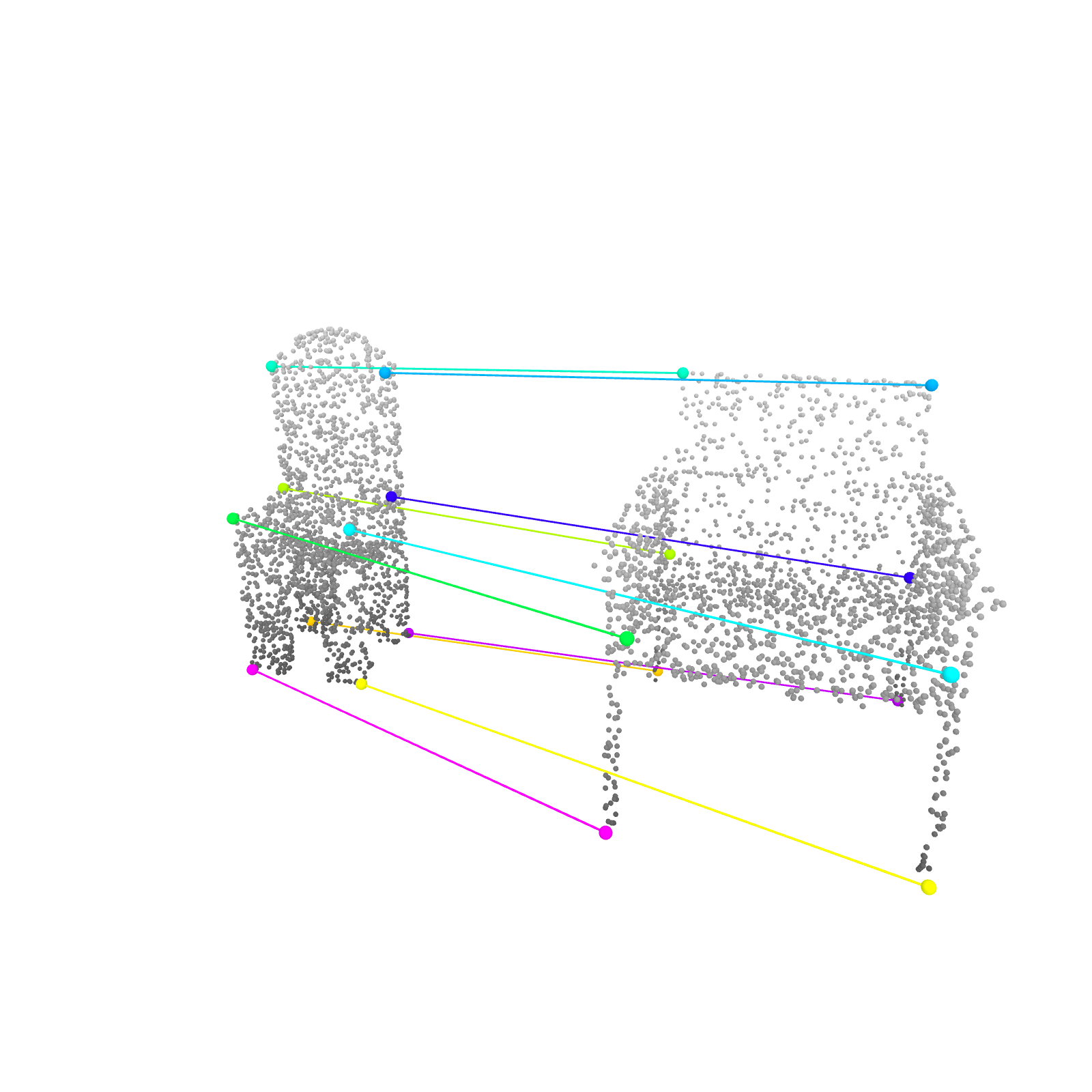}
        \caption{Chair}
        \label{fig:square_chair}
    \end{subfigure}

    \begin{subfigure}[b]{0.45\linewidth}
        \centering
        \includegraphics[trim={10 350 10 400},clip,width=\linewidth,height=\linewidth,keepaspectratio]{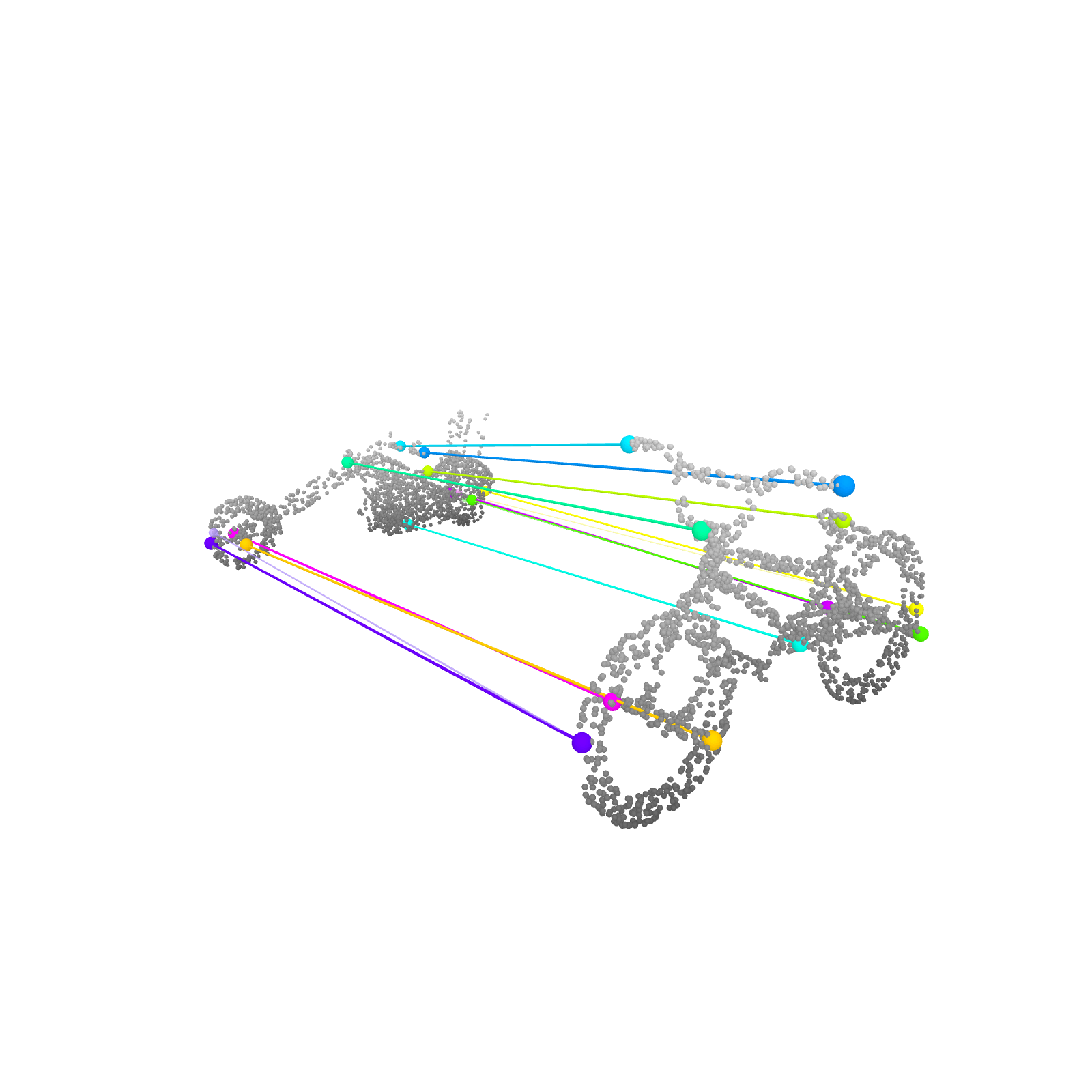}
        \caption{Motorcycle}
        \label{fig:square_motorcycle}
    \end{subfigure}
    \begin{subfigure}[b]{0.45\linewidth}
        \centering
        \includegraphics[trim={10 550 10 400},clip,width=\linewidth,height=\linewidth,keepaspectratio]{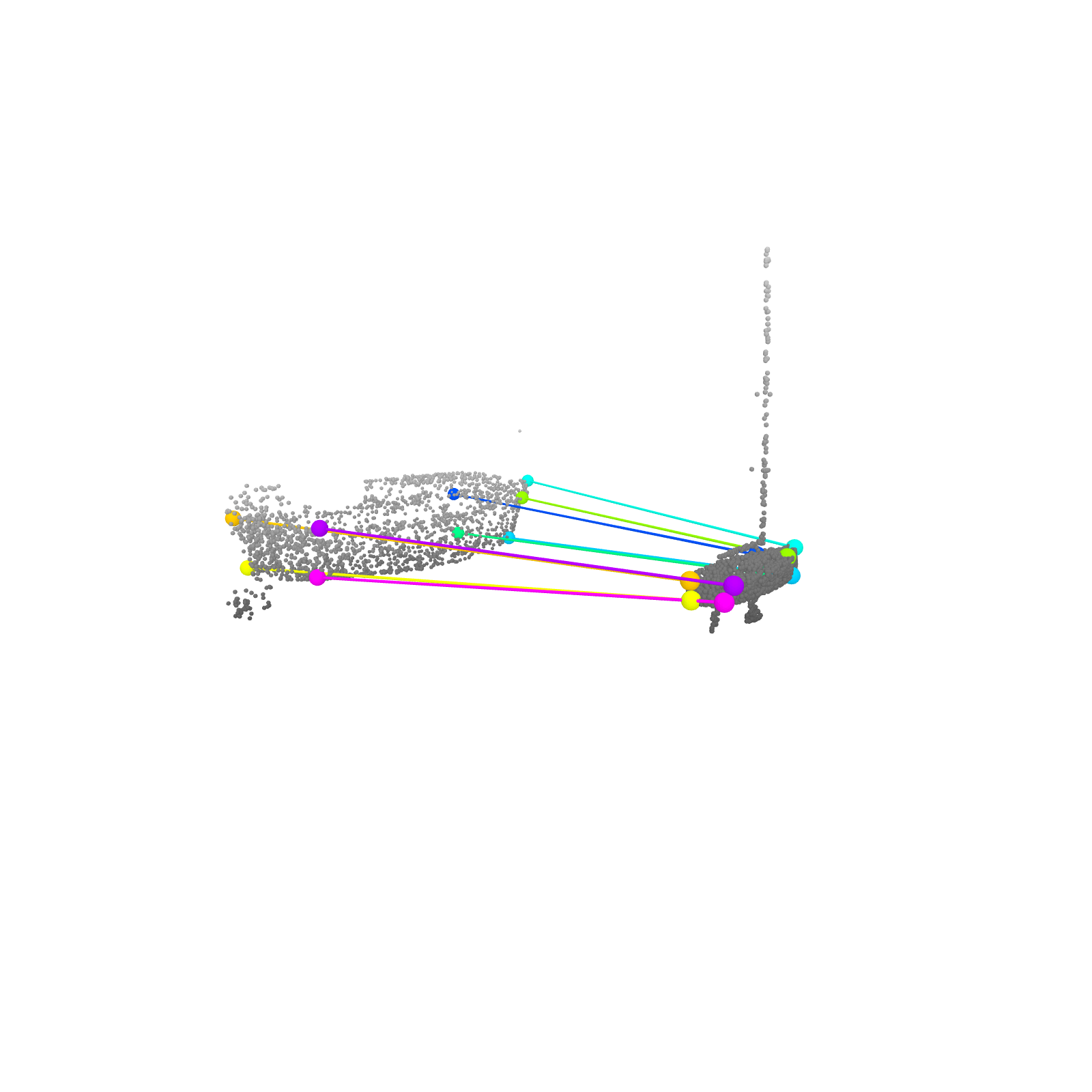}
        \caption{Vessel}
        \label{fig:square_vessel}
    \end{subfigure}

    \caption{\textbf{KeypointNet \cite{you2020} examples with strong annotation agreement.}
    Predicted correspondences are shown in lighter colors, while human annotations are shown in darker colors. The strong overlap between the two indicates that our pipeline recovers correspondences that are highly consistent with human semantic annotations.}
    \label{fig:keypointnet_strong_overlap}
\end{figure}

We further show examples with strong agreement in \autoref{fig:keypointnet_strong_overlap}. The close overlap between our predictions and the human annotations indicates that our generated correspondences are highly consistent with the sparse semantic keypoints used for evaluation.

\clearpage

\section{Additional Quantitative Results}
\label{sup:additional_quant}

We provide additional quantitative evaluations on SPair-71k \cite{min2019spair71k}
and KeypointNet \cite{you2020}. On SPair-71k, we examine different PCK
thresholds and training strategies, highlighting the
effectiveness of synthetic pretraining across architectures
and feature representations. On KeypointNet, we compare
our generated correspondences against the few-shot
semantic keypoint detector Back to 3D.

\subsection{Additional Results on SPair-71k}
\label{sup:supp_spair}
We provide additional evaluation results on SPair-71k \cite{min2019spair71k} across multiple PCK thresholds and training regimes. While the main paper reports PCK@0.10 (see \autoref{tab:spair_pck010}), here we analyze performance at stricter and looser thresholds, and compare four training strategies: \emph{Synth-only}, \emph{Both} (joint real and synthetic training), \emph{Real-only}, and \emph{Synth$\rightarrow$Real} (pretraining on synthetic correspondences followed by fine-tuning on real data).

\paragraph{MARCO \cite{cuttano2026marco} results.}
Across most thresholds (PCK@0.05--0.20), we observed that the Synth$\rightarrow$Real setting consistently outperforms Real-only training at practical thresholds. Improvements of approximately $+2.4$ to $+3.0$ PCK are observed for all PCK@0.05--0.20, and specifically at PCK@0.10, performance improves from $86.34$ to $89.30$ ($+2.96$), matching the results reported in \autoref{tab:spair_pck010}.

However, at a very strict threshold (PCK@0.01), Synth$\rightarrow$Real slightly underperforms Real-only ($-2.17$). However, this gap disappears at more relaxed thresholds, where Synth$\rightarrow$Real consistently achieves the best performance.

We also observe that joint training (\emph{Both}) underperforms Real-only, indicating that naive mixing of synthetic and real data is suboptimal, and that a two-stage training strategy is more effective.

\paragraph{GeoAware-SC \cite{zhang2024tellingLeftFromRight} results.}
We observe the similiar qualitative trend for GeoAware-SC. Synth$\rightarrow$Real consistently improves over Real-only training across all evaluated thresholds (including improvement at the very strict threshold of PCK@0.01), while Synth-only performs poorly in isolation. Joint training slightly outperforms Real-only, but remains inferior to synthetic pretraining followed by real-data fine-tuning, reinforcing that synthetic correspondences are most effective as a pretraining signal rather than as direct joint supervision.

At PCK@0.10, as we see in the results from the main paper (\autoref{tab:spair_pck010}), synthetic pretraining improves performance from $79.36$ to $83.73$, corresponding to a gain of $+4.37$ PCK.

\paragraph{Cross-architecture consistency.}
Using the PCK@0.10 results from \autoref{tab:spair_pck010}, we observe consistent improvements across architectures and feature choices:
\begin{center}
\begin{tabular}{lccccc}
\toprule
Architecture & DINOv2 \cite{oquab2024dinov2} & Stable Diffusion \cite{rombach2022stablediffusion} & Real-only & Synth$\rightarrow$Real & Gain \\
\midrule
GeoAware-SC \cite{zhang2024tellingLeftFromRight} & \checkmark & \checkmark & 79.36 & 83.73 & +4.37 \\
MARCO \cite{cuttano2026marco} & \checkmark & $\times$ & 86.34 & 89.30 & +2.96 \\
\bottomrule
\end{tabular}
\end{center}

GeoAware-SC \cite{zhang2024tellingLeftFromRight} uses both DINOv2 \cite{oquab2024dinov2} and Stable Diffusion \cite{rombach2022stablediffusion} features, while MARCO \cite{cuttano2026marco} relies on DINOv2 features without Stable Diffusion. The gains across both settings suggest that our synthetic pretraining signal is not tied to a specific feature family, but provides useful semantic-geometric supervision across different recent correspondence models.

These consistent gains suggest that the benefit of synthetic pretraining is largely architecture-agnostic. Rather than improving a specific model design, the generated correspondences provide a strong pretraining signal that improves representation quality across different backbones and matching heads.

\paragraph{Summary.}
Overall, these results support three key observations: (i) synthetic correspondences alone are insufficient for high performance, (ii) naive joint training is suboptimal, and (iii) a two-stage Synth$\rightarrow$Real strategy consistently yields the best results. This pattern holds across both architectures at practical evaluation thresholds, reinforcing the generality of the proposed data generation pipeline.

\subsection{Comparison with Back to 3D on KeypointNet}
\label{sup:backto3d}

We additionally compare against Back to 3D~\cite{wimmer2024back},
a few-shot semantic keypoint detection method rather than a
cross-instance correspondence method. Unlike SemGeo-Gen, it
predicts predefined semantic keypoints using annotated source
examples instead of generating dense correspondences without
keypoint supervision. We evaluate its 1-shot and 3-shot settings
on 100 object pairs per category (1,600 total).
The annotated source keypoints define semantic template slots,
and source-to-target associations are established through their
shared KeypointNet labels. Predicted target locations are then
evaluated using the same PCK thresholds as our correspondences.

As shown in \autoref{tab:backto3d_comparison},
SemGeo-Gen achieves 86.5\% PCK@0.10, outperforming
Back to 3D's 1-shot (63.2\%) and 3-shot (64.1\%)
settings, despite requiring no keypoint supervision.

\begin{table}[h]
\centering
\small
\setlength{\tabcolsep}{4pt}
\caption{\textbf{Comparison with Back to 3D \cite{wimmer2024back} on KeypointNet \cite{you2020}}
Mean PCK over 16 categories (100 pairs per category).
Back to 3D uses few-shot keypoint supervision;
SemGeo-Gen requires none.}
\label{tab:backto3d_comparison}
\vspace{4pt}
\begin{tabular}{lcccc}
\toprule
Method & @0.05 & @0.10 & @0.15 & @0.20 \\
\midrule
Back to 3D (1-shot) & 0.438 & 0.632 & 0.725 & 0.777 \\
Back to 3D (3-shot) & 0.436 & 0.641 & 0.740 & 0.799 \\
\textbf{SemGeo-Gen} & \textbf{0.648} & \textbf{0.865}
& \textbf{0.945} & \textbf{0.977} \\
\bottomrule
\end{tabular}
\end{table}

\clearpage

\section{Extended Ablation Study}
\label{sup:ablation_study}

We provide additional ablation studies examining the
contributions of the registration model and semantic filtering,
as well as the sensitivity of our method to imperfect
canonical alignment. These experiments further characterize
the accuracy, efficiency, and robustness of our method.

\subsection{Registration Method Ablation}
\label{sup:ablation_registration}

We ablate the deformation model used in the registration stage prior to correspondence extraction. All other components are kept fixed, including lifted DINOv2 \cite{oquab2024dinov2} features, nearest-neighbor assignment, and the semantic threshold $\tau_{\mathrm{sem}}=0.575$. We compare our default continuous piecewise-affine (CPA) model with two alternatives: a global affine transformation and a thin-plate spline (TPS) deformation.

\paragraph{Affine.}
As a global transformation baseline, we use an affine transformation

\[
f(\mathbf{x}) = \mathbf{A}\mathbf{x} + \mathbf{t},
\]
where $\mathbf{A} \in \mathbb{R}^{3\times 3}$ and $\mathbf{t} \in \mathbb{R}^{3}$. This model has 12 learnable parameters and is optimized with Adam using the same Chamfer objective. Since it cannot model non-rigid variation, it serves as a baseline for how much alignment can be explained by global pose normalization alone.

\paragraph{Continuous Piecewise-Affine (CPA).}
Our default model uses a $6 \times 6 \times 6$ control grid (216 vertices) over the normalized 3D domain, tetrahedralized via Delaunay triangulation. Each vertex has a learnable displacement, and each point is warped using barycentric interpolation within its enclosing tetrahedron. This yields a deformation that is locally affine and globally continuous. We enforce smoothness by penalizing squared differences between neighboring vertex displacements, with $\lambda_{\mathrm{smooth}}=0.3$. The model has 648 degrees of freedom.

\paragraph{Thin-Plate Spline (TPS).}
We compare with a classical smooth non-rigid deformation
\[
f(\mathbf{x}) = \mathbf{a} + \mathbf{B}\mathbf{x}
+ \sum_{k=1}^{K} \mathbf{w}_k \, U(\|\mathbf{x}-\mathbf{c}_k\|),
\qquad U(r) = -r,
\]
where $\mathbf{B}\in\mathbb{R}^{3\times3}$, $\mathbf{a}\in\mathbb{R}^{3}$, and $\mathbf{w}_k\in\mathbb{R}^{3}$. We use $K=216$ control points sampled from the source, resulting in a similar number of parameters to CPA (660 vs.\ 648). Smoothness is enforced with an $\ell_2$ penalty on the radial weights.

\paragraph{Protocol.}
We evaluate all three methods on the 16 KeypointNet \cite{you2020} categories. For each category, we sample 100 source-target pairs from random instances and run all methods on them. All hyperparameters are shared: 10,000 Adam iterations with early stopping on a loss plateau, a learning rate of $0.005$, and no pair-level filtering. Performance is measured using PCK at multiple thresholds, where a prediction is correct if it falls within $\tau \cdot d$ of the ground-truth target keypoint, with $d$ the target bounding-box diagonal.

\paragraph{Results.}
\autoref{tab:ablation_registration} reports mean PCK across all categories. CPA achieves the best average performance at four of the five PCK thresholds, while TPS is slightly better only at PCK@0.05. Overall, CPA and TPS obtain very similar accuracy, indicating that the choice of non-rigid basis has limited impact on correspondence quality.

\begin{table}[h!]
\centering
\small
\caption{\textbf{Registration ablation on KeypointNet.}
Mean PCK across all 16 categories. PCK thresholds define the allowed normalized distance from the ground-truth keypoint, measured as a fraction of the target bounding-box diagonal. DOF denotes the number of learnable parameters. Time reports the mean registration time per pair on a single A6000 GPU. Best results are shown in \textbf{bold}.}
\label{tab:ablation_registration}
\setlength{\tabcolsep}{4pt}
\begin{tabular}{lccccccc}
\toprule
Method & DOF & Time (s/pair) & PCK@0.01 & PCK@0.05 & PCK@0.10 & PCK@0.15 & PCK@0.20 \\
\midrule
Affine            & 12  & 1.15 & 0.221 & 0.686 & 0.898 & 0.960 & 0.982 \\
TPS ($K\!=\!216$) & 660 & 3.71 & 0.226 & \textbf{0.739} & 0.919 & 0.966 & 0.983 \\
CPA (ours)        & 648 & \textbf{1.13} & \textbf{0.236} & 0.729 & \textbf{0.922} & \textbf{0.971} & \textbf{0.986} \\
\bottomrule
\end{tabular}
\end{table}

In contrast, CPA is substantially more efficient. It is about $3.3\times$ faster per pair (1.13s vs.\ 3.71s), despite having a similar number of parameters. We also observe that CPA typically converges earlier. This difference arises from the underlying structure: CPA uses local barycentric interpolation within tetrahedra, while TPS requires dense kernel interactions between all points and control points at each iteration.

Compared with the affine baseline, CPA provides clear gains, especially at tighter PCK thresholds, highlighting the importance of non-rigid alignment for cross-instance correspondence.

\paragraph{Conclusion.}
This ablation supports two conclusions: (i) non-rigid deformation is essential for accurate alignment, as affine transformations perform significantly worse; and (ii) among non-rigid models, CPA offers a better accuracy-efficiency tradeoff, achieving similar performance to TPS while being significantly faster and more stable in practice.

\subsection{Ablation on Semantic Filtering}
\label{sup:ablation_semantic_filtering}

We analyze the role of semantic filtering in our pipeline. Specifically, we study (i) the effect of removing semantic filtering entirely, (ii) the contribution of correspondence-level filtering alone, and (iii) the impact of the correspondence-level semantic threshold $\tau_{\mathrm{sem}}$ when combined with pair-level filtering. All experiments are conducted on KeypointNet using the same evaluation protocol as in the main paper.

\subsubsection{No Semantic Filtering}
\label{sup:no_semantic_filtering}

We first evaluate the pipeline without any semantic filtering. In this setting, all object pairs are retained, and correspondences are obtained directly from CPA registration followed by nearest-neighbor assignment.

\autoref{tab:unfiltered} shows that this setting produces significantly lower accuracy. While geometric alignment alone can recover many plausible correspondences, it frequently matches inconsistent regions across instances, leading to degraded PCK, especially at stricter thresholds (e.g., PCK@0.05 and PCK@0.10). This highlights the limitation of relying solely on geometry for cross-instance correspondence and the need for a semantic pruning step.

\begin{table}[h]
  \caption{Evaluation of annotation quality on KeypointNet correspondences without filters. We report PCK at normalized distance thresholds of $5\%$, $10\%$, $15\%$, and $20\%$ of the object scale, computed per category. (on 100 objects)}
  \label{tab:unfiltered}

  \centering
  \begin{tabular}{lcccc}
    \toprule
    Category & PCK@0.05 $\uparrow$ & PCK@0.10 $\uparrow$ & PCK@0.15 $\uparrow$ & PCK@0.20 $\uparrow$ \\
    \midrule
    Airplane   & 0.73 & 0.90 & 0.95 & 0.98 \\
    Bathtub    & 0.41 & 0.73 & 0.91 & 0.97 \\
    Bed        & 0.45 & 0.71 & 0.84 & 0.90 \\
    Bottle     & 0.65 & 0.83 & 0.92 & 0.98 \\
    Car        & 0.75 & 0.94 & 0.97 & 0.99 \\
    Chair      & 0.51 & 0.83 & 0.93 & 0.97 \\
    Guitar     & 0.78 & 0.94 & 0.98 & 0.99 \\
    Laptop     & 0.87 & 0.99 & 1.00 & 1.00 \\
    Motorcycle & 0.69 & 0.88 & 0.95 & 0.99 \\
    Mug        & 0.70 & 0.95 & 0.99 & 1.00 \\
    Skateboard & 0.70 & 0.87 & 0.96 & 0.99 \\
    Table      & 0.63 & 0.86 & 0.94 & 0.97 \\
    Vessel     & 0.38 & 0.66 & 0.82 & 0.89 \\
    Knife     & 0.52 & 0.68 & 0.82 & 0.91 \\ 
    Helmet & 0.36 & 0.69 & 0.86 & 0.93 \\ 
    Cap & 0.74 & 0.92 & 0.99 & 1.0\\
    \midrule
    Mean   & 0.62 & 0.84 & 0.93 & 0.97 \\
    Median & 0.67 & 0.86 & 0.94 & 0.98 \\
    \bottomrule
  \end{tabular}

\end{table}

\subsubsection{Correspondence-Level Semantic Filtering Only}
\label{sup:only_corr_filtering}

We next evaluate correspondence-level semantic filtering alone, without pair-level filtering. In this setting, all object pairs are retained, and individual correspondences are pruned using a fixed threshold $\tau_{\mathrm{sem}}=0.575$.

\autoref{tab:pck_results} shows that this step alone leads to a substantial improvement over the no-filtering baseline. In particular, mean PCK@0.10 improves from 0.84 to 0.91, confirming that a large portion of the error arises from semantically inconsistent matches that are geometrically plausible but incorrect.

\begin{table}[t]
  \centering
  \caption{Ablation of only semantic filtering of correspondences, reporting PCK results per category. calculated over 100 models.}
  \label{tab:pck_results}
  \begin{tabular}{lcccc}
    \toprule
    Category & PCK@0.05 & PCK@0.10 & PCK@0.15 & PCK@0.20 \\
    \midrule
    Airplane & 0.78 & 0.94 & 0.98 & 0.99 \\
    Bathtub & 0.68 & 0.86 & 0.94 & 0.97 \\
    Bed & 0.63 & 0.85 & 0.94 & 0.98 \\
    Bottle & 0.80 & 0.95 & 0.97 & 0.99 \\
    Car & 0.77 & 0.95 & 0.98 & 1.00 \\
    Chair & 0.71 & 0.94 & 0.98 & 0.99 \\
    Guitar & 0.80 & 0.96 & 0.98 & 0.99 \\
    Laptop & 0.89 & 0.99 & 1.00 & 1.00 \\
    Table & 0.74 & 0.95 & 0.99 & 1.00 \\
    Mug & 0.70 & 0.95 & 0.99 & 1.00 \\
    Skateboard & 0.86 & 0.97 & 0.99 & 1.00 \\
    Vessel & 0.42 & 0.68 & 0.84 & 0.93 \\
    Motorcycle & 0.75 & 0.93 & 0.98 & 0.99 \\
    Cap & 0.80 & 0.96 & 1.00 & \textbf{1.00} \\
    Knife & 0.86 & 0.93 & 0.97 & 0.98 \\
    Helmet & 0.46 & 0.77 & 0.91 & 0.95 \\
    \midrule
     Mean & 0.73 & 0.91 & 0.97 & 0.99 \\
    Median & 0.76 & 0.94 & 0.98 & 0.99 \\
    \bottomrule
  \end{tabular}
\end{table}

\subsubsection{Correspondence-Level Semantic Threshold}
\label{sup:ablation_semantic_threshold_corr}

We finally analyze the effect of the semantic threshold $\tau_{\mathrm{sem}}$ when combined with pair-level filtering, as used in the full pipeline.

This threshold controls an accuracy-coverage tradeoff. A stricter threshold keeps fewer but more reliable matches, while a looser threshold retains more matches but allows more semantically inconsistent correspondences. To quantify coverage, we report the \emph{median GT-used rate} across categories. A ground-truth keypoint is counted as used if at least one surviving correspondence remains within $0.05$ of the object bounding-box diagonal from the ground-truth location.

Importantly, this metric is conservative. KeypointNet annotations are sparse and typically located on salient corners or extreme parts, while our method produces dense correspondences across the object surface (see Fig.~1). As a result, semantically valid correspondences that do not fall within the $0.05$ evaluation radius are still counted as missed.

\begin{table}[h!]
\centering
\small
\caption{\textbf{Ablation on the correspondence-level semantic threshold (with pair-level filtering).}
Results are averaged over three random seeds and 16 KeypointNet categories. Median GT-used rate denotes the median fraction across categories of ground-truth keypoints that remain evaluable after correspondence-level pruning.}
\label{tab:semantic_threshold_ablation}
\setlength{\tabcolsep}{5pt}
\begin{tabular}{lccccc}
\toprule
$\tau_{\mathrm{sem}}$ & PCK@0.05 $\uparrow$ & PCK@0.10 $\uparrow$ & PCK@0.15 $\uparrow$ & PCK@0.20 $\uparrow$ & Median GT-used rate $\uparrow$ \\
\midrule
0.75  & 0.758 & 0.930 & 0.974 & 0.988 & 0.260 \\
0.575 & 0.749 & 0.926 & 0.973 & 0.989 & 0.576 \\
0.25  & 0.699 & 0.896 & 0.960 & 0.983 & 0.915 \\
\bottomrule
\end{tabular}
\end{table}

\autoref{tab:semantic_threshold_ablation} shows that the choice of threshold controls a clear tradeoff between
accuracy and coverage. A strict threshold ($\tau_{\mathrm{sem}}=0.75$) achieves the highest accuracy at tight
thresholds (75.8\% PCK@0.05) but retains only 26.0\% of ground-truth keypoints, while a loose threshold
($\tau_{\mathrm{sem}}=0.25$) retains 91.5\% but degrades accuracy (69.9\% PCK@0.05, 89.6\% PCK@0.10).
Our default ($\tau_{\mathrm{sem}}=0.575$) retains 57.6\% of the keypoints with almost no loss in accuracy
compared to the strict setting.

We therefore use $\tau_{\mathrm{sem}}=0.575$ as a practical middle point: it maintains similar accuracy to the stricter threshold while more than doubling the median GT-used rate (26.0\% $\rightarrow$ 57.6\%).

\paragraph{Discussion.}
Overall, these results show that semantic filtering is essential for removing geometrically plausible but semantically incorrect correspondences. Most of the improvement comes from correspondence-level filtering, while pair-level filtering provides an additional refinement by removing object pairs with insufficient semantic overlap.

\subsection{Sensitivity to Rotation Misalignment}
\label{sup:rotation_sensitivity}

We evaluate the robustness of SemGeo-Gen to imperfect
canonical alignment. For each of the 16 KeypointNet
categories, we sample 100 object pairs and randomly
select either the source or target object to perturb
with a random-axis rotation of $0^\circ$, $5^\circ$,
$10^\circ$, $30^\circ$, $60^\circ$, or $90^\circ$.
We measure correspondence accuracy using the same
KeypointNet evaluation protocol.

As shown in \autoref{tab:rotation_sensitivity},
mean PCK@0.10 decreases from 0.93 for aligned objects
to 0.89 and 0.85 at $5^\circ$ and $10^\circ$,
respectively. Performance drops substantially at
$30^\circ$ and beyond, confirming that approximate
canonical alignment remains an important prerequisite
for accurate correspondence generation.

\begin{table}[h]
    \centering
    \captionsetup{justification=centering}
    \fontsize{11}{13}\selectfont
    \setlength{\tabcolsep}{2pt}
    \renewcommand{\arraystretch}{1.05}

    \caption{
        \textbf{Sensitivity to rotation misalignment.}
        Mean and median KeypointNet PCK under random-axis
        rotations applied to either the source or target.
    }
    \label{tab:rotation_sensitivity}
    \vspace{4pt}

    \begin{tabular}{llcccccc}
        \toprule
        PCK & Stat.
        & $0^\circ$ & $5^\circ$ & $10^\circ$
        & $30^\circ$ & $60^\circ$ & $90^\circ$ \\
        \midrule
        \multirow{2}{*}{@0.05}
        & Mean   & .75 & .71 & .65 & .32 & .10 & .04 \\
        & Median & .78 & .75 & .69 & .25 & .06 & .02 \\
        \midrule
        \multirow{2}{*}{@0.10}
        & Mean   & .93 & .89 & .85 & .55 & .20 & .08 \\
        & Median & .94 & .91 & .88 & .52 & .13 & .05 \\
        \midrule
        \multirow{2}{*}{@0.15}
        & Mean   & .98 & .94 & .90 & .66 & .26 & .12 \\
        & Median & .98 & .95 & .91 & .64 & .20 & .07 \\
        \midrule
        \multirow{2}{*}{@0.20}
        & Mean   & .99 & .95 & .91 & .70 & .30 & .14 \\
        & Median & .99 & .95 & .91 & .67 & .26 & .09 \\
        \bottomrule
    \end{tabular}
\end{table}

\clearpage
\section{Qualitative Analysis of DINOv2 Based Semantic Filtering}
\label{sup:supp_semantic_analy}

\begin{figure}[h]
    \centering
    \includegraphics[width=\linewidth]{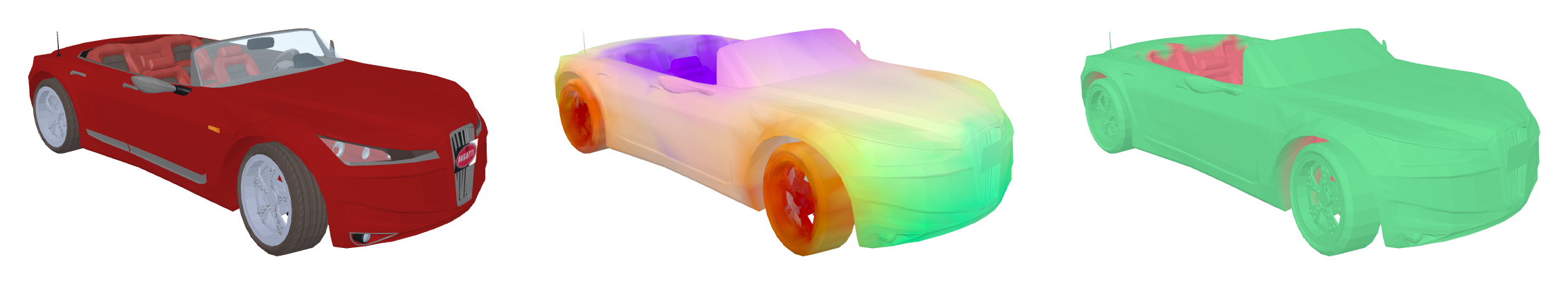}\\[-1mm]
    \includegraphics[width=\linewidth]{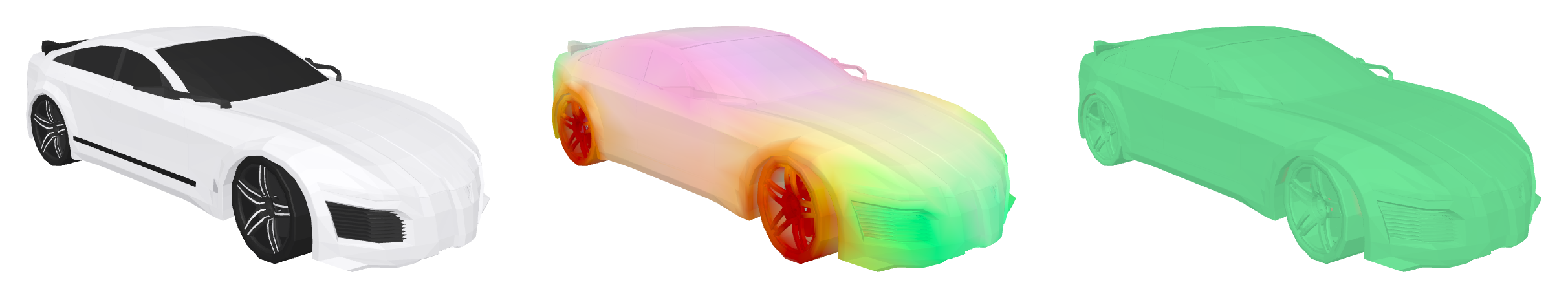}\\[1mm]
    \includegraphics[width=\linewidth]{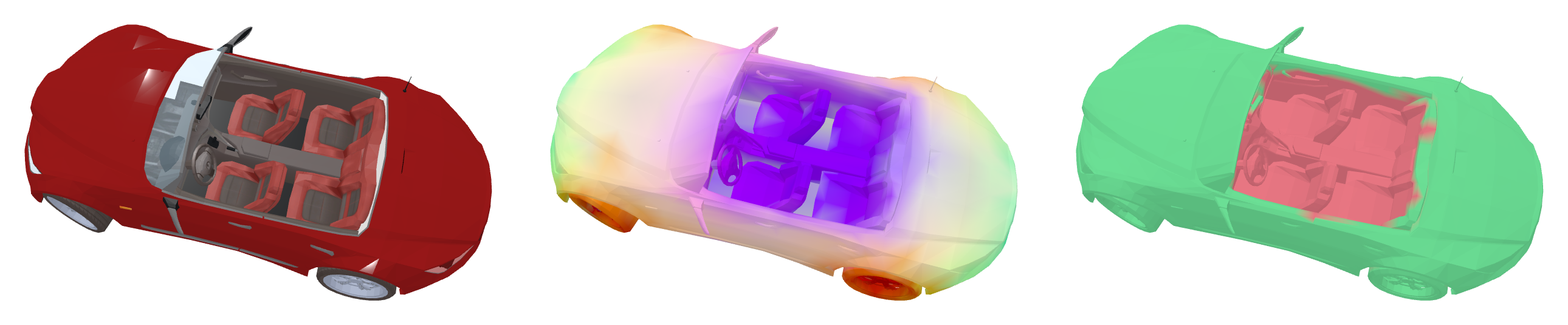}\\[-1mm]
    \includegraphics[width=\linewidth]{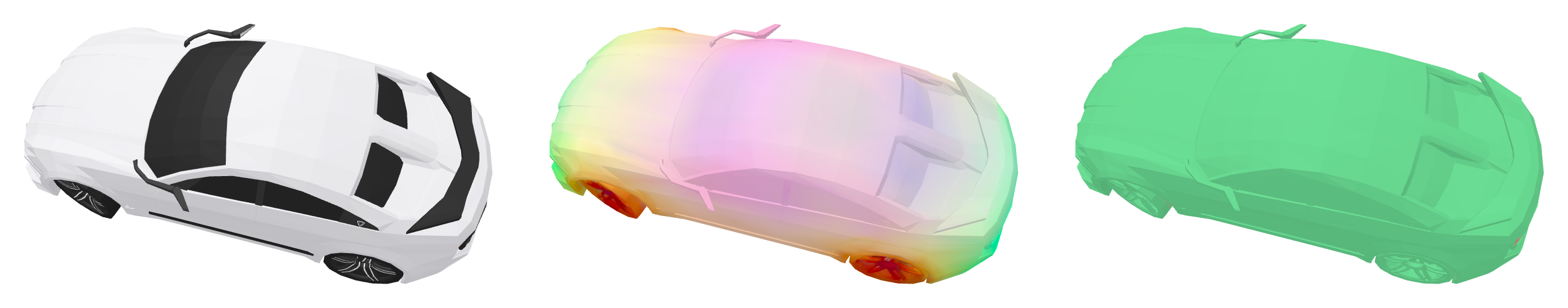}
\caption{
Qualitative analysis of DINOv2 based semantic coverage. Each row shows the original rendering, a PCA visualization of DINOv2 features computed jointly for the object pair, and the resulting semantic coverage mask. Green points have at least one semantically similar point in the paired object, while red points have none.
}
    
    \label{fig:semantic_coverage_analysis}
\end{figure}

We provide a qualitative analysis of the DINOv2 based semantic filtering used in our pipeline, both at the pair level and at the correspondence level. As shown in \autoref{fig:semantic_coverage_analysis}, we visualize the lifted DINOv2 features after joint PCA to three dimensions, and mark points according to whether they have any sufficiently similar semantic counterpart in the paired object. Green points have at least one point in the paired object with cosine similarity above our threshold of \(0.575\), while red points have no such candidate. We use this semantic coverage to filter object pairs that do not share enough semantic structure. In addition, for pairs that survive pair-level filtering, red points are guaranteed to be removed by our semantic pruning after registration. This filtering is complementary to the correspondence-level pruning, which further removes geometrically matched points whose assigned matches are semantically inconsistent.

\begin{figure}[t]
    \centering
    \includegraphics[
        width=0.9\linewidth,
        height=0.26\textheight,
        keepaspectratio,
        trim={20pt 80pt 20pt 80pt},
        clip
    ]{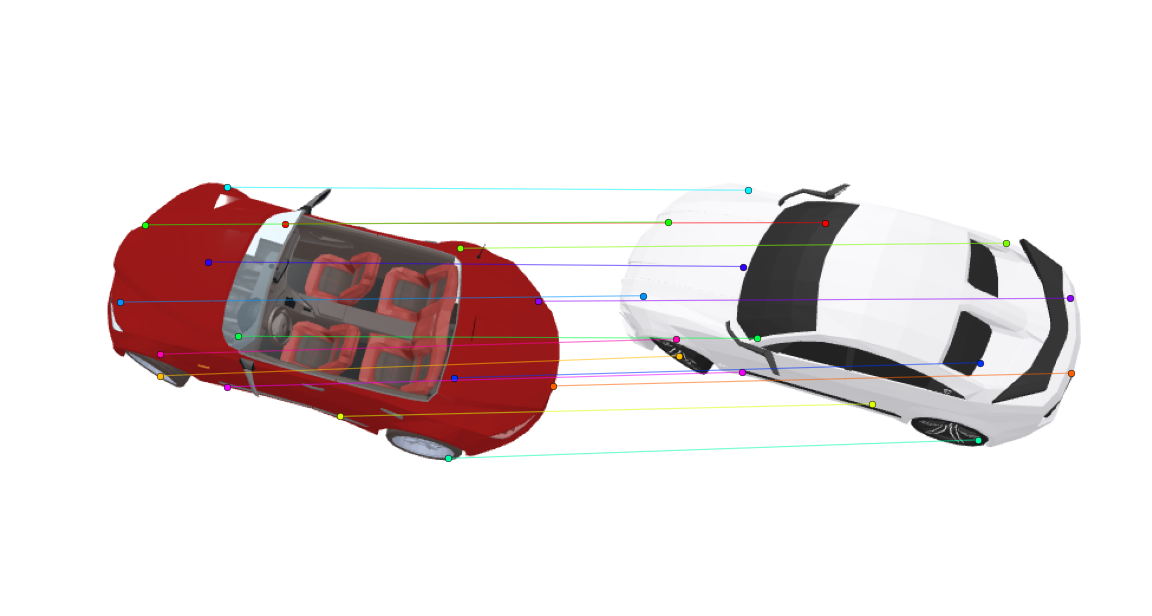}\\[-2mm]
    \includegraphics[
        width=0.9\linewidth,
        height=0.26\textheight,
        keepaspectratio,
        trim={20pt 50pt 20pt 80pt},
        clip
    ]{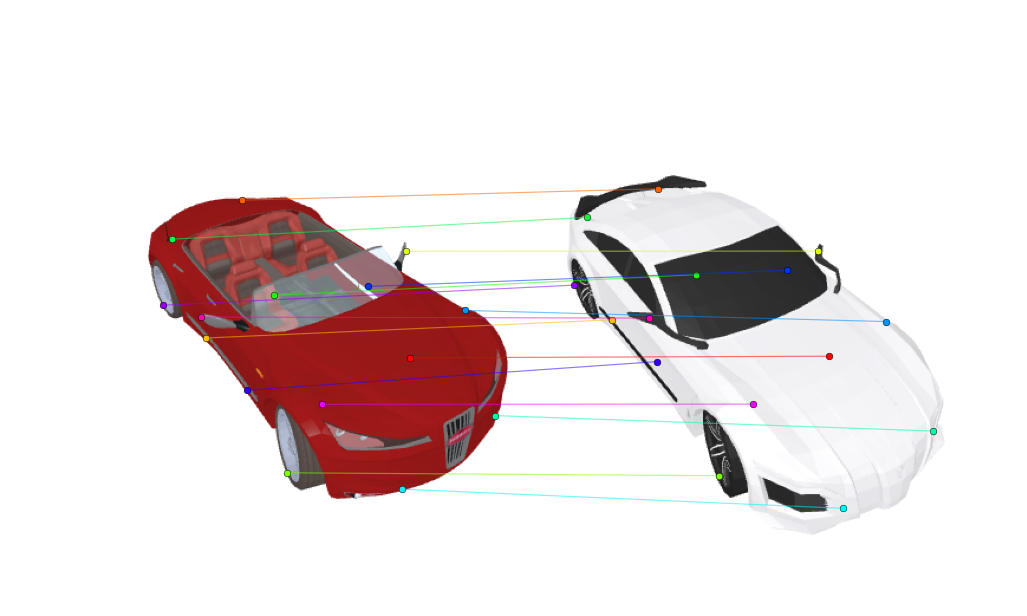}
    \caption{
    Sampled correspondences after semantic pruning. Each row shows a different view of the same object pair.
    }
    \label{fig:semantic_correspondence_pruning}
\end{figure}

We further visualize sampled correspondences established by our method in \autoref{fig:semantic_correspondence_pruning}. For visualization, we sample farthest points from the final correspondence set after semantic pruning. Each row shows a different view of the same object pair. As can be seen, semantically unmatched regions such as the exposed seats and steering wheel of the roofless car do not obtain correspondences. Similarly, the roof region of the white car is also removed, even though the CPA registration geometrically deforms these regions toward other object parts. These correspondences are filtered by our semantic correspondence pruning stage due to semantic inconsistency.
\clearpage

\section{3D--2D Correspondence Evaluation on KeypointNet}
\label{sup:3d2d_eval}

We further evaluate the quality of the 3D--2D correspondences
generated by SemGeo-Gen using KeypointNet~\cite{you2020}.
Since our 3D--3D correspondences are approximate, geometric
inaccuracies may introduce projection errors or visibility
inconsistencies when transferred to image space.
To assess their impact, we project the generated correspondences
onto rendered target views and evaluate their accuracy against
human-annotated semantic keypoints, accounting for visibility
and occlusion.

\paragraph{Evaluation protocol.}
We follow the 3D--3D evaluation protocol described in the main
paper, using the same 16 categories, model subsets, pair filtering,
and generated correspondences. Each target model is rendered
from four random viewpoints, with azimuth sampled from
$[0^\circ,360^\circ)$ and elevation from $[-10^\circ,70^\circ]$.

We evaluate only ground-truth keypoints visible in the rendered
target image, since occluded keypoints cannot provide valid
image-space annotations. As in our data-generation pipeline,
predicted correspondences are retained only if their target
points are also visible. Visibility is determined by comparing
each point's depth with the rendered depth map using a tolerance
of $0.02$ to account for small rasterization and surface
discrepancies.

For each visible ground-truth keypoint, we select the
correspondence with the nearest source point among those
with visible targets, provided its distance is within
$0.05$ times the target 3D bounding-box diagonal.
Keypoints without such a correspondence are considered
not found.

\paragraph{Evaluation metrics.}
We report PCK on found keypoints. A prediction is considered
correct at threshold $\alpha$ if
\begin{equation}
    \|\mathbf{u}_{pred}-\mathbf{u}_{gt}\|_2
    < \alpha d_{2D},
\end{equation}
where $\mathbf{u}_{pred}$ and $\mathbf{u}_{gt}$ denote
the projected predicted and ground-truth target locations,
and $d_{2D}$ is the projected 2D bounding-box diagonal
of the target object.

We additionally report coverage, defined as the fraction
of visible ground-truth keypoints for which a valid
correspondence is found (similar to the 3D-3D experiment). This measures coverage of the
sparse KeypointNet annotations rather than the fraction
of dense correspondences retained by our data-generation
pipeline.

\paragraph{Results.}
As shown in \autoref{tab:kpn_2d}, SemGeo-Gen achieves
94.2\% PCK@0.10 across 16 categories, compared to
92.7\% in the original 3D--3D evaluation over all
found keypoints. The higher overall 2D accuracy reflects
the restriction to visible keypoints, as required for
valid 3D--2D correspondences. When evaluated on the
same retained keypoints, 3D PCK reaches 95.1\%, compared
to 94.2\% in 2D. The slight decrease may arise from
geometric inaccuracies in approximate correspondences,
which can affect their projected locations and normalized
image-space errors. Nevertheless, the small gap indicates
that correspondence accuracy is largely preserved after
projection into image space.

The average coverage is 45.3\% of visible ground-truth
keypoints, reflecting the combined effect of correspondence
availability and visibility filtering. Results remain
stable across independently sampled viewpoint sets,
varying by at most $\pm0.1$ percentage points.

\begin{table*}[t]
    \centering
    \captionsetup{justification=centering}
    \small
    \setlength{\tabcolsep}{5pt}
    \renewcommand{\arraystretch}{0.95}

    \caption{
        \textbf{3D--2D correspondence evaluation on KeypointNet.}
        PCK on found correspondences with visible ground-truth
        keypoints and visible predicted target points.
        Coverage denotes the fraction of visible ground-truth
        keypoints for which a valid correspondence is found.
        The last column reports the original 3D PCK@0.10
        over all found keypoints.
    }
    \label{tab:kpn_2d}
    \vspace{4pt}

    \begin{tabular}{lccccc|c}
        \toprule
        \textbf{Category} &
        \textbf{PCK@.05} &
        \textbf{PCK@.10} &
        \textbf{PCK@.15} &
        \textbf{PCK@.20} &
        \textbf{Coverage} &
        \textbf{3D PCK@.10} \\
        \midrule
        Airplane   & 0.79 & 0.94 & 0.98 & 0.99 & 0.73 & 0.94 \\
        Bathtub    & 0.81 & 0.95 & 0.99 & 1.00 & 0.26 & 0.88 \\
        Bed        & 0.81 & 0.96 & 0.99 & 1.00 & 0.45 & 0.91 \\
        Bottle     & 0.81 & 0.95 & 0.98 & 0.99 & 0.40 & 0.96 \\
        Cap        & 0.83 & 0.97 & 1.00 & 1.00 & 0.30 & 0.97 \\
        Car        & 0.76 & 0.95 & 0.98 & 1.00 & 0.88 & 0.95 \\
        Chair      & 0.80 & 0.97 & 0.99 & 1.00 & 0.31 & 0.96 \\
        Guitar     & 0.81 & 0.97 & 0.99 & 1.00 & 0.61 & 0.96 \\
        Helmet     & 0.57 & 0.85 & 0.96 & 0.99 & 0.20 & 0.84 \\
        Knife      & 0.79 & 0.92 & 0.96 & 0.98 & 0.32 & 0.93 \\
        Laptop     & 0.93 & 1.00 & 1.00 & 1.00 & 0.55 & 0.99 \\
        Motorcycle & 0.70 & 0.90 & 0.97 & 0.99 & 0.74 & 0.93 \\
        Mug        & 0.79 & 0.97 & 0.99 & 1.00 & 0.59 & 0.96 \\
        Skateboard & 0.84 & 0.97 & 0.99 & 1.00 & 0.42 & 0.97 \\
        Table      & 0.87 & 0.98 & 1.00 & 1.00 & 0.40 & 0.96 \\
        Vessel     & 0.70 & 0.86 & 0.93 & 0.97 & 0.08 & 0.71 \\
        \midrule
        \textbf{Mean}
        & \textbf{0.79} & \textbf{0.94}
        & \textbf{0.98} & \textbf{0.99}
        & \textbf{0.45} & \textbf{0.93} \\
        \bottomrule
    \end{tabular}
\end{table*}

\section{Detailed PCK Results by Semantic Label and KeypointNet Category}
\label{Sup:per_label}

We provide a detailed analysis of correspondence quality at the level of individual semantic keypoints. For each category, we evaluate one subset of up to 100 objects (38 for Cap and 90 for Helmet) and all
induced source-target pairs, corresponding to up to $\binom{100}{2}$ object pairs per category, across
15 categories and 69{,}058 pairs overall, of which 46{,}849 pass the pair-level filter.

These results are computed using our full pipeline, including correspondence-level semantic filtering with $\tau_{\mathrm{sem}}=0.575$ and pair-level filtering. For each semantic label, we report results as $\text{correct}/\text{total}$, where \emph{total} denotes the number of ground-truth keypoints that remain evaluable after filtering. A keypoint is considered evaluable if at least one predicted correspondence remains within $0.05$ of the object bounding-box diagonal from the ground-truth location. We report raw $\text{correct}/\text{total}$ counts rather than percentages to make the amount of retained evaluable annotations explicit for each semantic label.

KeypointNet annotations are sparse and typically located on salient corners or extreme parts, while our method produces dense correspondences across the object surface. Therefore, these per-label results should be interpreted as a conservative estimate of correspondence quality around the sparse annotated locations.

The full per-label results are reported in 
Tables~\ref{tab:pck_airplane}-\ref{tab:pck_vessel}.

\begin{table}[h]
\centering
\small
\caption{Per-keypoint PCK results for category \texttt{airplane}. Values shown as $\text{correct}/\text{total}$.}
\label{tab:pck_airplane}
\begin{tabular}{cl ccc}
\toprule
ID & Description & PCK@0.1 & PCK@0.15 & PCK@0.2 \\
\midrule
0 & nose of the airplane & $3805/3834$ & $3829/3834$ & $3834/3834$ \\
1 & cockpit of the airplane & $2768/3289$ & $3113/3289$ & $3205/3289$ \\
2 & tail of the airplane & $3059/3457$ & $3437/3457$ & $3455/3457$ \\
3 & wingtip of the left wing of the airplane & $3597/3790$ & $3707/3790$ & $3716/3790$ \\
4 & wing root of the left wing of the airplane & $3046/3192$ & $3181/3192$ & $3192/3192$ \\
5 & wingtip of the right wing of the airplane & $3606/3795$ & $3709/3795$ & $3719/3795$ \\
6 & wing root of the right wing of the airplane & $3042/3195$ & $3182/3195$ & $3193/3195$ \\
7 & wingtip of the left horizontal stabilizer of the airplane & $3569/3711$ & $3615/3711$ & $3682/3711$ \\
8 & wingtip of the right horizontal stabilizer of the airplane & $3487/3719$ & $3534/3719$ & $3683/3719$ \\
9 & top of vertical stabilizer of the airplane & $3146/3369$ & $3307/3369$ & $3368/3369$ \\
10 & wingtip of the left diagonal stabilizer of the airplane & $34/34$ & $34/34$ & $34/34$ \\
11 & wingtip of the right diagonal stabilizer of the airplane & $38/38$ & $38/38$ & $38/38$ \\
12 & wing root of the left wing of the airplane & $2923/3095$ & $3063/3095$ & $3090/3095$ \\
13 & wing root of the right wing of the airplane & $2920/3093$ & $3056/3093$ & $3091/3093$ \\
14 & inner turbine of the left wing of the airplane & $1402/1422$ & $1421/1422$ & $1422/1422$ \\
15 & inner turbine of the right wing of the airplane & $1400/1424$ & $1423/1424$ & $1424/1424$ \\
16 & outer turbine of the left wing of the airplane & $105/130$ & $130/130$ & $130/130$ \\
17 & outer turbine of the right wing of the airplane & $117/138$ & $138/138$ & $138/138$ \\
18 & outer turbine of the right wing of the airplane & $0/0$ & $0/0$ & $0/0$ \\
\midrule
\multicolumn{2}{r}{Total} & $42064/44725$ & $43917/44725$ & $44414/44725$ \\
\bottomrule
\end{tabular}
\end{table}

\begin{table}[h]
\centering
\small
\caption{Per-keypoint PCK results for category \texttt{bathtub}. Values shown as $\text{correct}/\text{total}$.}
\label{tab:pck_bathtub}
\begin{tabular}{cl ccc}
\toprule
ID & Description & PCK@0.1 & PCK@0.15 & PCK@0.2 \\
\midrule
0 & top outer corner of the bathtub & $1340/1350$ & $1346/1350$ & $1350/1350$ \\
1 & top outer corner of the bathtub & $1300/1314$ & $1310/1314$ & $1314/1314$ \\
2 & top outer corner of the bathtub & $1277/1287$ & $1285/1287$ & $1287/1287$ \\
3 & top outer corner of the bathtub & $1234/1250$ & $1248/1250$ & $1250/1250$ \\
4 & bottom outer corner of the bathtub & $566/649$ & $595/649$ & $616/649$ \\
5 & bottom outer corner of the bathtub & $478/558$ & $511/558$ & $530/558$ \\
6 & bottom outer corner of the bathtub & $471/588$ & $526/588$ & $550/588$ \\
7 & bottom outer corner of the bathtub & $486/571$ & $530/571$ & $548/571$ \\
8 & top inner corner of the bathtub & $775/991$ & $925/991$ & $975/991$ \\
9 & top inner corner of the bathtub & $766/986$ & $908/986$ & $957/986$ \\
10 & top inner corner of the bathtub & $731/938$ & $878/938$ & $933/938$ \\
11 & top inner corner of the bathtub & $696/899$ & $848/899$ & $893/899$ \\
12 & bottom inner corner of the bathtub & $552/679$ & $635/679$ & $656/679$ \\
13 & bottom inner corner of the bathtub & $562/733$ & $666/733$ & $701/733$ \\
14 & bottom inner corner of the bathtub & $584/698$ & $651/698$ & $686/698$ \\
15 & bottom inner corner of the bathtub & $466/545$ & $508/545$ & $534/545$ \\
16 & center of the short side upper edge of the bathtub & $49/49$ & $49/49$ & $49/49$ \\
17 & center of the short side upper edge of the bathtub & $55/56$ & $56/56$ & $56/56$ \\
18 & center of the short side lower edge of the bathtub & $24/27$ & $27/27$ & $27/27$ \\
19 & center of the short side lower edge of the bathtub & $21/24$ & $24/24$ & $24/24$ \\
20 & center of the short side upper edge of the bathtub & $159/165$ & $162/165$ & $164/165$ \\
21 & center of the short side upper edge of the bathtub & $141/144$ & $143/144$ & $144/144$ \\
22 & center of the short side lower edge of the bathtub & $80/93$ & $92/93$ & $93/93$ \\
23 & center of the short side lower edge of the bathtub & $66/69$ & $69/69$ & $69/69$ \\
\midrule
\multicolumn{2}{r}{Total} & $12879/14663$ & $13992/14663$ & $14406/14663$ \\
\bottomrule
\end{tabular}
\end{table}

\begin{table}[h]
\centering
\small
\caption{Per-keypoint PCK results for category \texttt{bed}. Values shown as $\text{correct}/\text{total}$.}
\label{tab:pck_bed}
\begin{tabular}{cl ccc}
\toprule
ID & Description & PCK@0.1 & PCK@0.15 & PCK@0.2 \\
\midrule
0 & lower end of the front leg of the bed & $989/1044$ & $1032/1044$ & $1042/1044$ \\
1 & lower end of the headboard leg of the bed & $880/965$ & $924/965$ & $960/965$ \\
2 & lower end of the headboard leg of the bed & $825/906$ & $857/906$ & $901/906$ \\
3 & lower end of the front leg of the bed & $937/988$ & $979/988$ & $984/988$ \\
4 & foot corner of the mattress & $606/663$ & $640/663$ & $648/663$ \\
5 & head corner of the mattress & $420/511$ & $464/511$ & $491/511$ \\
6 & head corner of the mattress & $416/492$ & $461/492$ & $475/492$ \\
7 & foot corner of the mattress & $521/566$ & $550/566$ & $553/566$ \\
8 & foot corner of the upper mattress of the bunk bed & $489/521$ & $516/521$ & $516/521$ \\
9 & head corner of the upper mattress of the bunk bed & $464/507$ & $488/507$ & $496/507$ \\
10 & head corner of the upper mattress of the bunk bed & $493/545$ & $522/545$ & $527/545$ \\
11 & foot corner of the upper mattress of the bunk bed & $480/510$ & $507/510$ & $510/510$ \\
12 & upper corner of the bed frame of the bunk bed & $567/568$ & $568/568$ & $568/568$ \\
13 & upper corner of the bed frame of the bunk bed & $515/553$ & $525/553$ & $552/553$ \\
14 & upper corner of the bed frame of the bunk bed & $527/556$ & $537/556$ & $555/556$ \\
15 & upper corner of the bed frame of the bunk bed & $548/554$ & $554/554$ & $554/554$ \\
16 & upper corner of the footboard of the bed & $0/0$ & $0/0$ & $0/0$ \\
17 & upper corner of the headboard of the bed & $79/84$ & $83/84$ & $84/84$ \\
18 & upper corner of the headboard of the bed & $86/87$ & $87/87$ & $87/87$ \\
19 & upper corner of the footboard of the bed & $0/0$ & $0/0$ & $0/0$ \\
\midrule
\multicolumn{2}{r}{Total} & $9842/10620$ & $10294/10620$ & $10503/10620$ \\
\bottomrule
\end{tabular}
\end{table}

\begin{table}[h]
\centering
\small
\caption{Per-keypoint PCK results for category \texttt{bottle}. Values shown as $\text{correct}/\text{total}$.}
\label{tab:pck_bottle}
\begin{tabular}{cl ccc}
\toprule
ID & Description & PCK@0.1 & PCK@0.15 & PCK@0.2 \\
\midrule
0 & lower rim of the bottle & $2836/2866$ & $2860/2866$ & $2864/2866$ \\
1 & lower rim of the bottle & $2845/2868$ & $2866/2868$ & $2866/2868$ \\
2 & lower rim of the bottle & $2693/2711$ & $2707/2711$ & $2711/2711$ \\
3 & lower rim of the bottle & $2829/2861$ & $2857/2861$ & $2860/2861$ \\
4 & edge between the bottle body and the bottle neck & $243/261$ & $259/261$ & $261/261$ \\
5 & edge between the bottle body and the bottle neck & $257/275$ & $272/275$ & $275/275$ \\
6 & edge between the bottle body and the bottle neck & $250/268$ & $267/268$ & $268/268$ \\
7 & edge between the bottle body and the bottle neck & $256/273$ & $272/273$ & $272/273$ \\
8 & bottle cap center & $1438/1453$ & $1453/1453$ & $1453/1453$ \\
9 & edge of the bottle cap & $2096/2135$ & $2115/2135$ & $2134/2135$ \\
10 & edge of the bottle cap & $2095/2135$ & $2116/2135$ & $2134/2135$ \\
11 & edge of the bottle cap & $2088/2133$ & $2112/2133$ & $2132/2133$ \\
12 & edge of the bottle cap & $2083/2123$ & $2104/2123$ & $2123/2123$ \\
13 & lower edge of the bottle neck & $856/1005$ & $923/1005$ & $968/1005$ \\
14 & lower edge of the bottle neck & $824/970$ & $896/970$ & $933/970$ \\
15 & lower edge of the bottle neck & $805/961$ & $877/961$ & $921/961$ \\
16 & lower edge of the bottle neck & $798/958$ & $880/958$ & $922/958$ \\
17 & lower edge of the bottle neck & $0/0$ & $0/0$ & $0/0$ \\
\midrule
\multicolumn{2}{r}{Total} & $25292/26256$ & $25836/26256$ & $26097/26256$ \\
\bottomrule
\end{tabular}
\end{table}

\begin{table}[h]
\centering
\small
\caption{Per-keypoint PCK results for category \texttt{cap}. Values shown as $\text{correct}/\text{total}$.}
\label{tab:pck_cap}
\begin{tabular}{cl ccc}
\toprule
ID & Description & PCK@0.1 & PCK@0.15 & PCK@0.2 \\
\midrule
0 & top button of the cap & $528/558$ & $557/558$ & $558/558$ \\
1 & front of the visor of the cap & $558/558$ & $558/558$ & $558/558$ \\
2 & corner of the visor of the cap & $481/558$ & $546/558$ & $554/558$ \\
3 & corner of the visor of the cap & $484/558$ & $540/558$ & $555/558$ \\
4 & center of the back of the cap & $555/558$ & $558/558$ & $558/558$ \\
5 & center of the visor attachment of the cap & $493/502$ & $502/502$ & $502/502$ \\
\midrule
\multicolumn{2}{r}{Total} & $3099/3292$ & $3261/3292$ & $3285/3292$ \\
\bottomrule
\end{tabular}
\end{table}

\begin{table}[h]
\centering
\small
\caption{Per-keypoint PCK results for category \texttt{car}. Values shown as $\text{correct}/\text{total}$.}
\label{tab:pck_car}
\begin{tabular}{cl ccc}
\toprule
ID & Description & PCK@0.1 & PCK@0.15 & PCK@0.2 \\
\midrule
0 & front right wheel of the car & $3118/3118$ & $3118/3118$ & $3118/3118$ \\
1 & rear right wheel of the car & $3097/3100$ & $3100/3100$ & $3100/3100$ \\
2 & rear left wheel of the car & $3102/3113$ & $3113/3113$ & $3113/3113$ \\
3 & front left wheel of the car & $3107/3111$ & $3111/3111$ & $3111/3111$ \\
4 & corner of the front bumper of the car & $2890/2909$ & $2909/2909$ & $2909/2909$ \\
5 & corner of the rear bumper of the car & $3046/3067$ & $3067/3067$ & $3067/3067$ \\
6 & corner of the rear bumper of the car & $3068/3084$ & $3084/3084$ & $3084/3084$ \\
7 & corner of the front bumper of the car & $2919/2935$ & $2935/2935$ & $2935/2935$ \\
8 & top corner of the windshield of the car & $3044/3116$ & $3112/3116$ & $3115/3116$ \\
9 & top corner of the windshield of the car & $1919/2631$ & $2283/2631$ & $2521/2631$ \\
10 & top corner of the rear window of the car & $1976/2644$ & $2360/2644$ & $2558/2644$ \\
11 & top corner of the rear window of the car & $3053/3118$ & $3112/3118$ & $3118/3118$ \\
12 & bottom corner of the windshield of the car & $2817/3068$ & $3032/3068$ & $3064/3068$ \\
13 & bottom corner of the windshield of the car & $2892/3055$ & $3032/3055$ & $3053/3055$ \\
14 & bottom corner of the rear window of the car & $1830/2171$ & $2037/2171$ & $2117/2171$ \\
15 & bottom corner of the rear window of the car & $1821/2145$ & $2049/2145$ & $2119/2145$ \\
16 & right headlight of the car & $2857/2903$ & $2902/2903$ & $2903/2903$ \\
17 & right taillight of the car & $2907/2919$ & $2919/2919$ & $2919/2919$ \\
18 & left taillight of the car & $2906/2916$ & $2916/2916$ & $2916/2916$ \\
19 & left headlight of the car & $2849/2905$ & $2904/2905$ & $2905/2905$ \\
20 & right side mirror of the car & $2147/2231$ & $2225/2231$ & $2231/2231$ \\
21 & left side mirror of the car & $2158/2245$ & $2236/2245$ & $2245/2245$ \\
\midrule
\multicolumn{2}{r}{Total} & $59523/62504$ & $61556/62504$ & $62221/62504$ \\
\bottomrule
\end{tabular}
\end{table}

\begin{table}[h]
\centering
\small
\caption{Per-keypoint PCK results for category \texttt{chair}. Values shown as $\text{correct}/\text{total}$.}
\label{tab:pck_chair}
\begin{tabular}{cl ccc}
\toprule
ID & Description & PCK@0.1 & PCK@0.15 & PCK@0.2 \\
\midrule
0 & upper corner of the back of the chair & $743/779$ & $771/779$ & $778/779$ \\
1 & upper corner of the back of the chair & $713/759$ & $757/759$ & $758/759$ \\
2 & lower corner of the back of the chair & $501/548$ & $537/548$ & $547/548$ \\
3 & lower corner of the back of the chair & $489/539$ & $526/539$ & $537/539$ \\
4 & front corner of the seat of the chair & $822/840$ & $839/840$ & $840/840$ \\
5 & front corner of the seat of the chair & $816/840$ & $839/840$ & $840/840$ \\
6 & front corner of the armrest of the chair & $109/111$ & $110/111$ & $111/111$ \\
7 & back corner of the armrest of the chair & $67/80$ & $73/80$ & $77/80$ \\
8 & front corner of the armrest of the chair & $105/109$ & $109/109$ & $109/109$ \\
9 & back corner of the armrest of the chair & $71/83$ & $74/83$ & $78/83$ \\
10 & lower end of the front leg of the chair & $1/3$ & $1/3$ & $1/3$ \\
11 & lower end of the back leg of the chair & $1/3$ & $1/3$ & $1/3$ \\
12 & lower end of the front leg of the chair & $1/3$ & $1/3$ & $1/3$ \\
13 & lower end of the back leg of the chair & $1/3$ & $1/3$ & $1/3$ \\
14 & lower end of the leg of the chair & $1/3$ & $1/3$ & $1/3$ \\
15 & top of the chair cylinder & $9/9$ & $9/9$ & $9/9$ \\
16 & bottom of the chair cylinder & $8/10$ & $10/10$ & $10/10$ \\
17 & lower end of the front leg of the chair & $1322/1352$ & $1333/1352$ & $1339/1352$ \\
18 & lower end of the back leg of the chair & $1135/1165$ & $1156/1165$ & $1160/1165$ \\
19 & lower end of the back leg of the chair & $1147/1174$ & $1167/1174$ & $1171/1174$ \\
20 & lower end of the front leg of the chair & $1308/1342$ & $1332/1342$ & $1336/1342$ \\
\midrule
\multicolumn{2}{r}{Total} & $9370/9755$ & $9647/9755$ & $9705/9755$ \\
\bottomrule
\end{tabular}
\end{table}

\begin{table}[h]
\centering
\small
\caption{Per-keypoint PCK results for category \texttt{guitar}. Values shown as $\text{correct}/\text{total}$.}
\label{tab:pck_guitar}
\begin{tabular}{cl ccc}
\toprule
ID & Description & PCK@0.1 & PCK@0.15 & PCK@0.2 \\
\midrule
0 & head of the guitar & $3885/3904$ & $3904/3904$ & $3904/3904$ \\
1 & connection of the body and neck of the guitar & $2690/2926$ & $2883/2926$ & $2916/2926$ \\
2 & corner of the body of the guitar & $1964/2039$ & $2014/2039$ & $2033/2039$ \\
3 & corner of the body of the guitar & $2046/2060$ & $2057/2060$ & $2060/2060$ \\
4 & corner of the body of the guitar & $2104/2135$ & $2128/2135$ & $2135/2135$ \\
5 & corner of the body of the guitar & $2038/2130$ & $2112/2130$ & $2130/2130$ \\
6 & strap pin on the bottom of the body of the guitar & $2015/2236$ & $2102/2236$ & $2148/2236$ \\
7 & waist of the body of the guitar & $1966/2115$ & $2007/2115$ & $2085/2115$ \\
8 & waist of the body of the guitar & $1814/1856$ & $1855/1856$ & $1856/1856$ \\
\midrule
\multicolumn{2}{r}{Total} & $20522/21401$ & $21062/21401$ & $21267/21401$ \\
\bottomrule
\end{tabular}
\end{table}

\begin{table}[h]
\centering
\small
\caption{Per-keypoint PCK results for category \texttt{helmet}. Values shown as $\text{correct}/\text{total}$.}
\label{tab:pck_helmet}
\begin{tabular}{cl ccc}
\toprule
ID & Description & PCK@0.1 & PCK@0.15 & PCK@0.2 \\
\midrule
0 & top of the helmet & $1765/1976$ & $1949/1976$ & $1974/1976$ \\
1 & front of the helmet peak & $1331/1976$ & $1694/1976$ & $1872/1976$ \\
2 & back of the helmet peak & $1451/1976$ & $1789/1976$ & $1915/1976$ \\
3 & left side of the helmet peak & $1300/1976$ & $1738/1976$ & $1895/1976$ \\
4 & right side of the helmet peak & $1343/1976$ & $1762/1976$ & $1912/1976$ \\
5 & right bottom of the helmet & $1296/1834$ & $1592/1834$ & $1697/1834$ \\
6 & left bottom of the helmet & $1351/1834$ & $1635/1834$ & $1722/1834$ \\
7 & bottom of the visor of the helmet & $1265/1654$ & $1451/1654$ & $1588/1654$ \\
8 & bottom of the back of the helmet & $1590/1766$ & $1715/1766$ & $1743/1766$ \\
\midrule
\multicolumn{2}{r}{Total} & $12692/16968$ & $15325/16968$ & $16318/16968$ \\
\bottomrule
\end{tabular}
\end{table}

\begin{table}[h]
\centering
\small
\caption{Per-keypoint PCK results for category \texttt{laptop}. Values shown as $\text{correct}/\text{total}$.}
\label{tab:pck_laptop}
\begin{tabular}{cl ccc}
\toprule
ID & Description & PCK@0.1 & PCK@0.15 & PCK@0.2 \\
\midrule
0 & corner of the laptop body & $2862/2870$ & $2870/2870$ & $2870/2870$ \\
1 & hinge of the laptop & $2334/2342$ & $2342/2342$ & $2342/2342$ \\
2 & hint of the laptop & $2193/2223$ & $2222/2223$ & $2223/2223$ \\
3 & corner of the laptop body & $2774/2780$ & $2780/2780$ & $2780/2780$ \\
4 & corner of the laptop screen & $2778/2791$ & $2791/2791$ & $2791/2791$ \\
5 & corner of the laptop screen & $2763/2770$ & $2770/2770$ & $2770/2770$ \\
\midrule
\multicolumn{2}{r}{Total} & $15704/15776$ & $15775/15776$ & $15776/15776$ \\
\bottomrule
\end{tabular}
\end{table}

\begin{table}[h]
\centering
\small
\caption{Per-keypoint PCK results for category \texttt{motorcycle}. Values shown as $\text{correct}/\text{total}$.}
\label{tab:pck_motorcycle}
\begin{tabular}{cl ccc}
\toprule
ID & Description & PCK@0.1 & PCK@0.15 & PCK@0.2 \\
\midrule
0 & headlight of the motorcycle & $2392/2875$ & $2788/2875$ & $2864/2875$ \\
1 & end of the handlebar of the motorcycle & $3005/3549$ & $3320/3549$ & $3478/3549$ \\
2 & end of the handlebar of the motorcycle & $3009/3545$ & $3301/3545$ & $3481/3545$ \\
3 & seat of the motorcycle & $3763/4092$ & $4020/4092$ & $4087/4092$ \\
4 & axle of the front wheel of the motorcycle & $4225/4267$ & $4264/4267$ & $4267/4267$ \\
5 & axle of the back wheel of the motorcycle & $3873/3914$ & $3911/3914$ & $3914/3914$ \\
6 & bottom part of the frame of the motorcycle & $4261/4398$ & $4385/4398$ & $4397/4398$ \\
7 & axle of the front wheel of the motorcycle & $4202/4247$ & $4242/4247$ & $4247/4247$ \\
8 & axle of the back wheel of the motorcycle & $3910/3958$ & $3957/3958$ & $3958/3958$ \\
9 & front part of the front wheel of the motorcycle & $4502/4511$ & $4511/4511$ & $4511/4511$ \\
10 & back part of the back wheel of the motorcycle & $3983/4019$ & $4018/4019$ & $4019/4019$ \\
11 & top of the fuel tank of the motorcycle & $2583/3002$ & $2945/3002$ & $3002/3002$ \\
12 & front end of the front fender of the motorcycle & $2578/2817$ & $2800/2817$ & $2815/2817$ \\
13 & back end of the rear fender of the motorcycle & $1239/2041$ & $1654/2041$ & $1928/2041$ \\
\midrule
\multicolumn{2}{r}{Total} & $47525/51235$ & $50116/51235$ & $50968/51235$ \\
\bottomrule
\end{tabular}
\end{table}

\begin{table}[h]
\centering
\small
\caption{Per-keypoint PCK results for category \texttt{mug}. Values shown as $\text{correct}/\text{total}$.}
\label{tab:pck_mug}
\begin{tabular}{cl ccc}
\toprule
ID & Description & PCK@0.1 & PCK@0.15 & PCK@0.2 \\
\midrule
0 & mug rim & $1985/1996$ & $1996/1996$ & $1996/1996$ \\
1 & mug rim & $2038/2045$ & $2044/2045$ & $2045/2045$ \\
2 & mug rim & $2283/2307$ & $2305/2307$ & $2306/2307$ \\
3 & mug rim & $2274/2294$ & $2293/2294$ & $2294/2294$ \\
4 & mug base rim & $2718/2870$ & $2843/2870$ & $2868/2870$ \\
5 & mug base rim & $2733/2874$ & $2845/2874$ & $2874/2874$ \\
6 & mug base rim & $2642/2761$ & $2735/2761$ & $2760/2761$ \\
7 & mug base rim & $2655/2794$ & $2780/2794$ & $2794/2794$ \\
8 & upper attachment point for the mug handle & $2854/2897$ & $2897/2897$ & $2897/2897$ \\
9 & center of the mug handle & $2928/3179$ & $3159/3179$ & $3178/3179$ \\
10 & lower attachment point for the mug handle & $2597/2849$ & $2807/2849$ & $2841/2849$ \\
\midrule
\multicolumn{2}{r}{Total} & $27707/28866$ & $28704/28866$ & $28853/28866$ \\
\bottomrule
\end{tabular}
\end{table}

\begin{table}[h]
\centering
\small
\caption{Per-keypoint PCK results for category \texttt{skateboard}. Values shown as $\text{correct}/\text{total}$.}
\label{tab:pck_skateboard}
\begin{tabular}{cl ccc}
\toprule
ID & Description & PCK@0.1 & PCK@0.15 & PCK@0.2 \\
\midrule
0 & corner of the tail of the skateboard & $1308/1409$ & $1380/1409$ & $1404/1409$ \\
1 & corner of the tail of the skateboard & $1314/1421$ & $1386/1421$ & $1415/1421$ \\
2 & corner of the nose of the skateboard & $1289/1445$ & $1404/1445$ & $1440/1445$ \\
3 & corner of the nose of the skateboard & $1233/1395$ & $1337/1395$ & $1392/1395$ \\
4 & wheel of the skateboard & $2680/2694$ & $2692/2694$ & $2693/2694$ \\
5 & wheel of the skateboard & $2606/2622$ & $2615/2622$ & $2619/2622$ \\
6 & wheel of the skateboard & $2613/2639$ & $2634/2639$ & $2638/2639$ \\
7 & wheel of the skateboard & $2591/2608$ & $2604/2608$ & $2608/2608$ \\
8 & nose of the skateboard & $2028/2030$ & $2030/2030$ & $2030/2030$ \\
9 & tail of the skateboard & $1814/1814$ & $1814/1814$ & $1814/1814$ \\
\midrule
\multicolumn{2}{r}{Total} & $19476/20077$ & $19896/20077$ & $20053/20077$ \\
\bottomrule
\end{tabular}
\end{table}

\begin{table}[h]
\centering
\small
\caption{Per-keypoint PCK results for category \texttt{table}. Values shown as $\text{correct}/\text{total}$.}
\label{tab:pck_table}
\begin{tabular}{cl ccc}
\toprule
ID & Description & PCK@0.1 & PCK@0.15 & PCK@0.2 \\
\midrule
0 & corner of the table top & $1466/1523$ & $1515/1523$ & $1521/1523$ \\
1 & corner of the table top & $1604/1654$ & $1646/1654$ & $1652/1654$ \\
2 & corner of the table top & $1640/1676$ & $1675/1676$ & $1675/1676$ \\
3 & corner of the table top & $1638/1681$ & $1670/1681$ & $1674/1681$ \\
4 & end of the table leg & $2419/2584$ & $2547/2584$ & $2574/2584$ \\
5 & end of the table leg & $2443/2583$ & $2552/2583$ & $2580/2583$ \\
6 & end of the table leg & $2420/2592$ & $2557/2592$ & $2586/2592$ \\
7 & end of the table leg & $2435/2583$ & $2554/2583$ & $2578/2583$ \\
8 & end of the stretcher between the table legs & $0/0$ & $0/0$ & $0/0$ \\
9 & end of the stretcher between the table legs & $0/0$ & $0/0$ & $0/0$ \\
10 & end of the stretcher between the table legs & $0/0$ & $0/0$ & $0/0$ \\
11 & end of the stretcher between the table legs & $0/0$ & $0/0$ & $0/0$ \\
\midrule
\multicolumn{2}{r}{Total} & $16065/16876$ & $16716/16876$ & $16840/16876$ \\
\bottomrule
\end{tabular}
\end{table}

\begin{table}[h]
\centering
\small
\caption{Per-keypoint PCK results for category \texttt{vessel}. Values shown as $\text{correct}/\text{total}$.}
\label{tab:pck_vessel}
\begin{tabular}{cl ccc}
\toprule
ID & Description & PCK@0.1 & PCK@0.15 & PCK@0.2 \\
\midrule
0 & bow of a boat & $1133/1142$ & $1142/1142$ & $1142/1142$ \\
1 & lower part of the end of the bow of a boat (starboard) & $396/668$ & $553/668$ & $634/668$ \\
2 & upper part of the end of the bow of a boat (starboard) & $179/292$ & $245/292$ & $276/292$ \\
3 & upper part of the end of the bow of a boat (port) & $213/323$ & $277/323$ & $313/323$ \\
4 & lower part of the end of the bow of a boat (port) & $445/717$ & $611/717$ & $684/717$ \\
5 & lower corner of the boat stern (starboard & $207/265$ & $248/265$ & $257/265$ \\
6 & upper corner of the boat stern (starboard) & $97/113$ & $113/113$ & $113/113$ \\
7 & upper corner of the boat stern (port) & $104/118$ & $111/118$ & $117/118$ \\
8 & lower corner of the boat stern (port) & $211/251$ & $241/251$ & $247/251$ \\
9 & mast step of a sailboat & $23/28$ & $27/28$ & $28/28$ \\
10 & mast top of a sailboat & $72/72$ & $72/72$ & $72/72$ \\
11 & lower front corner of the deckhouse (starboard) & $45/142$ & $73/142$ & $94/142$ \\
12 & lower front corner of deckhouse (port) & $36/125$ & $66/125$ & $85/125$ \\
13 & lower aft corner of the deckhouse (starboard) & $29/76$ & $44/76$ & $54/76$ \\
14 & lower aft corner of the deckhouse (port) & $45/108$ & $67/108$ & $82/108$ \\
15 & upper front corner of the deckhouse (starboard) & $11/26$ & $18/26$ & $22/26$ \\
16 & upper front corner of the deckhouse (port) & $12/20$ & $16/20$ & $18/20$ \\
17 & upper aft corner of the deckhouse (starboard) & $14/25$ & $16/25$ & $21/25$ \\
18 & upper aft corner of the deckhouse (port) & $15/24$ & $17/24$ & $22/24$ \\
19 & foremost elevation point on the deck of the ship & $51/58$ & $54/58$ & $57/58$ \\
20 & lower corner of the windshield of the ship (starboard) & $2/2$ & $2/2$ & $2/2$ \\
21 & lower corner of the windshield of the ship (port) & $2/2$ & $2/2$ & $2/2$ \\
22 & upper corner of the windshield of the ship (starboard) & $0/0$ & $0/0$ & $0/0$ \\
23 & upper corner of the windshield of the ship (port) & $0/1$ & $1/1$ & $1/1$ \\
\midrule
\multicolumn{2}{r}{Total} & $3342/4598$ & $4016/4598$ & $4343/4598$ \\
\bottomrule
\end{tabular}
\end{table}

\clearpage

\begin{table}[h]
\centering
\caption{PCK averaged over seeds, semantic\_threshold = 0.625}
\label{tab:pck_thresh_0_75}
\begin{tabular}{lcccc}
\toprule
Category & PCK@0.05 & PCK@0.10 & PCK@0.15 & PCK@0.20 \\
\midrule
airplane & 0.80 & 0.94 & 0.98 & 0.99 \\
bathtub & 0.72 & 0.89 & 0.96 & 0.99 \\
bed & 0.69 & 0.91 & 0.97 & 0.99 \\
bottle & 0.82 & 0.96 & 0.99 & 0.99 \\
cap & 0.80 & 0.97 & 1.00 & 1.00 \\
car & 0.78 & 0.96 & 0.99 & 1.00 \\
chair & 0.78 & 0.97 & 0.99 & 0.99 \\
guitar & 0.81 & 0.96 & 0.98 & 0.99 \\
helmet & 0.56 & 0.84 & 0.95 & 0.98 \\
knife & 0.85 & 0.93 & 0.97 & 0.98 \\
laptop & 0.89 & 0.99 & 1.00 & 1.00 \\
motorcycle & 0.76 & 0.94 & 0.98 & 1.00 \\
mug & 0.73 & 0.96 & 0.99 & 1.00 \\
skateboard & 0.87 & 0.98 & 0.99 & 1.00 \\
table & 0.77 & 0.96 & 0.99 & 1.00 \\
vessel & 0.44 & 0.69 & 0.86 & 0.94 \\
\midrule
Mean & 0.75 & 0.93 & 0.97 & 0.99 \\
Median & 0.78 & 0.96 & 0.98 & 0.99 \\
\bottomrule
\end{tabular}
\end{table}

\begin{table}[h]
\centering
\caption{PCK averaged over seeds, semantic\_threshold = 0.575}
\label{tab:pck_thresh_0_85}
\begin{tabular}{lcccc}
\toprule
Category & PCK@0.05 & PCK@0.10 & PCK@0.15 & PCK@0.20 \\
\midrule
airplane & 0.79 & 0.94 & 0.98 & 0.99 \\
bathtub & 0.70 & 0.88 & 0.96 & 0.99 \\
bed & 0.68 & 0.91 & 0.96 & 0.99 \\
bottle & 0.83 & 0.96 & 0.98 & 0.99 \\
cap & 0.80 & 0.97 & 1.00 & 1.00 \\
car & 0.78 & 0.95 & 0.99 & 1.00 \\
chair & 0.77 & 0.96 & 0.99 & 0.99 \\
guitar & 0.81 & 0.96 & 0.98 & 0.99 \\
helmet & 0.55 & 0.84 & 0.94 & 0.98 \\
knife & 0.83 & 0.93 & 0.96 & 0.98 \\
laptop & 0.89 & 0.99 & 1.00 & 1.00 \\
motorcycle & 0.75 & 0.93 & 0.98 & 0.99 \\
mug & 0.73 & 0.96 & 0.99 & 1.00 \\
skateboard & 0.86 & 0.97 & 0.99 & 1.00 \\
table & 0.76 & 0.96 & 0.99 & 1.00 \\
vessel & 0.45 & 0.71 & 0.87 & 0.94 \\
\midrule
Mean & 0.75 & 0.93 & 0.97 & 0.99 \\
Median & 0.78 & 0.96 & 0.98 & 0.99 \\
\bottomrule
\end{tabular}
\end{table}

\begin{table}[h]
\centering
\caption{PCK averaged over seeds, semantic\_threshold = 0.525}
\label{tab:pck_thresh_0_95}
\begin{tabular}{lcccc}
\toprule
Category & PCK@0.05 & PCK@0.10 & PCK@0.15 & PCK@0.20 \\
\midrule
airplane & 0.78 & 0.94 & 0.98 & 0.99 \\
bathtub & 0.68 & 0.87 & 0.96 & 0.99 \\
bed & 0.67 & 0.90 & 0.96 & 0.98 \\
bottle & 0.82 & 0.95 & 0.98 & 0.99 \\
cap & 0.79 & 0.97 & 1.00 & 1.00 \\
car & 0.78 & 0.95 & 0.98 & 0.99 \\
chair & 0.75 & 0.96 & 0.99 & 0.99 \\
guitar & 0.80 & 0.96 & 0.98 & 0.99 \\
helmet & 0.53 & 0.83 & 0.94 & 0.98 \\
knife & 0.82 & 0.92 & 0.96 & 0.98 \\
laptop & 0.90 & 0.99 & 1.00 & 1.00 \\
motorcycle & 0.74 & 0.92 & 0.97 & 0.99 \\
mug & 0.74 & 0.96 & 0.99 & 1.00 \\
skateboard & 0.86 & 0.97 & 0.99 & 1.00 \\
table & 0.75 & 0.95 & 0.99 & 1.00 \\
vessel & 0.47 & 0.73 & 0.88 & 0.95 \\
\midrule
Mean & 0.74 & 0.92 & 0.97 & 0.99 \\
Median & 0.76 & 0.95 & 0.98 & 0.99 \\
\bottomrule
\end{tabular}
\end{table}

\begin{table}[h]
\centering
\caption{PCK averaged over seeds, semantic\_threshold = 	0.25}
\label{tab:pck_thresh_1_5}
\begin{tabular}{lcccc}
\toprule
Category & PCK@0.05 & PCK@0.10 & PCK@0.15 & PCK@0.20 \\
\midrule
airplane & 0.76 & 0.92 & 0.97 & 0.99 \\
bathtub & 0.53 & 0.79 & 0.93 & 0.98 \\
bed & 0.62 & 0.86 & 0.93 & 0.97 \\
bottle & 0.75 & 0.91 & 0.97 & 0.99 \\
cap & 0.79 & 0.95 & 0.99 & 1.00 \\
car & 0.77 & 0.95 & 0.98 & 0.99 \\
chair & 0.68 & 0.93 & 0.98 & 0.99 \\
guitar & 0.78 & 0.95 & 0.98 & 0.99 \\
helmet & 0.45 & 0.77 & 0.91 & 0.96 \\
knife & 0.73 & 0.87 & 0.94 & 0.97 \\
laptop & 0.90 & 0.99 & 1.00 & 1.00 \\
motorcycle & 0.70 & 0.89 & 0.96 & 0.99 \\
mug & 0.74 & 0.96 & 0.99 & 1.00 \\
skateboard & 0.82 & 0.95 & 0.98 & 1.00 \\
table & 0.72 & 0.93 & 0.98 & 0.99 \\
vessel & 0.44 & 0.72 & 0.86 & 0.93 \\
\midrule
Mean & 0.70 & 0.90 & 0.96 & 0.98 \\
Median & 0.74 & 0.92 & 0.98 & 0.99 \\
\bottomrule
\end{tabular}
\end{table}

\clearpage

\end{document}